%% file: arxiv_submission.tex
\documentclass{article}
\usepackage{ijcai26}

\usepackage{times}
\usepackage{soul}
\usepackage{url}
\usepackage[hidelinks]{hyperref}

\usepackage[utf8]{inputenc}
\usepackage[small]{caption}
\usepackage{graphicx}
\usepackage{amsmath}
\usepackage{amsthm}
\usepackage{algorithm}
\usepackage{algorithmic}
\usepackage[switch]{lineno}
\usepackage{enumitem}
\usepackage{mathtools}
\usepackage{siunitx}
\newcommand{\mnum}[1]{\makebox[2.3em][r]{#1}}
\newcommand{\msep}{\hspace{0pt}}

\newtheorem{theorem}{Theorem}

\title{Probably Approximately Correct\\Maximum A Posteriori Inference}

\author{
Matthew Shorvon$^1$
\and
Frederik Mallmann-Trenn$^1$\and
David S. Watson$^{1}$
\affiliations
$^1$Department of Informatics, King's College London\\
\emails
\{matthew.shorvon, frederik.mallmann-trenn, david.watson\}@kcl.ac.uk
}

\usepackage[round]{natbib}

\usepackage{amsmath} 
\usepackage{amssymb}  
\usepackage{pifont}
\DeclareMathOperator*{\argmax}{argmax}

\usepackage{tikz}
\usepackage{pctex}
\usepackage{algorithm}
\usepackage{verbatim}
\usepackage{subcaption}
\usepackage{amsthm}
\usepackage{natbib}
\usepackage{algorithmic}
\usepackage{soul}
\usepackage{tabularx}
\usepackage{booktabs}
\usepackage{xcolor}
\usepackage[font={small}]{caption}

\theoremstyle{definition}

\newtheorem{lemma}{Lemma}

\newtheorem{corollary}{Corollary}[theorem]

\newcommand{\fnote}[1]{
\textsf{{\color{teal}***F:#1***}}}

\begin{document}

\maketitle

\begin{abstract}
Computing the conditional mode of a distribution, better known as the \textit{maximum a posteriori} (MAP) assignment, is a fundamental task in probabilistic inference. However, MAP is generally intractable, and remains hard even under many common structural constraints and approximation schemes. 
We take a novel approach inspired by multi-armed bandits, recasting MAP as a best arm identification task.
We introduce \textit{probably approximately correct} (PAC) algorithms for MAP that provide provably optimal solutions in both the fixed-confidence and fixed-budget regimes. 
We characterize tractability conditions using information theoretic measures that can be estimated from finite samples. 
Our PAC-MAP solvers are efficiently implemented using probabilistic circuits and graphical models with appropriate architectures. 
The algorithms we develop can be used either as standalone MAP estimators or to improve on standard heuristics, fortifying their solutions with rigorous guarantees.
Experiments confirm the benefits of our method in a range of benchmarks.
\end{abstract}

\input{main_text}



\section*{Acknowledgments}


MS was supported by the UKRI CDT in Safe and Trusted Artificial Intelligence, grant number EP/S023356/1.\footnote{\url{https://safeandtrustedai.org/}.} DSW was supported by EPSRC grant number UKRI918.

\bibliographystyle{named}
\bibliography{bibbythebib}

\appendix
\input{arxiv_appx}


\end{document}

%% file: main_text.tex
\section{Introduction}
Given evidence $\bm E=\bm e$, what is the most likely value of one or more query variables $\bm Q$? Solving such questions is a fundamental task in probabilistic inference, where it is generally known as the \textit{maximum a posteriori} (MAP) problem. It comes up naturally in prediction and planning, where we frequently need to make decisions under uncertainty on the basis of available information. Common examples include diagnosing disease status from physical symptoms or correcting corrupted portions of an image based on the content of surrounding pixels. 

Supervised learning provides adaptive procedures for tackling such problems, but these methods face important limitations. In the continuous setting, regression algorithms are generally designed to estimate conditional \textit{expectations}, which may correspond to low-density regions of the target distribution. Classifiers, on the other hand, typically compute MAP by default, if constrained to return a single label. However, supervised learners of either type are \textit{discriminative} models that require an \textit{a priori} selection of the outcome variable $Y$ and conditioning on a complete feature vector $\bm X = \bm x$. 
A full \textit{generative} model, by contrast, allows users to compute MAP for arbitrary query variables and potentially use partial evidence, conditioning on a strict subset of the available features $\bm e \subsetneq \bm x$. 
When query and evidence variables exhaust the feature space, a MAP solution is also called the \textit{most probable explanation} (MPE); when nuisance variables must be ignored, we have an instance of marginal MAP (MMAP). We will generally use MAP as the umbrella term, unless it is important to specify that we are in one setting or the other.


These maximization problems have been widely studied in probabilistic reasoning, e.g. via Bayesian networks (BNs) \citep{darwiche2009modeling}, probabilistic graphical models (PGMs) \citep{koller2009}, and probabilistic circuits (PCs) \citep{vergari_tutorial}.
The latter formalism is especially useful in analyzing the complexity of MAP, as salient properties of PCs can often guarantee the existence of polytime solutions. 
In general, however, MPE is $\text{NP}$-complete \citep{SHIMONY1994399}, while MMAP is
$\text{NP}^{\text{PP}}$-complete \citep{park_darwiche_2004}. Both tasks remain intractable even under various constraints on graph structure and variable cardinality \citep{deCampos2011, decampos_no_news}.
Previous authors \citep{darwiche2009modeling, choi2020} have established sufficient conditions for exact polytime MPE inference in PCs, including a property known as \textit{determinism} (formally defined in Appx. D).
This constraint is highly restrictive and limits the expressive capacity of resulting circuits. 



We take a different approach, striking a pragmatic balance between accuracy and efficiency to tackle MAP (with or without nuisance variables). Specifically, we propose methods for \textit{probably approximately correct} (PAC)-MAP inference, recasting the problem as a best arm identification task. These methods are not guaranteed to produce exact or even $\varepsilon$-approximate MAP solutions (although, as we demonstrate below, they can and often do). 
Instead, we provide a strictly weaker guarantee: for user-supplied parameters $\varepsilon, \delta \in (0,1)$, PAC-MAP solutions are within a factor $1-\varepsilon$ of the true mode, with probability at least $1 - \delta$. 
Our results illustrate that introducing nonzero error tolerance and failure probability can improve practical performance in many cases. 
Though randomized MAP solvers have been proposed in the Bayesian modeling literature \citep{wang2018, hazan2019, Tolpin_Wood_2021}, these approaches generally rely on strong parametric assumptions and come with few or no theoretical guarantees. 
To the best of our knowledge, our work is the first to take a bandit approach to MAP inference.


Our primary contributions are fourfold. (1) We reformulate MAP as a best arm identification task and develop provably optimal PAC-MAP solvers under minimal assumptions. (2) We characterize tractability conditions in terms of information theoretic quantities that can be estimated or bounded from finite samples. (3) When fixed-confidence results are infeasible, we invert our methods for fixed-budget solutions, deriving Pareto-optimal PAC-MAP policies. (4) We adapt our algorithms to issue PAC certificates for \textit{any} MAP approximator, providing rigorous guarantees for otherwise heuristic solutions. The remainder of this paper is structured as follows. Following a brief review of formal definitions in Sect. \ref{section:background}, we introduce and study oracle versions of our PAC-MAP solvers in Sect. \ref{section:method}. Experimental results are presented in Sect. \ref{section:experiments}, followed by a brief discussion in Sect. \ref{section:discussion}. 
Sect. \ref{section:conclusion} concludes.

\section{Background} \label{section:background}

Let $\bm Z \in \mathcal Z \subseteq \mathbb R^{n_Z}$ be a collection of random variables with joint distribution $P$ and probability mass/density function $p$. 
We partition the features into three disjoint sets, $\bm Z = \bm Q \cup \bm E \cup \bm V$, representing queries, evidence, and nuisance variables, respectively. 
The maximization task is to compute the assignment $\bm q^*$ (not necessarily unique) satisfying:
\begin{align*}
    \bm q^* := \argmax_{\bm q} ~\sum_{\bm v} ~p(\bm q, \bm v \mid \bm e). 
\end{align*}
The corresponding mass/density is $p^* := p(\bm q^*\mid \bm e)$. 
When $\bm V = \emptyset$, we have an instance of MPE; otherwise, we have an instance of MMAP. 


Even with binary data, MAP is notoriously difficult. Let $n$ be the dimensionality of the query space $\mathcal Q$. Because the number of possible assignments to queries grows exponentially with $n$, brute force strategies for solving the problem require $\Omega\big(\exp(n)\big)$ operations in the worst case, even with access to an oracle for computing marginals. 
Indeed, both MPE and MMAP are known to be intractable \citep{SHIMONY1994399, park_darwiche_2004}. Neither problem can be efficiently approximated to within a constant factor. 

These considerations motivate a randomized approach. We say that the PAC-MAP solver $\pi$ is \textit{sound} if, for any $\varepsilon, \delta \in (0,1)$, $\pi$ returns some $\hat{\bm q}$ such that
\begin{align*}
    \Pr\big( p(\hat{\bm q} \mid \bm e) \geq p^* (1- \varepsilon)  \big) \geq 1-\delta,
\end{align*}
with randomness over the sampling distribution of candidates, $\bm q \sim P(\bm Q \mid \bm e)$.
If the inequality holds, we say that the true maximizer $\bm q^*$ is \textit{PAC-identified}.
Moreover, $\pi$ is \textit{tractable} if its runtime can be upperbounded by a polynomial function of $n$ 
and $\log (1 / \delta)$.\footnote{Typically, we include $1/\varepsilon$ among the inputs that characterize feasibility for PAC algorithms. This is not the case for PAC-MAP. See Appx. A.} 
We say that $\pi$ is \textit{purely random} if its search procedure relies exclusively on i.i.d. sampling, as opposed to \textit{adaptive} strategies that employ a mix of exploration and exploitation. 

An important point to acknowledge upfront is that we cannot guarantee the existence of tractable PAC-MAP solutions for arbitrary distributions. 
Indeed, for sufficiently small $\varepsilon, \delta$, the task is effectively indistinguishable from classical MAP, and therefore at least NP-hard.
However, sampling provides a principled strategy for navigating the state space and qualifying our best guess with appropriate precision and confidence.

\paragraph{Bandits}
Our approach builds on foundational work in multi-armed bandits \citep{Lattimore2020bandit}.  
Specifically, we treat MAP as a \textit{best arm identification} (BAI) task, in which each $\bm q \in \mathcal Q$ represents an arm with reward \mbox{$p(\bm q \mid \bm e)$} (or any rank-preserving transformation thereof).
There is a rich literature on BAI, including provably optimal procedures under various assumptions on arms and rewards \citep{audibert2010, garivier2016, russo2016, kaufman2016}. 

Our setting differs from the classic setup in several important respects. We assume access to a \textit{sampler}, which generates i.i.d. draws from the distribution $P(\bm Q \mid \bm e)$; and an \textit{oracle}, which provides exact solutions to queries of the form $p(\bm q \mid \bm e) = ~?$, for any $\bm q \in \mathcal Q$. 
These assumptions are reasonable given access to graphical models or circuits representing the target distribution. 
Our sampler and oracle remove two difficulties that dominate classic BAI---the former by fixing the exploration policy upfront, the latter by removing noise from the reward signal. These gains are offset by the complication that the support of $P(\bm Q \mid \bm e)$ may grow exponentially with $n$, making it impossible to pull every arm.

Several authors have studied the case of bandits with infinite or continuous arms \citep{kleinberg2008, wang2008, bubeck2011, carpentier2015}.
The goal in these works is generally to minimize some notion of \textit{regret}, i.e. the expected gap to the optimal arm. 
In BAI, the focus is typically on \textit{simple} (as opposed to cumulative) regret. For our purposes, we define this quantity as follows. Let $\pi = (\pi_t)_{t=1}^{m+1}$ be a policy that, at each step $t$, returns some $\hat{\bm q}_t$ with mass/density $\hat p_t = p(\hat{\bm q}_t \mid \bm e)$. Then $\pi$'s simple regret at time $m$ is:
\begin{align*}
    R_m(\pi, p) := \mathbb E_{p \pi}\big[\log_2 (p^* / \hat p_{m+1})\big],
\end{align*}
which represents the expected gap in bits between the true MAP and $\pi$'s output after $m$ steps. 
Though we derive regret bounds for our algorithms below, our primary objective is different and arguably more ambitious: to return optimal PAC solutions under fixed confidence or sample budget. With infinite arms, such guarantees have been obtained only under a reservoir model, where the target is a quantile rather than the mode \citep{aziz2018, gong2023}. The latter is provably unidentifiable in this setting because near-optimal arms may be arbitrarily rare and rewards are noisy. Neither obstruction arises in our setting: reward-proportional sampling naturally steers us toward promising candidates, while the oracle eliminates reward noise. We thus target the true mode directly, with tractability governed by how sharply the distribution concentrates. 
Our challenge is not to separate arms with similar rewards, as in standard BAI, but to ensure that we have covered enough of the arm space to issue a valid PAC certificate.

\section{Oracle PAC-MAP}\label{section:method}

In this section, we propose and study novel algorithms for MAP inference.

\subsection{Binary data}
Consider the binary setting, with $\mathcal Q = \{0,1\}^n$. Generalization to arbitrary discrete queries is straightforward, while continuous data is handled separately below. 



\paragraph{Purely random PAC-MAP}
A trivial strategy for randomized MAP inference is to draw $m$ samples from $P(\bm Q \mid \bm e)$ and return the maximizing assignment 
\begin{align*}
    \hat{\bm q} := \argmax_{i \in [m]} ~p(\bm q_i \mid \bm e).
\end{align*}
This is what \citet[Ch.~33]{Lattimore2020bandit} call \textit{uniform exploration}. 
We call this na\"{i}ve approach Alg. 0.

\begin{lemma}[Discovery complexity]\label{lemma:sample_complexity}
    Fix any $\varepsilon, \delta \in (0,1)$. 
    Define the $\varepsilon$-superlevel set: 
    \begin{align*}
    G_\varepsilon := \big\{\bm q: p(\bm q \mid \bm e) \geq p^*(1 - \varepsilon)\big\},
    \end{align*}
    with corresponding mass $\mu_\varepsilon = \sum_{\bm q \in G_\varepsilon} p(\bm q \mid \bm e)$. 
    Then Alg.~0 PAC-identifies $\bm q^*$ in $\mathcal O\big(\mu_\varepsilon^{-1} \log\delta^{-1}\big)$ samples.
\end{lemma}
\noindent (All proofs in Appx. A.) Lemma \ref{lemma:sample_complexity} bounds the complexity of a straightforward sampling approach. 
Of course, Alg. 0 is not much use without prior information on $p^*$ or $\mu_\varepsilon$.
However, the result highlights an important distinction between \textit{discovering} and \textit{certifying} PAC-MAP solutions. 
Just as computing and verifying answers to decision problems characterize distinct complexity classes (assuming the polynomial hierarchy does not collapse), we must distinguish between the two aspects of PAC-MAP. 
Suppose that, unbeknownst to us, the true distribution is uniform. Then \textit{discovering} a solution is trivial, since every $\bm q$ is in $G_\varepsilon$ and so $\mu_\varepsilon=1$. \textit{Certifying} that a candidate solution is PAC, by contrast, is maximally challenging, since we must search through an exponentially large number of candidates before we can be confident that we have not missed any more probable atoms. 
\begin{algorithm}[tb] \small
\caption{Purely random PAC-MAP: $\pi_{\mathrm R}(P, p, \varepsilon, \delta)$}
\label{alg:pacmap_b}
\textbf{Input}: Sampler $P$, oracle $p$, tolerance $\varepsilon$, level $\delta$\\
\textbf{Output}: PAC-MAP solution $\hat{\bm q}, \hat p$\\
\begin{algorithmic}[1]
\STATE $S \gets \emptyset$ \quad \texttt{// Initialize candidate set}
\STATE $m \gets 0$ \quad \texttt{// Initialize sample count}
\STATE $M \gets \infty$ \quad \texttt{// Initialize stop time}

\WHILE{$m < M$}
    \STATE $m \gets m + 1$ \quad \texttt{// Update count}
    \STATE $\bm q \sim P(\bm Q \mid \bm e)$ \quad \texttt{// Draw sample} 
    \STATE $S \gets S \cup \bm q$ \quad \texttt{// Admit new candidate}
    \STATE $\hat{\bm{q}} \leftarrow \argmax_{\bm q \in S} ~p(\bm q \mid \bm{e})$ \quad \texttt{// Leading arm} 
    \STATE $\hat p \gets p(\hat{\bm q} \mid \bm e)$ \quad \texttt{// Associated probability}
    
    \STATE $\check p \gets 1 - \sum_{\bm q \in S} ~p(\bm q \mid \bm e)$ \texttt{// Residual mass}
    \IF{$\hat p \geq \check p(1 - \varepsilon)$}
        \STATE \textbf{break} \quad \texttt{// Algorithm has converged}
    \ELSE 
        \STATE $M \gets \hat p^{-1} ~(1 - \varepsilon) ~\log  \delta^{-1} $ ~~\texttt{// Update stop time}
    \ENDIF
\ENDWHILE

\RETURN $\hat{\bm q}, ~\hat p$
\end{algorithmic}
\end{algorithm}

In lieu of any distributional assumptions, we need some adaptive stopping criteria.
Our strategy, outlined in Alg. \ref{alg:pacmap_b}, is to approach the maximal probability $p^*$ from both directions: a lower bound $\hat p$ that increases every time we find a new maximizer among the sampled atoms, and an upper bound $\check p$ that decreases as we gradually remove assignments from consideration. 
If $\hat p$ rises above $\check p(1-\varepsilon)$, the algorithm halts, since this ensures an $\varepsilon$-approximate solution at any $\delta$. (Note that the solution is exact if $\hat p \geq \check p$.) 
Otherwise, we continue sampling until we reach the threshold $\hat p^{-1} (1 - \varepsilon) \log \delta^{-1}$. If this stopping criterion is triggered, we issue a PAC certificate for the current leading candidate $\hat{\bm q}$. We call this procedure $\pi_{\mathrm R}$, for \textit{random}.

$\pi_{\mathrm R}$ can be uncompetitive in high-entropy distributions. For example, with access to a pmf oracle, we can always solve MAP exactly in $2^n$ queries by simply enumerating all possible assignments. $\pi_{\mathrm R}$ requires $2^n (1 - \varepsilon) \log \delta^{-1}$ samples 
to PAC-identify $\bm q^*$ in the case of a uniform pmf, which exceeds the trivial bound whenever $(1 - \varepsilon) \log \delta^{-1} > 1$. However, $\pi_{\mathrm R}$ may be far more efficient than brute force methods when mass concentrates. 
To formalize this, we use the \textit{min-entropy}, aka the R\'{e}nyi entropy of infinite order:
\begin{align*}
    H_\infty(\bm Q \mid \bm e) := -\log_2 p^*,
\end{align*}
which represents the number of bits encoded by the MAP configuration. We henceforth abbreviate this statistic by $h := H_\infty(\bm Q \mid \bm e)$.
We also make use of the R\'{e}nyi divergence of infinite order between our true $P$ and a uniform $U$:
\begin{align*}
    D_\infty(P~||~U) &:= \log_2(p^* / 2^{-n}) = n - h,
\end{align*}
which measures how far the MAP state is from a flat baseline. We henceforth abbreviate this statistic by $d := D_\infty(P~||~U)$.


\begin{theorem}[Purely random PAC-MAP]
\label{thm:pacmap_bin}
    Let $\Pi$ be the class of purely random PAC-MAP solvers, with sound policy subset $\Pi^* \subset \Pi$. 
    Let $\mathcal P$ be the class of pmfs on $\mathcal Q$ with min-entropy $h$.
    We have:
    \begin{itemize}[noitemsep]
        \item[(a)] $\pi_{\mathrm{R}} \in \Pi^*$.
        \item[(b)] With probability at least $1 - \delta$, $\pi_{\mathrm{R}}$ terminates after $\mathcal O\big(\exp(h) \log \delta^{-1}\big)$ random samples. 
        This rate is uniformly optimal w.r.t. $\Pi^*$: for every $P \in \mathcal P$ and any $\{\bm q_i\}_{i=1}^m \sim P$, no sound policy terminates faster. 
        \item[(c)] After $m$ random samples, $\pi_{\mathrm{R}}$ achieves simple regret 
        \begin{align*}
            R_m(\pi_{\mathrm R}, p) = d \cdot \exp\big(- \Theta(m p^*) \big).
        \end{align*}
        This rate is minimax optimal w.r.t. $\Pi$ and $\mathcal P$.
    \end{itemize}
\end{theorem}
\noindent Despite its simplicity, our purely random PAC-MAP solver comes with powerful guarantees.
Not only is the procedure sound, but no competitor in $\Pi$ can reliably terminate faster or achieve better regret in the worst case. 
The following corollary is immediate.
\begin{corollary}\label{cor:tractability}
    $\pi_{\mathrm R}$ is tractable iff $h = \mathcal O(\log n)$.
\end{corollary}
\noindent In other words, a necessary and sufficient condition for tractable solutions in the purely random regime is that the min-entropy grows at a rate at most logarithmic in $n$. 
In the language of parameterized complexity theory \citep{downey2013}, we say that $\pi_\mathrm{R}$ is \textit{fixed-parameter tractable} w.r.t. $h$. Examples of distributions that satisfy this include the geometric, Poisson, and Zipf distributions with suitable parameters (Appx. A). Since $h$ cannot be computed directly without knowing the true $p^*$, we can only lower-bound it by sampling: a single non-uniform atom already falsifies the uniformity hypothesis, at which point $M$ shrinks quickly via line 14. Otherwise, stronger tools are needed.



\paragraph{Smooth PAC-MAP}
When $h$ is large, purely random sampling becomes inefficient because each draw only reveals the mass of its own atom. If nearby atoms tend to have similar probability, we can use the oracle to evaluate entire neighborhoods, providing more information per draw. This motivates the following smoothness assumption.

Let $\mathcal B_r(\bm q)$ be a Hamming ball of radius $r$ around point $\bm q$, with volume $v_r := \sum_{j=0}^r \binom{n}{j}$.  We say that pmf $p$ is \textit{modally Lipschitz} if, for all $\bm q \in \mathcal B_r(\bm q^*)$:
\begin{align*}
    \log_2 \frac{p^*}{p(\bm q \mid \bm e)} \leq L \cdot \lVert \bm q - \bm q^*\rVert_1,
\end{align*}
where $L>0$ is a constant. (Note that $L_1$ distance corresponds to Hamming distance on the Boolean hypercube.) 

For each $s \le r$, define the smoothness-weighted ball mass 
\begin{align*}
    w_s := \sum_{j=0}^s {n\choose j}2^{-Lj}.
\end{align*}
The modal Lipschitz condition implies that $\mathcal B_s(\bm q^*)$ has probability at least $p^* w_s$. Thus $w_s$ describes the effective size of the radius-$s$ near-modal region. For fixed tolerance $\varepsilon$, define
\begin{align*}
	k := \min \left\{ r,\left \lfloor \frac{\log_2(1/(1-\varepsilon))}{L} \right\rfloor \right\}.
\end{align*}
Then $\mathcal B_k(\bm q^*) \subseteq G_\varepsilon$, and this guaranteed $\varepsilon$-good region has mass at least $p^*w_k$.

We capitalize on this local structure by augmenting purely random search with an exploitation subroutine. 
At each step, the smooth PAC-MAP algorithm---denoted $\pi_{\mathrm S}$---either explores by drawing a fresh sample from $P(\bm Q \mid \bm e)$, or exploits by sweeping a Hamming ball of radius $r$ around the heaviest unswept candidate. 
Sweeps occur with frequency $\eta \asymp 1/v_r$. Hence, although each sweep costs $v_r$ oracle calls, the amortized sweep cost per random draw is constant. 
Sweeps can improve the incumbent solution but are not needed for the class-aware PAC certificate, whose accelerated stopping rule follows from the lower bound $\mu_\varepsilon \ge p^*w_k$. (For complete pseudocode, see Appx. C, Alg. 1.) 

For the regret lower bounds below, we also need to identify the regret scales at which the adversarial hidden-cap construction is valid. 
For $s<r$, define
\begin{align*}
	\beta_s := \frac{1-p^*w_s}{2^n-v_s}.
\end{align*}
We say that shell $s$ is \emph{feasible} if
\begin{align*}
	p^*2^{-L(s+1)} \le \beta_s < p^*2^{-Ls}.
\end{align*}
The lower inequality ensures that the flat background in the hidden-cap construction remains modally Lipschitz out to radius $r$, while the upper inequality ensures that points outside the cap are not $Ls$-good. When $v_s\ll 2^n$, this feasibility condition implies $Ls\lesssim d$. Thus the entropy gap $d$ determines how many regret shells admit matching minimax lower bounds.
Finally, let $\Delta_m^\pi := \log_2(p^*/\hat p_{m+1}^\pi)$ denote the loss of algorithm $\pi$ after $m$ calls to a $p$-oracle, such that $R_m(\pi, p) = \mathbb E_P[\Delta_m^\pi]$.
With these components in place, we may state the following result.

\begin{theorem}[Smooth PAC-MAP]\label{thm:smooth}
    Let $\Pi$ be the class of PAC-MAP solvers, with sound policy subset $\Pi^* \subset \Pi$.
    Let $\mathcal P$ be the class of pmfs on $\mathcal Q$ with min-entropy $h$ that are modally Lipschitz with parameters $L,r$.
    Assume that $v_r w_r = \mathcal O\big(\exp(h)\big)$.
    We have:
    \begin{itemize}[noitemsep]
        \item[(a)] $\pi_{\mathrm{S}} \in \Pi^*$.
        \item[(b)] Assume that $v_r \log \delta^{-1} = o\big(\exp(d) w_k \big)$. Then with probability at least $1 - \delta$, $\pi_{\mathrm{S}}$ terminates after $\mathcal O \big(\exp(h) w_k^{-1} \log \delta^{-1} \big)$ oracle calls. This rate is minimax optimal w.r.t. $\Pi^*$ and $\mathcal P$.
        \item[(c)] For some $c_1>0$, $\pi_{\mathrm S}$ achieves simple regret
        \begin{align*}
            R_m(\pi_{\mathrm S}, p) = \mathcal O\Big(L &\sum_{s=0}^{r-1} \exp(-c_1 m p^*w_s) +\\ 
            &(d - Lr)_+ \exp(-c_1 m p^* w_r)\Big).
        \end{align*}
        For each feasible shell $s < r$ satisfying $v_s=o(\exp(d)w_s)$, the corresponding tail rate is minimax optimal w.r.t. $\Pi$ and $\mathcal P$ up to constants in the exponent, i.e. there exist some $C_0, c_0 > 0$ such that:
        \begin{align*}
            \inf_{\pi \in \Pi} ~\sup_{P \in \mathcal P} ~\Pr_P \big( \Delta_m^\pi  > Ls \big) \geq C_0 \exp(- c_0 m p^* w_s).
        \end{align*}
    \end{itemize}
\end{theorem}
\noindent

\noindent The advantage of the modal Lipschitz assumption in the fixed-confidence setting is captured by the parameter $w_k$, which represents a speedup factor over the purely random policy of $\pi_\mathrm R$. This acceleration comes from the distributional geometry itself, not from Hamming sweeps per se. While the exploitation subroutine can improve incumbent solutions, the class-aware PAC certificate follows directly from the lower bound $\mu_\varepsilon \ge p^*w_k$. Notably, whereas Thm \ref{thm:pacmap_bin}(b) is a \textit{uniform} optimality claim that holds pointwise for all data streams, Thm \ref{thm:smooth}(b) is a \textit{minimax} result that holds in the worst case over $\mathcal P$. Pointwise optimality is unattainable once the competitor class admits adaptive algorithms, since the optimal explore/exploit balance is inevitably instance-dependent.

The regret guarantee of Thm \ref{thm:smooth}(c) is more nuanced than its purely random analogue Thm \ref{thm:pacmap_bin}(c). Rather than a single exponential rate, local smoothness induces a sequence of regret scales $Ls$, each associated with a weighted ball of mass $p^* w_s$. The regret of $\pi_{\mathrm S}$ is therefore controlled by the full near-modal profile $(w_s)_{s \le r}$. At each feasible shell, the corresponding tail rate is minimax optimal up to constants in the exponent; beyond the feasible range, the entropy gap $d$ limits how far hidden-cap lower bounds can be pushed. 

\begin{algorithm}[tb] \small
\caption{Budget PAC-MAP: $\pi_{\mathrm B}(P, p, M)$}
\label{alg:pacmap_binary_budget}
\textbf{Input}: Sampler $P$, oracle $p$, budget $M$\\
\textbf{Output}: Solution $\hat{\bm q}, \hat p$, PAC parameters $\bm \varepsilon, \delta(\hat p, \varepsilon)$\\
\begin{algorithmic}[1]
\STATE $S \gets \{\bm q_i\}_{i=1}^M \sim P(\bm Q \mid \bm e)$ \quad \texttt{// Draw samples}
\STATE $\hat{\bm{q}} \leftarrow \argmax_{\bm q \in S} ~p(\bm q \mid \bm{e})$ \quad \texttt{// Leading candidate}
\STATE $\hat p \gets p(\hat{\bm q} \mid \bm e)$ \quad \texttt{// Associated probability}
\STATE $\check p \gets 1 - \sum_{\bm q \in S} ~p(\bm q \mid \bm e)$ \texttt{// Residual mass}

\IF{$\hat p \geq \check p$}  
    \STATE $\bm \varepsilon \gets 0, \delta(\hat p, \varepsilon) \gets 0$ \quad \texttt{// Exact solution}
\ELSE
    \STATE $\bm \varepsilon \gets [0, 1 - \hat p)$ \quad \texttt{// Feasible range}
    \STATE $\delta(\hat p, \varepsilon) \gets \big( 1 - \hat p / (1 - \varepsilon) \big)^M$ \quad \texttt{// Pareto front}
\ENDIF

\RETURN $\hat{\bm q}, ~\hat p, ~\bm \varepsilon, ~\delta(\hat p, \varepsilon)$
\end{algorithmic}
\end{algorithm}

\paragraph{Budget PAC-MAP}
For high-dimensional, high-entropy distributions, even optimal rates may prove too onerous for fixed-confidence BAI. 
A practical alternative in such cases is to fix the sample budget instead. 
Observe that, given any plug-in estimate $\hat p$, there is a deterministic relationship between PAC parameters $\varepsilon, \delta$ and the number of samples $M$ required for a corresponding PAC certificate. 
Inverting our previous bounds and jointly optimizing for error tolerance and confidence level, we may construct a curve describing the tradeoff between the two for a given dataset. 
The solution $\hat{\bm q}, \hat p$ is PAC-certified at every point along the curve, providing a range of guarantees that practitioners can report or attempt to improve upon by increasing $M$. 

Let $\preceq$ be a partial ordering on PAC parameter pairs.
We say that one pair \textit{dominates} another if the former is strictly better in at least one respect (improved accuracy or confidence) and no worse in either respect. 
Formally, we write:
\begin{align*}
    (\varepsilon, \delta) \prec (\varepsilon', \delta') \Leftrightarrow (\varepsilon < \varepsilon' \lor \delta < \delta') \land (\varepsilon \leq \varepsilon' \land \delta \leq \delta').
\end{align*}
We say that a PAC pair is \textit{admissible} if it is not dominated by any other pair. The set of admissible pairs forms a Pareto frontier of PAC parameters that are optimal w.r.t. the sampled data, insomuch as no valid PAC certificate can be issued with greater accuracy without reducing confidence or vice versa.

Consider the purely random fixed-budget PAC-MAP policy $\pi_{\mathrm B}$ (see Alg. \ref{alg:pacmap_binary_budget}). The procedure works as follows. First, draw sample $S = \{\bm q_i\}_{i=1}^M$  from $P(\bm Q \mid \bm e)$ and record the maximal observed probability $\hat p$, as well as the residual mass $\check p$. These represent anytime-valid bounds on $p^*$ (provided they do not cross, in which case $\hat{\bm q}$ is an exact solution). Thus we can always get a cheap PAC certificate with $\varepsilon = \max\{0, 1 - \hat p / \check p\}, \delta=0$. 

Now suppose that we want to tighten $\varepsilon$ by introducing nonzero failure probability $\delta$. 
Our equations imply a feasible interval of $[0, 1 - \hat p)$ for $\varepsilon$ (see Appx. A for details). 
Plugging in values in this range and rearranging our worst-case stopping criterion to solve for $\delta$, we get minimal corresponding failure probabilities via $\big(1 - \hat p / (1 - \varepsilon) \big)^M$.
We use boldface notation for $\bm \varepsilon$ in the pseudocode for $\pi_{\mathrm B}$ to indicate that it is a set (albeit a singleton if $\hat p \geq \check p$), while $\delta$ is now a function of $\hat p, \varepsilon$. Resulting PAC parameter pairs describe a range of admissible solutions, where tightness and coverage trade off as expected. 
Each point along the curve is $S$-optimal, in the sense that stronger PAC guarantees are not possible without gathering more data. We summarize these points in the following theorem.

\begin{theorem}[Budget PAC-MAP]\label{thm:budget_pacmap_binary}
    $\pi_{\mathrm B}$ is sound and complete w.r.t. PAC parameters, returning all and only the admissible $(\varepsilon, \delta)$ pairs for a given sample $S$ and corresponding MAP estimate $\hat{\bm q}, \hat p$. The procedure executes in time $\mathcal O(M)$.
\end{theorem}
\noindent Unlike Thms \ref{thm:pacmap_bin} and \ref{thm:smooth}, whose stopping times depend on $h$, $\pi_\mathrm B$ executes in time linear in $M$ regardless of the distribution's min-entropy. The cost of this independence is potentially large PAC parameters, which improve with sample budget and are Pareto-optimal given the data.






\begin{table*}
\centering
\scriptsize
\definecolor{npgRed}{HTML}{E64B35}
\definecolor{npgBlue}{HTML}{4DBBD5}
\definecolor{npgGreen}{HTML}{00A087}
\definecolor{npgNavy}{HTML}{3C5488}
\definecolor{npgPurple}{HTML}{8491B4}
\definecolor{npgOrange}{HTML}{F39B7F}
\definecolor{npgBrown}{HTML}{7E6148}
\setlength{\tabcolsep}{1mm}

\begin{tabular}{l| c @{\hspace{0.2em}}| c @{\hspace{0.2em}}| c @{\hspace{0.2em}}| c}
\toprule
Dataset - Dimension & 20\% Q, 80\% E & 50\% Q, 50\% E & 20\% Q, 50\% E, 30\% V & 50\% Q, 30\% E, 20\% V \\
\midrule
\texttt{accidents - 111} & \mnum{$\textcolor{npgPurple}{1.8}$}\msep\mnum{$\textcolor{npgBlue}{\mathbf{1.4}}$}\msep\mnum{$\textcolor{npgRed}{\mathbf{1.4}}$}\msep\mnum{$\textcolor{npgBrown}{7.0}$}\msep\mnum{$\textcolor{npgOrange}{4.0}$}\msep\mnum{$\textcolor{npgGreen}{\mathbf{1.4}}$}\msep\mnum{$\textcolor{npgNavy}{\mathbf{1.4}}$}& \mnum{$\textcolor{npgPurple}{3.4}$}\msep\mnum{$\textcolor{npgBlue}{1.7}$}\msep\mnum{$\textcolor{npgRed}{1.4}$}\msep\mnum{$\textcolor{npgBrown}{7.0}$}\msep\mnum{$\textcolor{npgOrange}{3.9}$}\msep\mnum{$\textcolor{npgGreen}{2.3}$}\msep\mnum{$\textcolor{npgNavy}{\mathbf{1.3}}$} & \mnum{$\textcolor{npgPurple}{1.8}$}\msep\mnum{$\textcolor{npgBlue}{2.6}$}\msep\mnum{$\textcolor{npgRed}{1.8}$}\msep\mnum{$\textcolor{npgBrown}{6.0}$}\msep\mnum{$\textcolor{npgGreen}{\mathbf{1.0}}$}\msep\mnum{$\textcolor{npgNavy}{\mathbf{1.0}}$} & \mnum{$\textcolor{npgPurple}{5.0}$}\msep\mnum{$\textcolor{npgBlue}{2.8}$}\msep\mnum{$\textcolor{npgRed}{1.2}$}\msep\mnum{$\textcolor{npgBrown}{6.0}$}\msep\mnum{$\textcolor{npgGreen}{1.2}$}\msep\mnum{$\textcolor{npgNavy}{\mathbf{1.0}}$} \\
\texttt{adult - 123} & \mnum{$\textcolor{npgPurple}{1.5}$}\msep\mnum{$\textcolor{npgBlue}{1.6}$}\msep\mnum{$\textcolor{npgRed}{\mathbf{1.1}}$}\msep\mnum{$\textcolor{npgBrown}{7.0}$}\msep\mnum{$\textcolor{npgOrange}{5.4}$}\msep\mnum{$\textcolor{npgGreen}{\mathbf{1.1}}$}\msep\mnum{$\textcolor{npgNavy}{\mathbf{1.1}}$} & \mnum{$\textcolor{npgPurple}{3.4}$}\msep\mnum{$\textcolor{npgBlue}{2.3}$}\msep\mnum{$\textcolor{npgRed}{\mathbf{1.0}}$}\msep\mnum{$\textcolor{npgBrown}{7.0}$}\msep\mnum{$\textcolor{npgOrange}{5.9}$}\msep\mnum{$\textcolor{npgGreen}{\mathbf{1.0}}$}\msep\mnum{$\textcolor{npgNavy}{\mathbf{1.0}}$} & \mnum{$\textcolor{npgPurple}{1.4}$}\msep\mnum{$\textcolor{npgBlue}{1.8}$}\msep\mnum{$\textcolor{npgRed}{\mathbf{1.0}}$}\msep\mnum{$\textcolor{npgBrown}{4.9}$}\msep\mnum{$\textcolor{npgGreen}{\mathbf{1.0}}$}\msep\mnum{$\textcolor{npgNavy}{\mathbf{1.0}}$} & \mnum{$\textcolor{npgPurple}{3.0}$}\msep\mnum{$\textcolor{npgBlue}{2.2}$}\msep\mnum{$\textcolor{npgRed}{\mathbf{1.0}}$}\msep\mnum{$\textcolor{npgBrown}{6.0}$}\msep\mnum{$\textcolor{npgGreen}{\mathbf{1.0}}$}\msep\mnum{$\textcolor{npgNavy}{\mathbf{1.0}}$} \\
\texttt{baudio - 100} & \mnum{$\textcolor{npgPurple}{3.5}$}\msep\mnum{$\textcolor{npgBlue}{2.6}$}\msep\mnum{$\textcolor{npgRed}{\mathbf{1.0}}$}\msep\mnum{$\textcolor{npgBrown}{6.8}$}\msep\mnum{$\textcolor{npgOrange}{5.8}$}\msep\mnum{$\textcolor{npgGreen}{\mathbf{1.0}}$}\msep\mnum{$\textcolor{npgNavy}{\mathbf{1.0}}$} & \mnum{$\textcolor{npgPurple}{5.9}$}\msep\mnum{$\textcolor{npgBlue}{1.6}$}\msep\mnum{$\textcolor{npgRed}{\mathbf{1.0}}$}\msep\mnum{$\textcolor{npgBrown}{6.1}$}\msep\mnum{$\textcolor{npgOrange}{5.9}$}\msep\mnum{$\textcolor{npgGreen}{3.3}$}\msep\mnum{$\textcolor{npgNavy}{2.1}$} & \mnum{$\textcolor{npgPurple}{3.8}$}\msep\mnum{$\textcolor{npgBlue}{4.1}$}\msep\mnum{$\textcolor{npgRed}{1.2}$}\msep\mnum{$\textcolor{npgBrown}{5.7}$}\msep\mnum{$\textcolor{npgGreen}{\mathbf{1.0}}$}\msep\mnum{$\textcolor{npgNavy}{\mathbf{1.0}}$} & \mnum{$\textcolor{npgPurple}{5.7}$}\msep\mnum{$\textcolor{npgBlue}{3.0}$}\msep\mnum{$\textcolor{npgRed}{1.1}$}\msep\mnum{$\textcolor{npgBrown}{4.5}$}\msep\mnum{$\textcolor{npgGreen}{3.1}$}\msep\mnum{$\textcolor{npgNavy}{\mathbf{1.0}}$} \\
\texttt{bnetflix - 100} & \mnum{$\textcolor{npgPurple}{4.3}$}\msep\mnum{$\textcolor{npgBlue}{2.0}$}\msep\mnum{$\textcolor{npgRed}{1.3}$}\msep\mnum{$\textcolor{npgBrown}{6.9}$}\msep\mnum{$\textcolor{npgOrange}{5.6}$}\msep\mnum{$\textcolor{npgGreen}{\mathbf{1.1}}$}\msep\mnum{$\textcolor{npgNavy}{\mathbf{1.1}}$} & \mnum{$\textcolor{npgPurple}{5.2}$}\msep\mnum{$\textcolor{npgBlue}{1.7}$}\msep\mnum{$\textcolor{npgRed}{\mathbf{1.2}}$}\msep\mnum{$\textcolor{npgBrown}{6.8}$}\msep\mnum{$\textcolor{npgOrange}{5.6}$}\msep\mnum{$\textcolor{npgGreen}{4.1}$}\msep\mnum{$\textcolor{npgNavy}{1.7}$} & \mnum{$\textcolor{npgPurple}{4.8}$}\msep\mnum{$\textcolor{npgBlue}{2.3}$}\msep\mnum{$\textcolor{npgRed}{1.5}$}\msep\mnum{$\textcolor{npgBrown}{5.9}$}\msep\mnum{$\textcolor{npgGreen}{\mathbf{1.0}}$}\msep\mnum{$\textcolor{npgNavy}{\mathbf{1.0}}$} & \mnum{$\textcolor{npgPurple}{5.0}$}\msep\mnum{$\textcolor{npgBlue}{1.8}$}\msep\mnum{$\textcolor{npgRed}{\mathbf{1.0}}$}\msep\mnum{$\textcolor{npgBrown}{5.9}$}\msep\mnum{$\textcolor{npgGreen}{3.2}$}\msep\mnum{$\textcolor{npgNavy}{1.6}$} \\
\texttt{book - 500} & \mnum{$\textcolor{npgPurple}{\mathbf{1.1}}$}\msep\mnum{$\textcolor{npgBlue}{\mathbf{1.1}}$}\msep\mnum{$\textcolor{npgRed}{\mathbf{1.1}}$}\msep\mnum{$\textcolor{npgBrown}{6.1}$}\msep\mnum{$\textcolor{npgOrange}{4.7}$}\msep\mnum{$\textcolor{npgGreen}{6.8}$}\msep\mnum{$\textcolor{npgNavy}{\mathbf{1.1}}$} & \mnum{$\textcolor{npgPurple}{4.4}$}\msep\mnum{$\textcolor{npgBlue}{1.1}$}\msep\mnum{$\textcolor{npgRed}{\mathbf{1.0}}$}\msep\mnum{$\textcolor{npgBrown}{5.8}$}\msep\mnum{$\textcolor{npgOrange}{3.8}$}\msep\mnum{$\textcolor{npgGreen}{7.0}$}\msep\mnum{$\textcolor{npgNavy}{3.1}$} & \mnum{$\textcolor{npgPurple}{1.5}$}\msep\mnum{$\textcolor{npgBlue}{3.3}$}\msep\mnum{$\textcolor{npgRed}{\mathbf{1.1}}$}\msep\mnum{$\textcolor{npgBrown}{5.1}$}\msep\mnum{$\textcolor{npgGreen}{5.5}$}\msep\mnum{$\textcolor{npgNavy}{\mathbf{1.1}}$} & \mnum{$\textcolor{npgPurple}{3.6}$}\msep\mnum{$\textcolor{npgBlue}{1.9}$}\msep\mnum{$\textcolor{npgRed}{\mathbf{1.0}}$}\msep\mnum{$\textcolor{npgBrown}{4.5}$}\msep\mnum{$\textcolor{npgGreen}{6.0}$}\msep\mnum{$\textcolor{npgNavy}{2.3}$} \\
\texttt{connect4 - 126} & \mnum{$\textcolor{npgPurple}{2.9}$}\msep\mnum{$\textcolor{npgBlue}{2.2}$}\msep\mnum{$\textcolor{npgRed}{1.5}$}\msep\mnum{$\textcolor{npgBrown}{7.0}$}\msep\mnum{$\textcolor{npgOrange}{4.8}$}\msep\mnum{$\textcolor{npgGreen}{\mathbf{1.2}}$}\msep\mnum{$\textcolor{npgNavy}{\mathbf{1.2}}$} & \mnum{$\textcolor{npgPurple}{4.4}$}\msep\mnum{$\textcolor{npgBlue}{2.0}$}\msep\mnum{$\textcolor{npgRed}{\mathbf{1.0}}$}\msep\mnum{$\textcolor{npgBrown}{7.0}$}\msep\mnum{$\textcolor{npgOrange}{6.0}$}\msep\mnum{$\textcolor{npgGreen}{\mathbf{1.0}}$}\msep\mnum{$\textcolor{npgNavy}{\mathbf{1.0}}$} & \mnum{$\textcolor{npgPurple}{2.2}$}\msep\mnum{$\textcolor{npgBlue}{1.8}$}\msep\mnum{$\textcolor{npgRed}{\mathbf{1.0}}$}\msep\mnum{$\textcolor{npgBrown}{6.0}$}\msep\mnum{$\textcolor{npgGreen}{\mathbf{1.0}}$}\msep\mnum{$\textcolor{npgNavy}{\mathbf{1.0}}$} & \mnum{$\textcolor{npgPurple}{3.7}$}\msep\mnum{$\textcolor{npgBlue}{1.5}$}\msep\mnum{$\textcolor{npgRed}{1.2}$}\msep\mnum{$\textcolor{npgBrown}{6.0}$}\msep\mnum{$\textcolor{npgGreen}{\mathbf{1.0}}$}\msep\mnum{$\textcolor{npgNavy}{\mathbf{1.0}}$} \\
\texttt{dna - 180} & \mnum{$\textcolor{npgPurple}{3.4}$}\msep\mnum{$\textcolor{npgBlue}{1.9}$}\msep\mnum{$\textcolor{npgRed}{\mathbf{1.1}}$}\msep\mnum{$\textcolor{npgBrown}{6.6}$}\msep\mnum{$\textcolor{npgOrange}{4.9}$}\msep\mnum{$\textcolor{npgGreen}{2.6}$}\msep\mnum{$\textcolor{npgNavy}{1.6}$} & \mnum{$\textcolor{npgPurple}{4.6}$}\msep\mnum{$\textcolor{npgBlue}{1.4}$}\msep\mnum{$\textcolor{npgRed}{\mathbf{1.0}}$}\msep\mnum{$\textcolor{npgBrown}{6.4}$}\msep\mnum{$\textcolor{npgOrange}{5.1}$}\msep\mnum{$\textcolor{npgGreen}{5.5}$}\msep\mnum{$\textcolor{npgNavy}{2.0}$} & \mnum{$\textcolor{npgPurple}{3.2}$}\msep\mnum{$\textcolor{npgBlue}{2.3}$}\msep\mnum{$\textcolor{npgRed}{\mathbf{1.0}}$}\msep\mnum{$\textcolor{npgBrown}{5.8}$}\msep\mnum{$\textcolor{npgGreen}{1.5}$}\msep\mnum{$\textcolor{npgNavy}{\mathbf{1.0}}$} & \mnum{$\textcolor{npgPurple}{4.8}$}\msep\mnum{$\textcolor{npgBlue}{2.6}$}\msep\mnum{$\textcolor{npgRed}{\mathbf{1.0}}$}\msep\mnum{$\textcolor{npgBrown}{4.7}$}\msep\mnum{$\textcolor{npgGreen}{5.2}$}\msep\mnum{$\textcolor{npgNavy}{2.4}$} \\
\texttt{jester - 100} & \mnum{$\textcolor{npgPurple}{4.8}$}\msep\mnum{$\textcolor{npgBlue}{2.3}$}\msep\mnum{$\textcolor{npgRed}{1.2}$}\msep\mnum{$\textcolor{npgBrown}{6.8}$}\msep\mnum{$\textcolor{npgOrange}{5.8}$}\msep\mnum{$\textcolor{npgGreen}{\mathbf{1.0}}$}\msep\mnum{$\textcolor{npgNavy}{\mathbf{1.0}}$} & \mnum{$\textcolor{npgPurple}{5.5}$}\msep\mnum{$\textcolor{npgBlue}{1.3}$}\msep\mnum{$\textcolor{npgRed}{\mathbf{1.0}}$}\msep\mnum{$\textcolor{npgBrown}{6.6}$}\msep\mnum{$\textcolor{npgOrange}{5.7}$}\msep\mnum{$\textcolor{npgGreen}{4.1}$}\msep\mnum{$\textcolor{npgNavy}{1.5}$} & \mnum{$\textcolor{npgPurple}{3.8}$}\msep\mnum{$\textcolor{npgBlue}{3.3}$}\msep\mnum{$\textcolor{npgRed}{\mathbf{1.0}}$}\msep\mnum{$\textcolor{npgBrown}{5.9}$}\msep\mnum{$\textcolor{npgGreen}{\mathbf{1.0}}$}\msep\mnum{$\textcolor{npgNavy}{\mathbf{1.0}}$} & \mnum{$\textcolor{npgPurple}{5.3}$}\msep\mnum{$\textcolor{npgBlue}{2.1}$}\msep\mnum{$\textcolor{npgRed}{\mathbf{1.2}}$}\msep\mnum{$\textcolor{npgBrown}{5.5}$}\msep\mnum{$\textcolor{npgGreen}{3.6}$}\msep\mnum{$\textcolor{npgNavy}{2.1}$} \\
\texttt{kdd - 64} & \mnum{$\textcolor{npgPurple}{2.0}$}\msep\mnum{$\textcolor{npgBlue}{1.6}$}\msep\mnum{$\textcolor{npgRed}{\mathbf{1.1}}$}\msep\mnum{$\textcolor{npgBrown}{6.3}$}\msep\mnum{$\textcolor{npgOrange}{5.5}$}\msep\mnum{$\textcolor{npgGreen}{\mathbf{1.1}}$}\msep\mnum{$\textcolor{npgNavy}{\mathbf{1.1}}$} & \mnum{$\textcolor{npgPurple}{2.3}$}\msep\mnum{$\textcolor{npgBlue}{1.5}$}\msep\mnum{$\textcolor{npgRed}{\mathbf{1.0}}$}\msep\mnum{$\textcolor{npgBrown}{7.0}$}\msep\mnum{$\textcolor{npgOrange}{5.8}$}\msep\mnum{$\textcolor{npgGreen}{\mathbf{1.0}}$}\msep\mnum{$\textcolor{npgNavy}{\mathbf{1.0}}$} & \mnum{$\textcolor{npgPurple}{\mathbf{1.0}}$}\msep\mnum{$\textcolor{npgBlue}{\mathbf{1.0}}$}\msep\mnum{$\textcolor{npgRed}{\mathbf{1.0}}$}\msep\mnum{$\textcolor{npgBrown}{6.0}$}\msep\mnum{$\textcolor{npgGreen}{\mathbf{1.0}}$}\msep\mnum{$\textcolor{npgNavy}{\mathbf{1.0}}$} & \mnum{$\textcolor{npgPurple}{2.2}$}\msep\mnum{$\textcolor{npgBlue}{1.3}$}\msep\mnum{$\textcolor{npgRed}{\mathbf{1.0}}$}\msep\mnum{$\textcolor{npgBrown}{5.5}$}\msep\mnum{$\textcolor{npgGreen}{\mathbf{1.0}}$}\msep\mnum{$\textcolor{npgNavy}{\mathbf{1.0}}$} \\
\texttt{kosarek - 190} & \mnum{$\textcolor{npgPurple}{\mathbf{1.1}}$}\msep\mnum{$\textcolor{npgBlue}{1.3}$}\msep\mnum{$\textcolor{npgRed}{\mathbf{1.1}}$}\msep\mnum{$\textcolor{npgBrown}{7.0}$}\msep\mnum{$\textcolor{npgOrange}{5.4}$}\msep\mnum{$\textcolor{npgGreen}{4.0}$}\msep\mnum{$\textcolor{npgNavy}{1.3}$} & \mnum{$\textcolor{npgPurple}{4.4}$}\msep\mnum{$\textcolor{npgBlue}{1.7}$}\msep\mnum{$\textcolor{npgRed}{\mathbf{1.4}}$}\msep\mnum{$\textcolor{npgBrown}{7.0}$}\msep\mnum{$\textcolor{npgOrange}{2.9}$}\msep\mnum{$\textcolor{npgGreen}{5.7}$}\msep\mnum{$\textcolor{npgNavy}{2.5}$} & \mnum{$\textcolor{npgPurple}{1.9}$}\msep\mnum{$\textcolor{npgBlue}{1.4}$}\msep\mnum{$\textcolor{npgRed}{\mathbf{1.0}}$}\msep\mnum{$\textcolor{npgBrown}{5.9}$}\msep\mnum{$\textcolor{npgGreen}{3.2}$}\msep\mnum{$\textcolor{npgNavy}{\mathbf{1.0}}$} & \mnum{$\textcolor{npgPurple}{4.1}$}\msep\mnum{$\textcolor{npgBlue}{2.5}$}\msep\mnum{$\textcolor{npgRed}{\mathbf{1.0}}$}\msep\mnum{$\textcolor{npgBrown}{5.9}$}\msep\mnum{$\textcolor{npgGreen}{4.4}$}\msep\mnum{$\textcolor{npgNavy}{1.9}$} \\
\texttt{msnbc - 17} & \mnum{$\textcolor{npgPurple}{1.8}$}\msep\mnum{$\textcolor{npgBlue}{1.8}$}\msep\mnum{$\textcolor{npgRed}{\mathbf{1.4}}$}\msep\mnum{$\textcolor{npgBrown}{4.4}$}\msep\mnum{$\textcolor{npgOrange}{3.5}$}\msep\mnum{$\textcolor{npgGreen}{\mathbf{1.4}}$}\msep\mnum{$\textcolor{npgNavy}{\mathbf{1.4}}$} & \mnum{$\textcolor{npgPurple}{3.7}$}\msep\mnum{$\textcolor{npgBlue}{1.5}$}\msep\mnum{$\textcolor{npgRed}{\mathbf{1.1}}$}\msep\mnum{$\textcolor{npgBrown}{5.6}$}\msep\mnum{$\textcolor{npgOrange}{5.6}$}\msep\mnum{$\textcolor{npgGreen}{\mathbf{1.1}}$}\msep\mnum{$\textcolor{npgNavy}{\mathbf{1.1}}$} & \mnum{$\textcolor{npgPurple}{1.9}$}\msep\mnum{$\textcolor{npgBlue}{1.4}$}\msep\mnum{$\textcolor{npgRed}{\mathbf{1.0}}$}\msep\mnum{$\textcolor{npgBrown}{2.7}$}\msep\mnum{$\textcolor{npgGreen}{\mathbf{1.0}}$}\msep\mnum{$\textcolor{npgNavy}{\mathbf{1.0}}$} & \mnum{$\textcolor{npgPurple}{1.4}$}\msep\mnum{$\textcolor{npgBlue}{1.3}$}\msep\mnum{$\textcolor{npgRed}{\mathbf{1.0}}$}\msep\mnum{$\textcolor{npgBrown}{2.5}$}\msep\mnum{$\textcolor{npgGreen}{\mathbf{1.0}}$}\msep\mnum{$\textcolor{npgNavy}{\mathbf{1.0}}$} \\
\texttt{msweb - 294} & \mnum{$\textcolor{npgPurple}{1.8}$}\msep\mnum{$\textcolor{npgBlue}{\mathbf{1.3}}$}\msep\mnum{$\textcolor{npgRed}{\mathbf{1.3}}$}\msep\mnum{$\textcolor{npgBrown}{6.1}$}\msep\mnum{$\textcolor{npgOrange}{4.3}$}\msep\mnum{$\textcolor{npgGreen}{4.4}$}\msep\mnum{$\textcolor{npgNavy}{\mathbf{1.3}}$} & \mnum{$\textcolor{npgPurple}{4.2}$}\msep\mnum{$\textcolor{npgBlue}{\mathbf{1.3}}$}\msep\mnum{$\textcolor{npgRed}{\mathbf{1.3}}$}\msep\mnum{$\textcolor{npgBrown}{5.6}$}\msep\mnum{$\textcolor{npgOrange}{3.1}$}\msep\mnum{$\textcolor{npgGreen}{6.9}$}\msep\mnum{$\textcolor{npgNavy}{1.6}$} & \mnum{$\textcolor{npgPurple}{1.5}$}\msep\mnum{$\textcolor{npgBlue}{2.1}$}\msep\mnum{$\textcolor{npgRed}{\mathbf{1.0}}$}\msep\mnum{$\textcolor{npgBrown}{4.7}$}\msep\mnum{$\textcolor{npgGreen}{2.6}$}\msep\mnum{$\textcolor{npgNavy}{\mathbf{1.0}}$} & \mnum{$\textcolor{npgPurple}{3.2}$}\msep\mnum{$\textcolor{npgBlue}{1.1}$}\msep\mnum{$\textcolor{npgRed}{\mathbf{1.0}}$}\msep\mnum{$\textcolor{npgBrown}{4.6}$}\msep\mnum{$\textcolor{npgGreen}{6.0}$}\msep\mnum{$\textcolor{npgNavy}{1.1}$} \\
\texttt{mushrooms - 112} & \mnum{$\textcolor{npgPurple}{1.9}$}\msep\mnum{$\textcolor{npgBlue}{\mathbf{1.1}}$}\msep\mnum{$\textcolor{npgRed}{\mathbf{1.1}}$}\msep\mnum{$\textcolor{npgBrown}{7.0}$}\msep\mnum{$\textcolor{npgOrange}{5.0}$}\msep\mnum{$\textcolor{npgGreen}{\mathbf{1.1}}$}\msep\mnum{$\textcolor{npgNavy}{\mathbf{1.1}}$}& \mnum{$\textcolor{npgPurple}{2.7}$}\msep\mnum{$\textcolor{npgBlue}{1.5}$}\msep\mnum{$\textcolor{npgRed}{\mathbf{1.1}}$}\msep\mnum{$\textcolor{npgBrown}{7.0}$}\msep\mnum{$\textcolor{npgOrange}{5.5}$}\msep\mnum{$\textcolor{npgGreen}{\mathbf{1.1}}$}\msep\mnum{$\textcolor{npgNavy}{\mathbf{1.1}}$} & \mnum{$\textcolor{npgPurple}{2.2}$}\msep\mnum{$\textcolor{npgBlue}{2.0}$}\msep\mnum{$\textcolor{npgRed}{\mathbf{1.0}}$}\msep\mnum{$\textcolor{npgBrown}{6.0}$}\msep\mnum{$\textcolor{npgGreen}{\mathbf{1.0}}$}\msep\mnum{$\textcolor{npgNavy}{\mathbf{1.0}}$} & \mnum{$\textcolor{npgPurple}{3.4}$}\msep\mnum{$\textcolor{npgBlue}{1.6}$}\msep\mnum{$\textcolor{npgRed}{1.2}$}\msep\mnum{$\textcolor{npgBrown}{6.0}$}\msep\mnum{$\textcolor{npgGreen}{\mathbf{1.0}}$}\msep\mnum{$\textcolor{npgNavy}{\mathbf{1.0}}$} \\
\texttt{nips - 500} & \mnum{$\textcolor{npgPurple}{2.9}$}\msep\mnum{$\textcolor{npgBlue}{1.9}$}\msep\mnum{$\textcolor{npgRed}{\mathbf{1.1}}$}\msep\mnum{$\textcolor{npgBrown}{7.0}$}\msep\mnum{$\textcolor{npgOrange}{5.2}$}\msep\mnum{$\textcolor{npgGreen}{5.7}$}\msep\mnum{$\textcolor{npgNavy}{1.4}$} & \mnum{$\textcolor{npgPurple}{4.0}$}\msep\mnum{$\textcolor{npgBlue}{1.5}$}\msep\mnum{$\textcolor{npgRed}{\mathbf{1.0}}$}\msep\mnum{$\textcolor{npgBrown}{6.9}$}\msep\mnum{$\textcolor{npgOrange}{5.0}$}\msep\mnum{$\textcolor{npgGreen}{5.9}$}\msep\mnum{$\textcolor{npgNavy}{2.9}$} & \mnum{$\textcolor{npgPurple}{3.3}$}\msep\mnum{$\textcolor{npgBlue}{2.6}$}\msep\mnum{$\textcolor{npgRed}{1.3}$}\msep\mnum{$\textcolor{npgBrown}{6.0}$}\msep\mnum{$\textcolor{npgGreen}{5.0}$}\msep\mnum{$\textcolor{npgNavy}{\mathbf{1.1}}$} & \mnum{$\textcolor{npgPurple}{3.7}$}\msep\mnum{$\textcolor{npgBlue}{2.4}$}\msep\mnum{$\textcolor{npgRed}{\mathbf{1.1}}$}\msep\mnum{$\textcolor{npgBrown}{6.0}$}\msep\mnum{$\textcolor{npgGreen}{5.0}$}\msep\mnum{$\textcolor{npgNavy}{2.6}$} \\
\texttt{nltcs - 16} & \mnum{$\textcolor{npgPurple}{1.7}$}\msep\mnum{$\textcolor{npgBlue}{\mathbf{1.1}}$}\msep\mnum{$\textcolor{npgRed}{\mathbf{1.1}}$}\msep\mnum{$\textcolor{npgBrown}{4.4}$}\msep\mnum{$\textcolor{npgOrange}{4.3}$}\msep\mnum{$\textcolor{npgGreen}{\mathbf{1.1}}$}\msep\mnum{$\textcolor{npgNavy}{\mathbf{1.1}}$} & \mnum{$\textcolor{npgPurple}{2.7}$}\msep\mnum{$\textcolor{npgBlue}{\mathbf{1.1}}$}\msep\mnum{$\textcolor{npgRed}{\mathbf{1.1}}$}\msep\mnum{$\textcolor{npgBrown}{5.7}$}\msep\mnum{$\textcolor{npgOrange}{5.0}$}\msep\mnum{$\textcolor{npgGreen}{\mathbf{1.1}}$}\msep\mnum{$\textcolor{npgNavy}{\mathbf{1.1}}$} & \mnum{$\textcolor{npgPurple}{1.7}$}\msep\mnum{$\textcolor{npgBlue}{2.3}$}\msep\mnum{$\textcolor{npgRed}{\mathbf{1.0}}$}\msep\mnum{$\textcolor{npgBrown}{3.5}$}\msep\mnum{$\textcolor{npgGreen}{\mathbf{1.0}}$}\msep\mnum{$\textcolor{npgNavy}{\mathbf{1.0}}$} & \mnum{$\textcolor{npgPurple}{2.4}$}\msep\mnum{$\textcolor{npgBlue}{1.6}$}\msep\mnum{$\textcolor{npgRed}{1.2}$}\msep\mnum{$\textcolor{npgBrown}{4.3}$}\msep\mnum{$\textcolor{npgGreen}{\mathbf{1.0}}$}\msep\mnum{$\textcolor{npgNavy}{\mathbf{1.0}}$} \\
\texttt{ocr\_letters - 128} & \mnum{$\textcolor{npgPurple}{3.4}$}\msep\mnum{$\textcolor{npgBlue}{1.9}$}\msep\mnum{$\textcolor{npgRed}{1.9}$}\msep\mnum{$\textcolor{npgBrown}{--}$}\msep\mnum{$\textcolor{npgOrange}{--}$}\msep\mnum{$\textcolor{npgGreen}{\mathbf{1.0}}$}\msep\mnum{$\textcolor{npgNavy}{\mathbf{1.0}}$} & \mnum{$\textcolor{npgPurple}{4.2}$}\msep\mnum{$\textcolor{npgBlue}{1.8}$}\msep\mnum{$\textcolor{npgRed}{\mathbf{1.6}}$}\msep\mnum{$\textcolor{npgBrown}{--}$}\msep\mnum{$\textcolor{npgOrange}{5.9}$}\msep\mnum{$\textcolor{npgGreen}{3.8}$}\msep\mnum{$\textcolor{npgNavy}{2.3}$} & \mnum{$\textcolor{npgPurple}{4.3}$}\msep\mnum{$\textcolor{npgBlue}{4.5}$}\msep\mnum{$\textcolor{npgRed}{1.5}$}\msep\mnum{$\textcolor{npgBrown}{--}$}\msep\mnum{$\textcolor{npgGreen}{\mathbf{1.0}}$}\msep\mnum{$\textcolor{npgNavy}{\mathbf{1.0}}$} & \mnum{$\textcolor{npgPurple}{4.9}$}\msep\mnum{$\textcolor{npgBlue}{1.7}$}\msep\mnum{$\textcolor{npgRed}{\mathbf{1.0}}$}\msep\mnum{$\textcolor{npgBrown}{--}$}\msep\mnum{$\textcolor{npgGreen}{4.0}$}\msep\mnum{$\textcolor{npgNavy}{2.3}$} \\
\texttt{plants - 69} & \mnum{$\textcolor{npgPurple}{1.4}$}\msep\mnum{$\textcolor{npgBlue}{\mathbf{1.0}}$}\msep\mnum{$\textcolor{npgRed}{\mathbf{1.0}}$}\msep\mnum{$\textcolor{npgBrown}{7.0}$}\msep\mnum{$\textcolor{npgOrange}{5.5}$}\msep\mnum{$\textcolor{npgGreen}{\mathbf{1.0}}$}\msep\mnum{$\textcolor{npgNavy}{\mathbf{1.0}}$} & \mnum{$\textcolor{npgPurple}{4.8}$}\msep\mnum{$\textcolor{npgBlue}{1.7}$}\msep\mnum{$\textcolor{npgRed}{1.4}$}\msep\mnum{$\textcolor{npgBrown}{6.8}$}\msep\mnum{$\textcolor{npgOrange}{6.0}$}\msep\mnum{$\textcolor{npgGreen}{\mathbf{1.0}}$}\msep\mnum{$\textcolor{npgNavy}{\mathbf{1.0}}$} & \mnum{$\textcolor{npgPurple}{3.7}$}\msep\mnum{$\textcolor{npgBlue}{2.2}$}\msep\mnum{$\textcolor{npgRed}{1.3}$}\msep\mnum{$\textcolor{npgBrown}{5.9}$}\msep\mnum{$\textcolor{npgGreen}{\mathbf{1.0}}$}\msep\mnum{$\textcolor{npgNavy}{\mathbf{1.0}}$} & \mnum{$\textcolor{npgPurple}{4.6}$}\msep\mnum{$\textcolor{npgBlue}{3.0}$}\msep\mnum{$\textcolor{npgRed}{1.5}$}\msep\mnum{$\textcolor{npgBrown}{5.6}$}\msep\mnum{$\textcolor{npgGreen}{\mathbf{1.0}}$}\msep\mnum{$\textcolor{npgNavy}{\mathbf{1.0}}$} \\
\texttt{pumsb\_star - 163} & \mnum{$\textcolor{npgPurple}{2.3}$}\msep\mnum{$\textcolor{npgBlue}{2.5}$}\msep\mnum{$\textcolor{npgRed}{\mathbf{1.3}}$}\msep\mnum{$\textcolor{npgBrown}{7.0}$}\msep\mnum{$\textcolor{npgOrange}{4.5}$}\msep\mnum{$\textcolor{npgGreen}{\mathbf{1.3}}$}\msep\mnum{$\textcolor{npgNavy}{\mathbf{1.3}}$} & \mnum{$\textcolor{npgPurple}{3.9}$}\msep\mnum{$\textcolor{npgBlue}{1.6}$}\msep\mnum{$\textcolor{npgRed}{\mathbf{1.0}}$}\msep\mnum{$\textcolor{npgBrown}{7.0}$}\msep\mnum{$\textcolor{npgOrange}{5.9}$}\msep\mnum{$\textcolor{npgGreen}{\mathbf{1.0}}$}\msep\mnum{$\textcolor{npgNavy}{\mathbf{1.0}}$} & \mnum{$\textcolor{npgPurple}{2.2}$}\msep\mnum{$\textcolor{npgBlue}{2.8}$}\msep\mnum{$\textcolor{npgRed}{\mathbf{1.0}}$}\msep\mnum{$\textcolor{npgBrown}{6.0}$}\msep\mnum{$\textcolor{npgGreen}{\mathbf{1.0}}$}\msep\mnum{$\textcolor{npgNavy}{\mathbf{1.0}}$} & \mnum{$\textcolor{npgPurple}{2.9}$}\msep\mnum{$\textcolor{npgBlue}{2.8}$}\msep\mnum{$\textcolor{npgRed}{\mathbf{1.0}}$}\msep\mnum{$\textcolor{npgBrown}{6.0}$}\msep\mnum{$\textcolor{npgGreen}{\mathbf{1.0}}$}\msep\mnum{$\textcolor{npgNavy}{\mathbf{1.0}}$} \\
\texttt{tmovie - 500} & \mnum{$\textcolor{npgPurple}{2.1}$}\msep\mnum{$\textcolor{npgBlue}{2.4}$}\msep\mnum{$\textcolor{npgRed}{2.4}$}\msep\mnum{$\textcolor{npgBrown}{7.0}$}\msep\mnum{$\textcolor{npgOrange}{\mathbf{1.5}}$}\msep\mnum{$\textcolor{npgGreen}{5.3}$}\msep\mnum{$\textcolor{npgNavy}{2.0}$} & \mnum{$\textcolor{npgPurple}{3.8}$}\msep\mnum{$\textcolor{npgBlue}{1.7}$}\msep\mnum{$\textcolor{npgRed}{\mathbf{1.0}}$}\msep\mnum{$\textcolor{npgBrown}{7.0}$}\msep\mnum{$\textcolor{npgOrange}{4.2}$}\msep\mnum{$\textcolor{npgGreen}{5.8}$}\msep\mnum{$\textcolor{npgNavy}{3.6}$} & \mnum{$\textcolor{npgPurple}{2.1}$}\msep\mnum{$\textcolor{npgBlue}{2.9}$}\msep\mnum{$\textcolor{npgRed}{\mathbf{1.3}}$}\msep\mnum{$\textcolor{npgBrown}{6.0}$}\msep\mnum{$\textcolor{npgGreen}{4.6}$}\msep\mnum{$\textcolor{npgNavy}{1.9}$} & \mnum{$\textcolor{npgPurple}{3.5}$}\msep\mnum{$\textcolor{npgBlue}{1.9}$}\msep\mnum{$\textcolor{npgRed}{\mathbf{1.1}}$}\msep\mnum{$\textcolor{npgBrown}{5.9}$}\msep\mnum{$\textcolor{npgGreen}{4.9}$}\msep\mnum{$\textcolor{npgNavy}{3.0}$} \\
\texttt{tretail - 135} & \mnum{$\textcolor{npgPurple}{\mathbf{1.6}}$}\msep\mnum{$\textcolor{npgBlue}{\mathbf{1.6}}$}\msep\mnum{$\textcolor{npgRed}{\mathbf{1.6}}$}\msep\mnum{$\textcolor{npgBrown}{5.8}$}\msep\mnum{$\textcolor{npgOrange}{2.7}$}\msep\mnum{$\textcolor{npgGreen}{\mathbf{1.6}}$}\msep\mnum{$\textcolor{npgNavy}{\mathbf{1.6}}$} & \mnum{$\textcolor{npgPurple}{\mathbf{1.4}}$}\msep\mnum{$\textcolor{npgBlue}{\mathbf{1.4}}$}\msep\mnum{$\textcolor{npgRed}{\mathbf{1.4}}$}\msep\mnum{$\textcolor{npgBrown}{6.9}$}\msep\mnum{$\textcolor{npgOrange}{4.1}$}\msep\mnum{$\textcolor{npgGreen}{\mathbf{1.4}}$}\msep\mnum{$\textcolor{npgNavy}{\mathbf{1.4}}$} & \mnum{$\textcolor{npgPurple}{1.8}$}\msep\mnum{$\textcolor{npgBlue}{1.3}$}\msep\mnum{$\textcolor{npgRed}{\mathbf{1.0}}$}\msep\mnum{$\textcolor{npgBrown}{4.8}$}\msep\mnum{$\textcolor{npgGreen}{\mathbf{1.0}}$}\msep\mnum{$\textcolor{npgNavy}{\mathbf{1.0}}$} & \mnum{$\textcolor{npgPurple}{2.2}$}\msep\mnum{$\textcolor{npgBlue}{1.4}$}\msep\mnum{$\textcolor{npgRed}{\mathbf{1.0}}$}\msep\mnum{$\textcolor{npgBrown}{5.5}$}\msep\mnum{$\textcolor{npgGreen}{\mathbf{1.0}}$}\msep\mnum{$\textcolor{npgNavy}{\mathbf{1.0}}$} \\
\midrule
\quad Times Ranked Highest: & \mnum{$\textcolor{npgPurple}{3}$}\msep\mnum{$\textcolor{npgBlue}{7}$}\msep\mnum{$\textcolor{npgRed}{15}$}\msep\mnum{$\textcolor{npgBrown}{0}$}\msep\mnum{$\textcolor{npgOrange}{1}$}\msep\mnum{$\textcolor{npgGreen}{14}$}\msep\mnum{$\textcolor{npgNavy}{16}$} & \mnum{$\textcolor{npgPurple}{1}$}\msep\mnum{$\textcolor{npgBlue}{3}$}\msep\mnum{$\textcolor{npgRed}{18}$}\msep\mnum{$\textcolor{npgBrown}{0}$}\msep\mnum{$\textcolor{npgOrange}{0}$}\msep\mnum{$\textcolor{npgGreen}{9}$}\msep\mnum{$\textcolor{npgNavy}{10}$} & \mnum{$\textcolor{npgPurple}{1}$}\msep\mnum{$\textcolor{npgBlue}{1}$}\msep\mnum{$\textcolor{npgRed}{14}$}\msep\mnum{$\textcolor{npgBrown}{0}$}\msep\mnum{$\textcolor{npgGreen}{14}$}\msep\mnum{$\textcolor{npgNavy}{19}$} & \mnum{$\textcolor{npgPurple}{0}$}\msep\mnum{$\textcolor{npgBlue}{0}$}\msep\mnum{$\textcolor{npgRed}{14}$}\msep\mnum{$\textcolor{npgBrown}{0}$}\msep\mnum{$\textcolor{npgGreen}{9}$}\msep\mnum{$\textcolor{npgNavy}{11}$} \\
\bottomrule
\end{tabular}
\begin{center}
\scriptsize
\textbf{Method:} \quad
\textcolor{npgPurple}{Independent} \quad
\textcolor{npgBlue}{Max Product} \quad
\textcolor{npgRed}{ArgMax Product} \quad
\textcolor{npgBrown}{Hybrid Belief-Propagation} \quad
\textcolor{npgOrange}{ITSELF} \quad
\textcolor{npgGreen}{PACMAP} \quad
\textcolor{npgNavy}{smooth-PACMAP} \quad
\end{center}
\caption{Average rankings of methods by MPE and MMAP estimates across ten trials. Winning entries in bold. Underlined results denote at least one timeout. In these cases, the target PAC guarantee is not attained, but weaker, Pareto-optimal certificates can be generated via the fixed-budget algorithm $\pi_{\mathrm B}$. Missing results indicate the method could not produce estimates due to technical limitations.}
\label{tab:rankings_mapmmap}
\end{table*}

\subsection{Continuous data}

Continuous PAC-MAP is more difficult than its discrete counterpart. 
Since we cannot in principle enumerate all assignments over a continuous domain, extra assumptions are required to guarantee computability in this case. 
To formalize this, we introduce some notation. 
For continuous $\mathcal Q \subseteq \mathbb R^n$, let $\mu(B)$ denote the mass of subset $B \subseteq \mathcal Q$ w.r.t. Lebesgue measure over the query space, conditional on evidence:
\begin{align*}
    \mu(B) := \int_{\bm q \in B} ~p(\bm q \mid \bm e) ~d \bm q.
\end{align*}
(In the continuous setting, the oracle is presumed omniscient not only w.r.t. densities evaluated at particular points but also w.r.t. integrals over arbitrary subsets.) 
Recall that $G_\varepsilon$ denotes the superlevel set of $\varepsilon$-approximate solutions. 
Then we have the following identifiability conditions.

\begin{lemma}[Identifiability] \label{lemma:id}
Let $\mathcal Q \subseteq \mathbb R^{n}$ and let $\mathcal B_r(\bm q)$ be a Euclidean ball of radius $r$ around point $\bm q$.
Then $\bm q^*$ is PAC-identifiable iff there exists some $\bm q \in G_\varepsilon$ and $r>0$ such that (a) $\mathcal B_r(\bm q) \subseteq G_\varepsilon$; and (b) $\mu\big(\mathcal B_r(\bm q)\big) > 0$.
\end{lemma}

\noindent Condition (a) states that there exists an $r$-neighborhood of some high-density point in which all neighboring points are also part of the superlevel set $G_\varepsilon$.
Without this condition, we have no way of lower-bounding the set's mass.
Condition (b) ensures that the resulting $r$-neighborhood has nonzero mass and can therefore be discovered via sampling. These requirements are quite mild in practice, and automatically satisfied under standard smoothness assumptions.

For example, suppose our density satisfies a modal Lipschitz condition similar to the one posited above, but defined w.r.t. Euclidean rather than Hamming distance. 
This suffices to ensure that continuous PAC-MAP solutions are identifiable, and suggests a principled strategy for discretizing the query space into hyperspheres of radius $r = \varepsilon / L$. 
Alternatively, we could use data-driven techniques to partition $\mathcal Q$ into hyperrectangles using decision trees or more general polytopes using ReLU networks. 
However we instantiate the partition, the basic logic is the same: convert the pdf over continuous $\mathcal Q$ into a pmf over a discrete representation, then follow the steps laid out above for solving PAC-MAP with fixed confidence or budget. So long as resulting cells are small enough to reliably distinguish between low- and high-density regions, Thms. 1-3 apply. Alternatively, continuous-armed BAI algorithms such as HOO \citep{bubeck2011}, DOO, or SOO \citep{munos2011} could be adapted for our setting, exploiting local smoothness via hierarchical refinement of the query space. 

\section{Experiments} \label{section:experiments}


The algorithms presented above are idealized in several respects, most notably in their assumption that we have access to constant-cost samplers and oracles. In practice, we must implement these functions with models learned from data. We use PCs for this task in our main experiments, since they guarantee compatible samplers and likelihoods, as well as flexible selection of query variables. Additional experiments against leading PGM methods are included in Appx. B. Code for reproducing all results is available in our dedicated GitHub repository.\footnote{\url{https://github.com/mattShorvon/PAC-MAP} for experiments with \texttt{Ind}, \texttt{MP}, \texttt{AMP}, \texttt{HBP} and both \texttt{PACMAP} variants, and \url{https://github.com/mattShorvon/MAP-Benchmark-PGMs-ITSELF} for experiments with \texttt{ITSELF} and PGM methods.} 



\paragraph{Benchmark}
We evaluate the performance of purely random ($\pi_{\mathrm R}$) and smooth ($\pi_{\mathrm S}$) PAC-MAP solvers against a range of methods for approximate MAP inference on the popular Twenty Datasets benchmark \citep{Davis_2021}.
We train a model on each dataset and apply approximate MAP solvers to the resulting PCs with varying proportions of query, evidence and nuisance variables, recording and ranking the probability of each method's MAP estimate. This is repeated ten times per dataset and query proportion, using different random partitions of $\bm Q$, $\bm E$ and $\bm V$. 
We use sum-product networks (SPNs) \citep{poon2011sum} as our PC architecture.
Average ranks of various methods are reported in Table \ref{tab:rankings_mapmmap}.

PAC parameters are fixed at $\varepsilon = \delta = 0.01$ throughout, with a maximum sample size of $M \leq2.5 \times10^6$. 
For our smooth bandit $\pi_{\mathrm S}$, we set $r=1$ and run a Hamming sweep every 250 samples. We compare against five approximate MAP solvers: \texttt{Independent} (\texttt{Ind}), a na\"{i}ve baseline \cite{park_darwiche_2004}; \texttt{MaxProduct} (\texttt{MP}), a linear time heuristic \cite{poon2011sum}, \texttt{ArgMaxProduct} (\texttt{AMP}), a quadratic time solver  \cite{conaty2017}; \texttt{Hybrid Belief-Propagation} (\texttt{HBP}), a reformulation approach \cite{maua2020}; and \texttt{ITSELF}, a neural-network approach \cite{arya_nips2024}.
For further details on these methods, we refer readers to the original texts. We exclude exact solvers such as those proposed by \citet{mei_maxSPN} and \citet{choi_solving_2022} as out of scope.
We observe that $\pi_{\mathrm S}$ dominates $\pi_{\mathrm R}$ across all trials. While theory tells us that the adaptive procedure should do no worse in expectation than a purely random approach, these results indicate that smoothness assumptions are reasonable for these datasets. 

Our bandits perform best in relatively low-dimensional settings, beating competitors when query dimension is $\leq 20\%$. 
In the high-dimensional setting (query proportion $= 50\%$), \texttt{AMP} tends to pull ahead, with $\pi_{\mathrm S}$ typically coming in second place.
Note that this coincides with a higher incidence of timeouts, suggesting that stronger results are possible with more samples.
However, there are limits to what sampling can realistically achieve. The largest datasets in this experiment have $500$ features. At a query proportion of $50\%$, the resulting state space has $2^{250} \approx 1.8 \times 10^{75}$ configurations. 
To put that into perspective, at $n=266$, the state space grows to approximately $10^{80}$, which is the estimated number of particles in the observable universe. 

In a benchmark against leading PGM experiments (results table reported in Appx. B), $\pi_{\mathrm R}$ and $\pi_{\mathrm S}$ dominate four leading competitors across two query dimensions, demonstrating their effectiveness across multiple model classes.



\begin{figure}[t]
    \centering
    \includegraphics[width=0.9\linewidth]{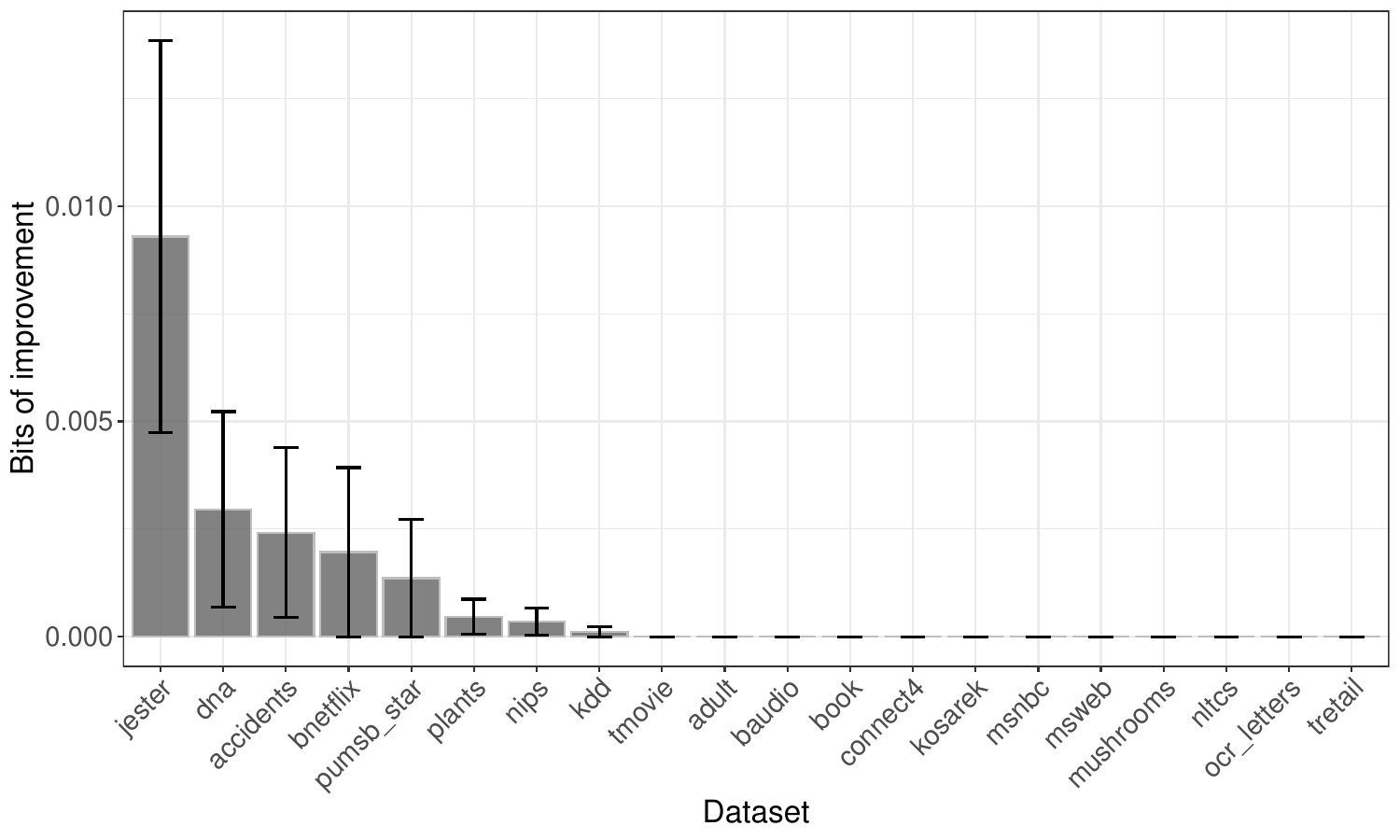}
    \caption{Average $\log_2$ ratio in MAP probability of \texttt{AMP} and smooth PAC-MAP (Alg. $\pi_{\mathrm S}$) across ten trials per dataset when using \texttt{AMP} outputs as a warm start for $\pi_{\mathrm S}$. Whiskers represent standard errors.}
    \label{fig:amp_pacmaph_barchart}
\end{figure}

\paragraph{Warm Starts} 
We combine PAC-MAP with existing heuristics by using outputs from an approximate solver as a warm start for randomized search. 
Specifically, we take the $\hat{\bm q}$ delivered by \texttt{AMP} as initialization for $\pi_{\mathrm S}$ across all Twenty Datasets.
In this collaborative setting, PAC techniques can only help---either elevating existing solutions with rigorous guarantees, or else finding more probable configurations to report. We find that $\pi_{\mathrm S}$ improves upon the results of \texttt{AMP} in $40\%$ of trials (see Fig. \ref{fig:amp_pacmaph_barchart}). Though some increases are modest, this reflects both the inherent difficulty of the task and strong performance of \texttt{AMP}.
More importantly, even in cases where randomized search does not find any atoms of greater mass, it still provides a PAC certificate for the \texttt{AMP} solution, giving practitioners greater confidence in their MAP estimates.

\begin{figure}[t]
    \centering
    \includegraphics[width=0.9\linewidth]{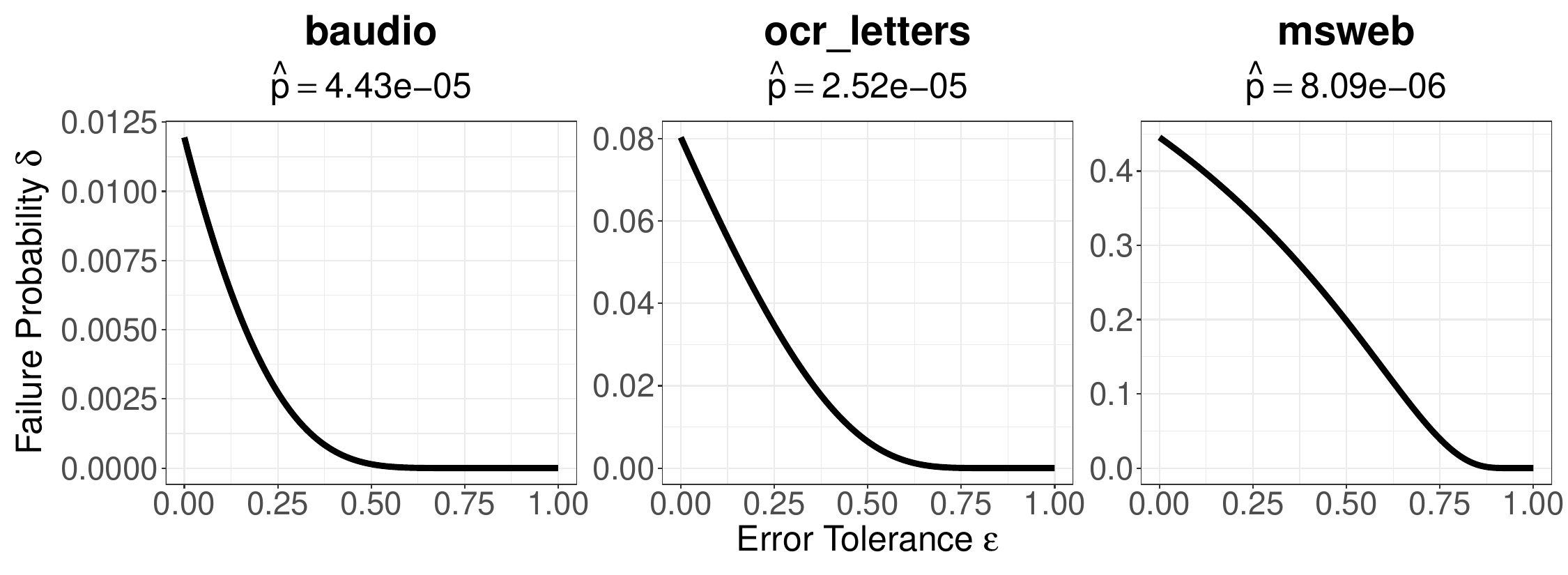}
    \caption{Pareto frontiers computed by $\pi_{\mathrm B}$ on three datasets where $10^6$ samples were insufficient for target guarantees. 
    Estimated MAP probabilities are given at the top of the graph.}
    \label{fig:pareto_fronts}
\end{figure}

\paragraph{Relaxing PAC Guarantees} In some instances, our fixed-confidence bandits time out before satisfying the target guarantee. In these cases, we revert to the fixed-budget bandit $\pi_{\mathrm B}$, using our data to recover a Pareto frontier of admissible ($\varepsilon$, $\delta$) pairs. Across the datasets where this occurs, we observe two consistent patterns: within a dataset, $\delta$ decreases monotonically as $\varepsilon$ increases, as expected; and across datasets, those with smaller estimated MAP probability $\hat p$ require substantially higher $\delta$ to reach any given $\varepsilon$, reflecting greater residual uncertainty in high-entropy, high-dimensional problems. In many cases, the relaxed guarantees are not much weaker than the original targets (see Fig. \ref{fig:pareto_fronts}). For example, on the \texttt{baudio} dataset, we have a PAC certificate at $\varepsilon = 0.01, \delta = 0.011$, just above our input level $\delta = 0.01$. On \texttt{ocrletters}, we have a solution at $\varepsilon = 0.05, \delta = 0.071$, a weaker but still suggestive result. The \texttt{msweb} dataset proved most challenging of the three, likely due to its high dimensionality and entropy: at $\varepsilon = 0.25$, the best failure probability we can guarantee is $\delta = 0.34$. Alternative PAC parameters to these points may be preferable depending on a user's priorities. 

\section{Discussion} \label{section:discussion}



We note several limitations of our method. First, and most obviously, it does not provide exact solutions except in relatively simple cases.
If exactness is essential, then alternative solvers
are more appropriate.
Second, PAC-MAP is intractable when min-entropy grows faster than $\mathcal O(\log n)$. 
This can lead to impractically large stopping times in some real world examples, making it hard to issue PAC certificates when optimizing over hundreds or thousands of variables.
This is an inevitable byproduct of the complexity of MAP itself. 
However, $\pi_\mathrm B$ remains a viable alternative in such cases, returning Pareto-optimal solutions on a fixed computational budget. 
Finally, the Hamming neighborhood scan is a fairly crude exploitation strategy that scales poorly with the radius hyperparameter. 
Still, we find that $\pi_{\mathrm S}$ fares reasonably well in our experiments with a fixed Hamming ball radius of $1$. 

To proponents of tractable probabilistic models, it may seem vaguely heretical to use PCs for randomized inference. A central motivation for this formalism is to render probabilistic reasoning exact and coherent. However, exact polytime MAP solutions are only possible in deterministic circuits, which show limited expressive efficiency in practice. 
We argue that randomization is underexplored in PCs, and likely has other valuable applications. 
There are also natural extensions to our methods that could use MCMC-based samplers and reinforcement-learning search strategies (e.g., Monte Carlo tree search) for incorporating additional structure into the sampling.

\section{Conclusion} \label{section:conclusion}

MAP inference is an important and challenging task in probabilistic reasoning. 
Because the problem is inherently intractable, creative strategies are required to compute approximate solutions that strike a reasonable balance between error and complexity. 
Our approach combines the theoretical rigor of PAC reasoning with the sampling and marginalization capabilities of probabilistic models to do just that. 
We have shown that a simple algorithm for PAC-MAP inference is uniformly optimal over the class of purely random strategies, and derived information theoretic tractability conditions for our solver.
We proved that an adaptive variant is minimax optimal, and found that it performs well in a range of experiments on real-world datasets.
Though the task remains infeasible in the worst case, we mitigate against this negative result with an anytime-valid algorithm that issues Pareto-optimal PAC certificates under arbitrary computational budgets. 
Experiments confirm that our method is not only competitive against state of the art solvers, but can team up with heuristic methods to improve upon their results and/or provide theoretical guarantees.
Future work will explore new directions for adaptive search and randomized MAP inference. 

%% file: arxiv_appx.tex
\section{A: Proofs}

In this section, we prove the paper's theoretical claims.\\ 

\noindent \textbf{Lemma 1} (Discovery complexity)\textbf{.} Fix any $\varepsilon, \delta \in (0,1)$. Define the $\varepsilon$-superlevel set: 
\begin{align*}
    G_\varepsilon := \big\{\bm q: p(\bm q \mid \bm e) \geq p^*(1 - \varepsilon)\big\},
\end{align*}
with corresponding mass $\mu_\varepsilon = \sum_{\bm q \in G_\varepsilon} p(\bm q \mid \bm e)$. 
Then Alg. 0 PAC-identifies $\bm q^*$ in $\mathcal O\big(\mu_\varepsilon^{-1} \log\delta^{-1}\big)$ samples.
    
\paragraph{Proof} To prove the lemma, we must show that $\mu_\varepsilon^{-1} \log\delta^{-1}$ samples are asymptotically sufficient for PAC-identification with Alg. 0. 

By the i.i.d. assumption, the probability of missing any $\varepsilon$-approximate query in $m$ random samples from $P(\bm Q \mid \bm e)$ is exactly $(1 - \mu_\varepsilon)^m$. To satisfy the PAC guarantee, this value must be upper bounded by $\delta$. Exploiting the fundamental inequality $1 - x \leq \exp(-x)$, we have
\begin{align*}
    (1 - \mu_\varepsilon)^m &\leq \exp(-m \mu_\varepsilon).
\end{align*}
We upper bound the rhs by $\delta$, take logs, and solve for $m$:
\begin{align*} 
    \exp(-m \mu_\varepsilon) &\leq \delta\\
    -m\mu_\varepsilon &\leq \log \delta \\ 
    m\mu_\varepsilon &\geq \log(1/\delta) \\ 
    m &\geq \frac{\log(1/\delta)}{\mu_\varepsilon}. 
\end{align*}
Thus $m = \mathcal O(\mu_\varepsilon^{-1} \log \delta^{-1})$, establishing that this many samples are sufficient in the worst case. $\Box$\\

\noindent \textbf{Remark}. To derive a lower bound, we start with the PAC inequality $(1 - \mu_\varepsilon)^m \leq \delta$. Taking logs and solving for $m$:
\begin{align*}
    m \log(1 - \mu_\varepsilon) &\leq \log \delta\\
    m &\geq \frac{\log \delta}{\log (1 - \mu_\varepsilon)}.
\end{align*}
This value converges to the upper bound as $\mu_\varepsilon \to 0$, but can be considerably lower when the superlevel set has relatively large mass. As noted in the main text, in the extreme case where $P$ is uniform, $\mu_\varepsilon = 1$ and $m=1$ sample suffices to PAC-identify $\bm q^*$.
\\

\noindent \textbf{Theorem 1} (Purely random PAC-MAP)\textbf{.} 
    Let $\Pi$ be the class of purely random PAC-MAP solvers, with sound policy subset $\Pi^* \subset \Pi$. 
    Let $\mathcal P$ be the class of pmfs on $\mathcal Q$ with min-entropy $h$.
    We have:
    \begin{itemize}[noitemsep]
        \item[(a)] $\pi_{\mathrm{R}} \in \Pi^*$.
        \item[(b)] With probability at least $1 - \delta$, $\pi_{\mathrm{R}}$ terminates after $\mathcal O\big(\exp(h) \log \delta^{-1}\big)$ random samples. 
        This rate is uniformly optimal: for every $P \in \mathcal P$ and any $\{\bm q_i\}_{i=1}^m \sim P$, no policy $\pi \in \Pi^*$ terminates faster. 
        \item[(c)] After $m$ random samples, $\pi_{\mathrm{R}}$ achieves simple regret 
        \begin{align*}
            R_m(\pi_{\mathrm R}, p) = d \cdot \exp\big(- \Theta(m p^*) \big).
        \end{align*}
        This rate is minimax optimal w.r.t. $\Pi$ and $\mathcal P$.
    \end{itemize}

\paragraph{Proof} The theorem makes three separate claims that we consider in turn. 

\paragraph{Soundness} To show that $\pi_{\mathrm R}$ is sound, we must demonstrate that any $\hat{\bm q}$ returned by the procedure is a valid PAC-MAP solution, i.e. satisfies the guarantee:
\begin{align}\label{eq:pac}
    \Pr\big(p(\hat{\bm q} \mid \bm e) \geq p^*(1 - \varepsilon)\big) \geq 1 - \delta.
\end{align}
This requires an analysis of two separate stopping criteria.
The first is triggered when $\hat p \geq \check p(1 - \varepsilon)$. Suppose this condition is met.
Note that the largest possible mass of any unsampled atom is $\check p$, so either (a) we have already seen the MAP configuration, in which case $\hat p = p^*$ ; or (b) we have not, in which case $p^* \leq \check p$. Under scenario (a), we can obviously stop since our solution is exact. Under scenario (b), we have
\begin{align*}
    \hat p \geq \check p(1 - \varepsilon) \geq p^*(1 - \varepsilon),
\end{align*}
which means we must have sampled at least one element of the superlevel set $G_\varepsilon$.
In either case, we can be certain that $\hat{\bm q}$ is (at least) an $\varepsilon$-approximate solution. 
That is why results under this condition are not just \textit{probably} approximately correct---they hold with $\delta=0$. Moreover, when $\hat p \geq \check p$, the solution is exact, i.e. $\hat p = p^*$ and $\varepsilon = \delta = 0$.

By contrast, a solution is at best PAC when $m \geq \hat p^{-1} (1 - \varepsilon) \log \delta^{-1}$. The argument here proceeds along similar lines to that for Lemma 1.
Specifically, we fail to return an $\varepsilon$-approximate solution if $\hat p < p^*(1 - \varepsilon)$. Thus the true MAP has to exceed $\hat p / (1 - \varepsilon)$ for $\pi_{\mathrm R}$ to err. What is the probability of such an error? Given that our sampler is i.i.d., the probability of missing some atom with mass greater than $\hat p / (1 - \varepsilon)$ in $m$ draws is exactly $\big( 1 - \hat p / (1 - \varepsilon) \big)^m \leq \exp\big(-m \hat p/(1 - \varepsilon)\big)$.
We upper bound the rhs by $\delta$ and solve for $m$:
\begin{align*}
    \exp\big(-m \hat p/(1 - \varepsilon)\big) &\leq \delta\\
    -m\hat p / (1 - \varepsilon) &\leq \log \delta\\
    m\hat p / (1 - \varepsilon) &\geq \log (1/\delta)\\
    m &\geq \frac{\log (1/\delta)}{\hat p / (1 - \varepsilon)}\\
    &= \hat p^{-1} ~(1 - \varepsilon) ~\log \delta^{-1}.
\end{align*}
This establishes the soundness of our second stopping criterion. Since $\pi_{\mathrm R}$ is purely random (i.e., relies only on i.i.d. sampling) and guaranteed to PAC-identify $\bm q^*$, we conclude that $\pi_{\mathrm R} \in \Pi^*$.

\paragraph{Certification complexity} 
Item (b) makes two claims: first, a high probability bound on the complexity of $\pi_{\mathrm R}$; second, that this bound is uniformly optimal w.r.t. $\Pi^*$ and $\mathcal P$. We will show each in turn.

Begin with the complexity bound. 
This analysis deliberately ignores the deterministic stopping condition of line 11, since only the probabilistic condition of line 14 holds uniformly for all distributions. The former, by contrast, applies exclusively in advantageous settings where mass is heavily concentrated. 
Let $S_m = \{\bm q_i\}_{i=1}^m \sim P(\bm Q \mid \bm e)$ denote a random sample of $m$ i.i.d. draws from the query distribution, with maximum observed probability $\hat p_m = \max_{\bm q \in S_m} ~p(\bm q \mid \bm e)$. 
Observe that $\hat p_m$ is a random variable that converges to $p^*$ as $m \rightarrow \infty$. 
We will show that for $\hat p_m$ to exceed $p^*(1 - \varepsilon)$ with high probability, it suffices to draw at most $M = \big \lceil 2^h \log \delta^{-1} \big \rceil$ samples.

If $S_m$ contains an element of the superlevel set $G_\varepsilon$, then the sample maximum satisfies $\hat p_m \geq p^*(1 - \varepsilon)$. Thus if the stopping condition is triggered, then soundness implies:
\begin{align*}
    \frac{(1 - \varepsilon) \log \delta^{-1}}{\hat p_m}~
    &\stackrel{\mathclap{\text{w.h.p.}}}{\leq} ~\frac{(1 - \varepsilon) \log \delta^{-1}}{p^*(1 - \varepsilon)}\\
    &= \frac{\log \delta^{-1}}{p^*}\\
    &= 2^{h} \log \delta^{-1},
\end{align*}
where the final line exploits the definition of the min-entropy, $h := H_\infty(\bm Q \mid \bm e) := -\log_2 p^*$.
This bound fails iff we do \textit{not} sample an element of $G_\varepsilon$, a failure mode that occurs with probability at most $\delta$. This establishes the following stochastic upper bound, which holds uniformly for all distributions $\mathcal P$ on $\mathcal Q$ with min-entropy $h$: 
\begin{align*}
    \forall P \in \mathcal P: ~\Pr_P \big( M_\mathrm R \leq 2^h \log \delta^{-1} \big) \geq 1 - \delta,
\end{align*}
where $M_\mathrm R$ denotes the random stopping time of $\pi_\mathrm R$.
This establishes the stated high probability upper bound on sample complexity. 

Next, we consider uniform optimality. To make this claim precise, we introduce some notation.
Let $(\bm q_m)_{m \geq 1} \sim P \in \mathcal P$ be an i.i.d. sample stream with induced filtration $\mathcal F_m^P = \sigma(\bm q_1, \dots, \bm q_m)$. 
Let $M_\pi(\mathcal F_m^P, \varepsilon, \delta)$ be a random variable denoting the stopping time of algorithm $\pi \in \Pi$, adapted to the filtration $\mathcal F_m^P$ and input PAC parameters. 
We write $M_{\mathrm R}(\mathcal F_m^P, \varepsilon, \delta)$ to denote the stopping time of $\pi_{\mathrm R}$, which we know to be an element of $\Pi$ by item (a).
We can now formalize the uniform optimality claim as follows: for any policy $\pi \in \Pi$, distribution $P \in \mathcal P$, and PAC parameters $\varepsilon, \delta \in (0,1)$: 
\begin{align*} 
    M_{\mathrm R}(\mathcal F_m^P, \varepsilon, \delta) \leq M_\pi(\mathcal F_m^P,\varepsilon, \delta).
\end{align*} 
This is a strong pointwise optimality claim, which states that no sound, purely random PAC-MAP solver terminates faster than $\pi_{\mathrm R}$ on any possible data stream. 

The proof proceeds by \textit{reductio}. Assume, for the sake of contradiction, that there exists some $\pi_0 \in \Pi$ such that $M_{0}(\mathcal F_m^P,\varepsilon, \delta) < M_{\mathrm R}(\mathcal F_m^P, \varepsilon, \delta)$.
We will show that $\pi_0$ cannot be an element of $\Pi$.

Recall that $\Pi$ is the class of all sound and purely random PAC-MAP solvers. Thus every element of $\Pi$ must satisfy the PAC guarantee of Eq. \ref{eq:pac}, using only i.i.d. draws from the sampling distribution to navigate the state space.  
The algorithms in this class therefore vary only with respect to their stopping criteria given some filtration $\mathcal F_m^P$.
We will show that any stopping condition faster than those of $M_{\mathrm R}(\mathcal F_m^P, \varepsilon, \delta)$ violates soundness, resulting in contradiction. 

We consider deterministic and stochastic stopping conditions in turn. For fixed $P, \varepsilon, \delta$, we abbreviate $m_1 := M_{\mathrm R}(\mathcal F_m^P, \varepsilon, \delta)$ and $m_0 := M_0(\mathcal F_m^P, \varepsilon, \delta)$, with $m_0 < m_1$ by assumption.

\paragraph{Case 1: Deterministic Stopping Condition} 
Suppose that $\pi_{\mathrm R}$ terminates at time $m_1$ because the deterministic condition $\hat p_{m_1} \geq \check p_{m_1}(1 - \varepsilon)$ is satisfied. 
Algorithm $\pi_0$ stops at time $m_0 < m_1$ on the same sample stream.
By definition of $m_1$, the deterministic stopping condition is not satisfied at any earlier time, so we have $\hat p_{m_0} < \check p_{m_0}(1 - \varepsilon)$.

At time $m_0$, $\pi_0$ has observed exactly the same information as $\pi_{\mathrm R}$, namely the filtration $\mathcal F_{m_0}^P$. 
After $m_0$ samples, we cannot rule out the possibility that the entire residual mass belongs to the as yet unobserved MAP configuration, in which case $p^* = \check p_{m_0}$. 
Under such a completion, the reported maximizer $\hat p_{m_0}$ is not in the $\varepsilon$-superlevel set:
\begin{align*}
    \hat p_{m_0} &< \check p_{m_0}(1 - \varepsilon)\\
    &= p^* (1 - \varepsilon),
\end{align*}
so any certificate issued at time $m_0$ is invalid. Since this failure mode occurs with nonzero probability under the sampling process, algorithm $\pi_0$ cannot issue a valid PAC certificate with $\delta=0$ in this setting, as required for the deterministic stopping condition.

\paragraph{Case 2: Probabilistic Stopping Condition}
Suppose instead that $\pi_{\mathrm R}$ terminates due to the probabilistic condition:
\begin{align*}
    m_1 \geq \frac{(1 - \varepsilon) \log \delta^{-1}}{\hat p_{m_1}},
\end{align*}
while $\pi_0$ stops at time $m_0 < m_1$ on the same sample stream.
By definition of $m_1$, the inequality fails for any $m < m_1$, giving:
\begin{align*}
    m_0 < \frac{(1 - \varepsilon) \log \delta^{-1}}{\hat p_{m_0}}.
\end{align*}
Solving for $\delta$, we have:
\begin{align*}
    \exp\big(-m \hat p_{m_0} / (1 - \varepsilon)\big) > \delta.
\end{align*}
Conditioned on the observed filtration $\mathcal F_{m_0}^P$, there exists a distribution consistent with the observations under which the superlevel set $G_\varepsilon$ has total mass exactly 
\begin{align*}
    \mu_\varepsilon = \frac{\hat p_{m_0}}{1 - \varepsilon},
\end{align*}
and none of its elements have yet been sampled. Under this distribution, the probability of failing to observe any element of $G_\varepsilon$ in $m_0$ samples is exactly $\exp\big(-m \hat p_{m_0} / (1 - \varepsilon)\big)$. Therefore, if $\pi_0$ stops at time $m_0$, then with probability strictly larger than $\delta$ its output fails to return a valid $\varepsilon$-approximation.
This violates the PAC guarantee, contradicting the assumption that $\pi_0 \in \Pi$. 

Thus in either case, any algorithm that terminates strictly before $m_1$ must violate PAC soundness. Hence the existence of some $\pi_0 \in \Pi$ such that $M_{0}(\mathcal F_m^P,\varepsilon, \delta) < M_{\mathrm R}(\mathcal F_m^P, \varepsilon, \delta)$ leads to contradiction. 

We therefore conclude that, for all $\pi \in \Pi, P \in \mathcal P$ and  $\varepsilon, \delta \in (0,1)$:
\begin{align*}
    M_{\mathrm R}(\mathcal F_m^P, \varepsilon, \delta) \leq M_\pi(\mathcal F_m^P,\varepsilon, \delta),
\end{align*}
thereby establishing uniform optimality.

\paragraph{Minimax regret} Item (c) of Thm. 1 states that $\pi_{\mathrm R}$ attains the minimax optimal regret w.r.t. $\Pi$ and $\mathcal P$. To show this, we first derive an upper bound on the simple regret of $\pi_{\mathrm R}$ w.r.t. $\mathcal P$, followed by a matching lower bound w.r.t. $\Pi$. 

We begin by decomposing the regret according to whether our sample $S_m$ includes $\bm q^*$.
Let $A$ be a binary variable with $A=a$ if $\bm q^* \in S_m$, and $A=a'$ if $\bm q^* \not \in S_m$.
Then we have:
\begin{align*}
    R_m(\pi_\mathrm R, p) &= \mathbb{E}\big[\log_2 (p^* / \hat p_{m+1})\big]\\ 
    &= \mathbb{E}\big[\log_2 (p^* / \hat p_{m+1}) \mid a\big] \cdot \Pr(a)\\
    &+ \mathbb{E}\big[\log_2 (p^* / \hat p_{m+1}) \mid a'\big] \cdot \Pr(a').
\end{align*}

\paragraph{Upper bound}
On the event that $\bm q^* \in S_m$, $\pi_\mathrm R$ has zero regret. 
Alternatively, if we fail to sample $\bm q^*$, then regret is large to the extent that $\hat p_{m+1}$ is small. 
Note that we can always set a trivial floor on $\hat p_{m+1}$ by initializing at $\hat p_0 = 2^{-n}$, as this represents the minimum possible MAP probability for all distributions on $\mathcal Q$.
Exploiting this fact, we have:
\begin{align*}
    \mathbb{E}\big[\log_2 (p^* / \hat p_{m+1}) \mid a'\big] \leq \log_2(p^* / 2^{-n}) = d,
\end{align*}
where recall that $d := D_\infty(P ~||~ U)$ represents the R\'{e}nyi divergence of infinite order between our true $P$ and a uniform alternative.
Since $\bm q^* \in S_m$ with probability $1 - (1-p^*)^m$, the miss event $a'$ has probability $\Pr(a') = (1-p^*)^m \le \exp(-mp^*)$. Therefore
\begin{align*}
    R_m(\pi_{\mathrm R}, p) &\le 0 \cdot \big(1 - (1-p^*)^m\big) + d \cdot \exp(-mp^*)\\
    &= d \cdot \exp(-mp^*).
\end{align*}

\paragraph{Lower bound}
We now show that no $\pi \in \Pi$ achieves better than $\Omega(d \cdot \exp(-mp^*))$ regret in the worst case.
Fix $h$ and define a family of distributions $\{p_{\bm q}\}_{\bm q \in \mathcal{Q}} \subset \mathcal{P}$, where each $p_{\bm q}$ places mass $p^* = 2^{-h}$ on a single mode at $\bm q$ and distributes the remaining mass uniformly over $\mathcal{Q} \setminus \{\bm q\}$. By construction, $p_{\bm q} \in \mathcal{P}$ for every $\bm q$.
This ``needle in the haystack'' distribution represents the adversarial setting for a purely random PAC-MAP solver, since for large $h$, we are likely to sample the flat region of the distribution rather than the single peak.

Let $\bm Q^* \sim U(\mathcal{Q})$ be a uniformly random mode location. By Yao's principle \citep{yao1977},
\begin{align*}
    \inf_{\pi \in \Pi} ~\sup_{p \in \mathcal{P}} ~R_m(\pi, p) \geq \inf_{\pi \in \Pi} ~\mathbb{E}_{\bm Q^*} \big[ R_m(\pi, p_{\bm Q^*}) \big].
\end{align*}
For any $\pi \in \Pi$ and any realization $\bm Q^* = \bm q^*$, the algorithm's samples $S_m$ are i.i.d. draws from $p_{\bm q^*}$, so $\bm q^* \in S_m$ with probability $1 - (1 - p^*)^m$. Equivalently, $\Pr(\bm q^* \notin S_m) = (1 - p^*)^m \geq \exp(-2mp^*)$ for $p^* \leq 1/2$.

By construction, each non-mode occurs with probability $\beta := (1 - p^*)/(2^n - 1)$.  
Conditional on $\bm q^* \notin S_m$, $\pi$'s recommendation therefore satisfies $\hat p_{m+1} = \beta \leq 2 \cdot 2^{-n}$. 
In this event, regret satisfies
\begin{align*}
    \log_2(p^* / \hat p_{m+1}) \geq \log_2(p^* \cdot 2^{n-1}) = d - 1.
\end{align*}
Combining,
\begin{align*}
    \mathbb{E}_{\bm Q^*} \big[ R_m(\pi, p_{\bm Q^*}) \big] \geq (d - 1) \cdot \exp(-2mp^*).
\end{align*}
Thus 
\begin{align*}
    \inf_{\pi \in \Pi} ~\sup_{p \in \mathcal{P}} ~R_m(\pi, p) =  d \cdot \exp\big(- \Omega (mp^*) \big),
\end{align*}
matching the upper bound. The minimax regret rate is therefore $d \cdot \exp(-\Theta(mp^*))$, attained by $\pi_{\mathrm R}$.
This completes the proof of Thm. 1.
$\Box$ \\

\noindent \textbf{Corollary 1.1.} With fixed PAC parameters $\varepsilon, \delta$, $\pi_{\mathrm R}$ is tractable if and only if $h = \mathcal O(\log n)$.

\paragraph{Proof} This corollary follows immediately from the fact that tractable PAC-MAP solutions must be upperbounded by a polynomial function of $n$ and $\log \delta^{-1}$. 
With fixed PAC parameters, the only relevant input is the dimensionality $n$ of the Boolean hypercube $\mathcal Q$, which enters into the complexity result via the distributional difficulty term. To see this, it helps to rewrite the min-entropy in terms of the R\'{e}nyi divergence $d = n-h$.
Observe that $2^{h} = \Theta\Big(\exp\big(n - d\big)\Big)$. 
To keep this factor from blowing up, the exponentiated difference must grow at most logarithmically in $n$, thereby reducing the term's overall complexity to $\mathcal O\big(\text{poly}(n)\big)$, as desired. $\Box$\\

\noindent \textbf{Theorem 2} (Smooth PAC-MAP) \textbf{.}
Let $\Pi$ be the class of PAC-MAP solvers, with sound policy subset $\Pi^* \subset \Pi$.
    Let $\mathcal P$ be the class of pmfs on $\mathcal Q$ with min-entropy $h$ that are modally Lipschitz with parameters $L,r$.
    Assume that $v_r w_r = \mathcal O\big(\exp(h)\big)$.
    We have:
    \begin{itemize}[noitemsep]
        \item[(a)] $\pi_{\mathrm{S}} \in \Pi^*$.
        \item[(b)] Assume that $v_r \log \delta^{-1} = o\big(\exp(d) w_k \big)$. Then with probability at least $1 - \delta$, $\pi_{\mathrm{S}}$ terminates after $\mathcal O \big(\exp(h) w_k^{-1} \log \delta^{-1} \big)$ oracle calls. This rate is minimax optimal w.r.t. $\Pi^*$ and $\mathcal P$.
        \item[(c)] For some $c_1>0$, $\pi_{\mathrm S}$ achieves simple regret
        \begin{align*}
            R_m(\pi_{\mathrm S}, p) = \mathcal O\Big(L &\sum_{s=0}^{r-1} \exp(-c_1 m p^*w_s) +\\ 
            &(d - Lr)_+ \exp(-c_1 m p^* w_r)\Big).
        \end{align*}
        For each feasible shell $s < r$ satisfying $v_s=o(\exp(d)w_s)$, the corresponding tail rate is minimax optimal w.r.t. $\Pi$ and $\mathcal P$ up to constants in the exponent, i.e. there exist some $C_0, c_0 > 0$ such that:
        \begin{align*}
            \inf_{\pi \in \Pi} ~\sup_{P \in \mathcal P} ~\Pr_P \big( \Delta_m^\pi  > Ls \big) \geq C_0 \exp(- c_0 m p^* w_s).
        \end{align*}
    \end{itemize}
\paragraph{Proof}
As with Thm. 1, we prove each of Thm. 2's claims in turn.
For completeness, we restate several relevant definitions:
\begin{align*}
    w_s &:= \sum_{j=0}^s {n\choose j}2^{-Lj}, \
    k:= \min \left\{ r,\left \lfloor \frac{\log_2(1/(1-\varepsilon))}{L} \right\rfloor \right\},\\ 
    \mu_k &:= p^*w_k,
    \qquad
    M_\delta := \left\lceil
    \mu_k^{-1}\log(2\delta^{-1})
    \right\rceil.
\end{align*}
Note that we suppress the dependence of $w_s$ on $L$, $k$ on $L, r$ and $\varepsilon$, and $\mu_k$ on $p^*$ (and therefore $h$). 

\paragraph{Soundness}
Two preliminary observations. First, exploitation cannot compromise soundness. Each sweep enlarges the candidate set $S$ without removing atoms and without incrementing the random-sample count $m$, so $\hat p$ only increases, $\check p$ only decreases, and it suffices to verify Eq.~\eqref{eq:pac} for the $m$ i.i.d.\ draws alone, since any $\hat{\bm q}$ that $\pi_{\mathrm S}$ reports is at least as probable. Second, the modal Lipschitz property inflates the relevant target. Every atom at Hamming distance $j \leq k$ from $\bm q^*$ has mass at least $p^* 2^{-Lj}$, and there are $\binom{n}{j}$ such atoms, so the ball $\mathcal B_{k}(\bm q^*)$ lies in $G_\varepsilon$ and carries mass
\begin{align*}
    P\big(\mathcal B_{k}(\bm q^*)\big) \;\geq\; \sum_{j=0}^{k} \binom{n}{j}\, p^* 2^{-Lj} \;=\; \mu_k.
\end{align*}

As before, we analyze two stopping criteria. The deterministic criterion, triggered when $\hat p \geq \check p(1-\varepsilon)$, is identical to that of $\pi_{\mathrm R}$. The argument of Thm.~\ref{thm:pacmap_bin} applies verbatim to the augmented set $S$ (sweeps only tighten it), so any solution returned here is exact or $\varepsilon$-approximate with $\delta = 0$.

The probabilistic criterion is where smoothness enters. Here $\pi_{\mathrm S}$ halts once $m \geq M := w_k^{-1}\, \hat p^{-1}(1-\varepsilon)\log\delta^{-1}$. We fail to return an $\varepsilon$-approximate solution only if $\hat p < p^*(1-\varepsilon)$, i.e.\ $p^* > \hat p/(1-\varepsilon)$. In this event the $\varepsilon$-good ball $\mathcal B_{k}(\bm q^*)$ has mass at least $\mu_k > w_k\, \hat p/(1-\varepsilon)$, and none of the $m$ random draws can lie in it: any such draw would have mass at least $p^*(1-\varepsilon) > \hat p$, contradicting the value of $\hat p$. Since the sampler is i.i.d., the probability of missing a region of mass exceeding $w_k \hat p/(1-\varepsilon)$ in $m$ draws is at most $\big(1 - w_k \hat p/(1-\varepsilon)\big)^m \leq \exp\big(-m\, w_k \hat p/(1-\varepsilon)\big)$. Bounding the rhs by $\delta$ and solving for $m$:
\begin{align*}
    \exp\big(-m\, w_k \hat p/(1-\varepsilon)\big) &\leq \delta\\
    m\, w_k \hat p/(1-\varepsilon) &\geq \log(1/\delta)\\
    m &\geq w_k^{-1}\, \hat p^{-1}(1-\varepsilon)\log\delta^{-1} =: M.
\end{align*}
So at stopping time, the failure probability is at most $\delta$. This is exactly the $\pi_{\mathrm R}$ bound with the missed mass scaled up by $w_k$.
Whereas the purely random solver must exclude a single atom of mass $\hat p/(1-\varepsilon)$, modal Lipschitz lets $\pi_{\mathrm S}$ exclude an entire $\varepsilon$-good ball $w_k$ times heavier, which is what divides the threshold by $w_k$. Combining both criteria, $\pi_{\mathrm S}$ satisfies Eq.~\eqref{eq:pac} for all $\varepsilon, \delta \in (0,1)$, hence $\pi_{\mathrm S} \in \Pi^*$.

\paragraph{Certification complexity}
For a nonnegative random variable $T$, define its $(1-\delta)$-quantile under $P$ by
\begin{align*}
	Q_{1-\delta}^{P}(T) := \inf \left \{ t : \Pr_{P}(T\le t)\ge 1-\delta \right\}.
\end{align*}
We prove an upper bound for $\pi_{\mathrm S}$ and a matching minimax lower bound on this quantile.

\paragraph{Upper bound}
By the inner-ball estimate from the soundness proof,
\begin{align*}
	\mathcal B_{k}(\bm q^*) \subseteq G_\varepsilon \qquad \text{and} \qquad
	P\left(\mathcal B_{k}(\bm q^*)\right) \ge \mu_k.
\end{align*}
Consequently, the probability that the first $M_\delta$ random draws all miss this ball is at most
\begin{align*}
	(1-\mu_k)^{M_\delta} \le \exp(-M_\delta\mu_k) \le \delta/2.
\end{align*}
Thus, with probability at least $1-\delta/2$, some random draw enters $\mathcal B_{k}(\bm q^*)$ by random-draw count $M_\delta$.
Let $m_0\le M_\delta$ denote the first such draw. Thereafter, $\hat p_m\ge p^*(1-\varepsilon)$, since sweeps can only increase the incumbent probability. Hence, for every $m\ge m_0$,
\begin{align*}
	M_m = w_k^{-1}\hat p_m^{-1}(1-\varepsilon) \log\delta^{-1} \le w_k^{-1}(p^*)^{-1}\log\delta^{-1} < M_\delta.
\end{align*}
It follows that, on this event, the probabilistic stopping criterion has fired by the time the random-draw count reaches $M_\delta$.

It remains to bound the oracle-call cost of the intervening sweeps. 
At each decision epoch, $\pi_{\mathrm S}$ performs a sweep of the best unswept candidate with probability $\eta$, and otherwise takes a random draw; if no unswept candidate exists, it takes a random draw. 
Let $S_{M_\delta}$ be the number of sweeps performed before the $M_\delta$-th random draw. Then $S_{M_\delta}$ is stochastically dominated by the number of failures before $M_\delta$ successes in independent Bernoulli trials with success probability $1-\eta$.
A standard negative-binomial concentration bound therefore gives, for some universal constant $C_0>0$:
\begin{align*}
	\Pr \left( S_{M_\delta} > C_0\left( M_\delta\eta+\log(2\delta^{-1}) \right) \right)
\le \delta/2.
\end{align*}
Each random draw requires at most one oracle evaluation and each sweep requires at most $v_r$ oracle evaluations. 
Therefore, except on an event of probability at most $\delta/2$,
\begin{align*}
	T_{\pi_{\mathrm S}} &\le M_\delta+v_r S_{M_\delta} \\
	&= \mathcal O\left(M_\delta + v_r M_\delta \eta + v_r \log(2\delta^{-1}) \right).
\end{align*}
Because $\eta \asymp 1/v_r$, the second term is $\mathcal O(M_\delta)$. 
Our assumption that $v_r w_r = \mathcal O\big(\exp(h)\big)$ entails that $v_rw_k=\mathcal O\big(\exp(h)\big)$ (since $w_r \geq w_k$), and hence
\begin{align*}
	v_r\log(2\delta^{-1}) = \mathcal O(M_\delta).
\end{align*}
Combining the good-ball event and the sweep-count event by a union bound, we obtain
\begin{align*}
	&\sup_{P \in \mathcal P} ~Q_{1-\delta}^{P}(T_{\pi_{\mathrm S}}) = 
	\mathcal O\left( \mu_k^{-1}\log\delta^{-1} \right)\\
    &=\mathcal O \left( \exp(h)w_k^{-1}\log\delta^{-1} \right).
\end{align*}

\paragraph{Lower bound}
We use a hidden-cap family of distributions. Write $N:=2^n$ and define
\[
p_{\bm x}(\bm q) :=
\begin{cases}
p^* 2^{-L\|\bm q-\bm x\|_1},
& \bm q\in \mathcal B_k(\bm x), \\[2mm]
\beta,
& \bm q\notin \mathcal B_k(\bm x),
\end{cases} \qquad \beta := \frac{1-\mu_k}{N - v_k}.
\]
Assume that this hidden-cap family is feasible, i.e. $p^*2^{-L(k+1)} \leq \beta$ and, whenever $k < r$, $\beta < p^*(1-\varepsilon)$.
The first inequality ensures that the modal Lipschitz condition continues to hold at all distances $k < j\le r$; the second ensures that the only $\varepsilon$-good atoms are those in $\mathcal B_k(\bm x)$.
Under these conditions, $P_{\bm x}\in\mathcal P$, its unique mode has mass $p^*$, and
\begin{align*}
	P_{\bm x}(\mathcal B_k(\bm x)) = p^*w_k =: \mu_k.
\end{align*}

Now draw the hidden mode location $\bm X \sim U(\mathcal Q)$, and consider an arbitrary, possibly randomized sound policy $\pi \in \Pi^*$. 
Each sampled candidate is evaluated by the oracle and counts as one oracle call. 
Extend the policy after it terminates using arbitrary dummy oracle queries; this does not change its stopping time or reported solution.

Let $E_t$ be the event that none of the first $t$ evaluated atoms lies in $\mathcal B_k(\bm X)$. 
On $E_t$, every oracle response equals $\beta$. 
Each nonrevealing evaluation can exclude at most $v_k$ possible values of $\bm X$. 
Consequently, conditional on any nonrevealing transcript of length $u \le t$, the posterior on $\bm X$ is uniform over a set of cardinality at least
\begin{align*}
	K_u\ge N-u v_k.
\end{align*}
In particular, provided $t v_k\le N/2$, a targeted oracle query reveals the hidden cap with conditional probability at most
\begin{align*}
	\frac{v_k}{K_u} \le \frac{2v_k}{N}.
\end{align*}
A random draw reveals the hidden cap with conditional probability exactly $\mu_k$. 
Thus, while $E_t$ holds, the conditional probability of a reveal at any oracle call is at most $\rho := \max \left\{ \mu_k, (2v_k / N) \right\}$.
Given our assumption that $v_r\log\delta^{-1} = o\left(\exp(d)w_k\right)$ and the fact that $v_k\le v_r$, we have
\begin{align*}
	\frac{v_k}{N} = o(\mu_k) \qquad\text{and}\qquad
	\frac{v_k}{N} \mu_k^{-1}\log\delta^{-1} = o(1).
\end{align*}
Hence, for sufficiently large problem instances, $\rho\le C\mu_k$ for some constant $C$, and $t v_k\le N/2$ for all $t = \mathcal O(\mu_k^{-1} \log \delta^{-1})$.
Assuming $C\mu_k\le 1/2$, we obtain
\begin{align*}
	\Pr(E_t) \ge (1-C\mu_k)^t \ge \exp(-2C\mu_k t).
\end{align*}
On $E_t$, if the policy has terminated by time $t$, its reported atom is $\varepsilon$-good for at most $v_k$ of the at least $K_t$ remaining mode locations. Therefore,
\begin{align*}
	\Pr\left(\hat{\bm q}_t^{\pi} \in G_\varepsilon \mid E_t,\ T_\pi\le t \right) \le \frac{v_k}{K_t} \le \frac{2v_k}{N}.
\end{align*}
For sufficiently large instances the final quantity is at most $1/2$.
Since $\pi$ is sound for every $P_{\bm x}$, its error probability averaged over the uniform prior on $\bm X$ is at most $\delta$.
It follows that
\begin{align*}
	\Pr(E_t,\ T_\pi\le t) \le 2\delta.
\end{align*}
Choose
\begin{align*}
	t_\delta := \left \lfloor \frac{1}{2C\mu_k} \log \frac{1}{4\delta} \right \rfloor.
\end{align*}
Then $\Pr(E_{t_\delta})\ge4\delta$, and consequently
\begin{align*}
	\Pr(T_\pi>t_\delta) \ge \Pr(E_{t_\delta}) - \Pr(E_{t_\delta},T_\pi\le t_\delta) \ge 2\delta > \delta.
\end{align*}
This probability is averaged over the uniform prior on $\bm X$.
Therefore, there exists at least one $\bm x\in\mathcal Q$ such that
\begin{align*}
	\Pr_{P_{\bm x},\pi}(T_\pi>t_\delta)>\delta,
\end{align*}
and hence
\begin{align*}
	Q_{1-\delta}^{P_{\bm x}}(T_\pi) > t_\delta = \Omega\left( \mu_k^{-1}\log\delta^{-1} \right).
\end{align*}
Because this holds for every sound policy, we have
\begin{align*}
	\inf_{\pi\in\Pi^*} ~\sup_{P\in\mathcal P} ~Q_{1-\delta}^{P}(T_\pi) = \Omega\left( \exp(h) w_k^{-1} \log \delta^{-1} \right).
\end{align*}
Together with the upper bound, this proves that
\begin{align*}
	\inf_{\pi\in\Pi^*} ~\sup_{P\in\mathcal P} ~Q_{1-\delta}^{P}(T_\pi) = \Theta\left( \exp(h) w_k^{-1} \log \delta^{-1} \right),
\end{align*}
and that $\pi_{\mathrm S}$ attains the minimax-optimal $(1-\delta)$-quantile certification rate.

\paragraph{Minimax regret}
We prove the regret upper bound for $\pi_{\mathrm S}$, followed by the shell-wise minimax lower bound.

\paragraph{Upper bound}
Let $\Delta_m^{\mathrm S} := \log_2(p^*/\hat p_{m+1}^{\mathrm S})$ denote the loss of $\pi_{\mathrm S}$ after $m$ oracle calls. For each $s\le r$, the modal Lipschitz condition implies
\begin{align*}
	\mathcal B_s(\bm q^*)\subseteq \{\bm q:\log_2 \big(p^*/p(\bm q\mid\bm e)\big)\le Ls\},
\end{align*}
and $P \left(\mathcal B_s(\bm q^*)\right) \ge p^*w_s$.
Write $\mu_s:=p^*w_s$.

We first relate oracle calls to random draws. Let $N_m$ be the number of genuine random draws completed by $\pi_{\mathrm S}$ after $m$ oracle calls. Since $\pi_{\mathrm S}$ performs a radius-$r$ sweep with probability $\eta\asymp 1/v_r$, and each sweep costs at most $v_r$ oracle calls, a standard negative-binomial concentration argument gives constants $a,b>0$ such that
\begin{align*}
	\Pr(N_m<am)\le \exp(-bm/v_r).
\end{align*}
Indeed, the number of sweeps before $g$ random draws is stochastically dominated by the number of failures before $g$ successes in Bernoulli trials with success probability $1-\eta$. Taking $g=\lfloor am\rfloor$ for sufficiently small $a>0$, the probability that the sweep cost prevents $g$ random draws from occurring by time $m$ is at most $\exp(-bm/v_r)$.

Fix $s\le r$. On the event ${N_m\ge am}$, if at least one of the first $\lfloor am\rfloor$ random draws lands in $\mathcal B_s(\bm q^*)$, then the incumbent returned by $\pi_{\mathrm S}$ has regret at most $Ls$, since sweeps can only improve the incumbent. Therefore
\begin{align*}
	\Pr(\Delta_m^{\mathrm S}>Ls) \le \Pr(N_m<am) + (1-\mu_s)^{\lfloor am\rfloor}.
\end{align*}
Using $1 - x \le \exp(-x)$, this yields
\begin{align*}
	\Pr(\Delta_m^{\mathrm S}>Ls) \le \exp(-bm/v_r) + \exp(-a m\mu_s).
\end{align*}
Under the sweep-amortization condition $v_rw_r=\mathcal O\big(\exp(h)\big)$, there exists some $C>0$ such that
\begin{align*}
	\frac{1}{v_r}\ge C^{-1}p^*w_r\ge C^{-1}p^*w_s=C^{-1}\mu_s,
\end{align*}
so the first term is absorbed into the second after adjusting constants. 
Therefore, for some $c_1, C_1>0$,
\begin{align*}
	\Pr(\Delta_m^{\mathrm S}>Ls) \le C_1 \exp(-c_1 m p^*w_s), \qquad s\le r.
\end{align*}

We now convert this tail bound into an expected regret bound. The truncated regret satisfies
\begin{align*}
	\mathbb E[\Delta_m^{\mathrm S}\wedge Lr] \le L\sum_{s=0}^{r-1} \Pr(\Delta_m^{\mathrm S}>Ls).
\end{align*}
Given our initialization at $\hat p_0^{\mathrm S} := 2^{-n}$, which represents a natural lower bound on MAP probabilities, we have
\begin{align*}
	\Delta_m^{\mathrm S} \leq \log_2 (p^* / \hat p_0^{\mathrm S}) = -h+n = d.
\end{align*}
Hence the layer-cake integral for the residual truncates at $d$:
\begin{align}
    &\mathbb E\big[(\Delta_m^{\mathrm S} - Lr)_+\big] \\ &= \int_{Lr}^{d}\Pr(\Delta_m^{\mathrm S} > \gamma)\,d\gamma \le (d-Lr)_+\,\Pr(\Delta_m^{\mathrm S} > Lr) \\
    &\le C_1(d-Lr)_+\exp(-c_1 m p^* w_r),
\end{align}
where the integrand vanishes above $d$ by the ceiling, is bounded on $[Lr,d]$ by its value at the left endpoint, and the last step applies the shell tail bound at $s=r$.

Combining,
\begin{align}
	&R_m(\pi_{\mathrm S},p) = \mathbb E[\Delta_m^{\mathrm S}] = \\
    &\mathcal O \bigg( L\sum_{s=0}^{r-1} \exp(-c_1 m p^*w_s) \\
    &+ (d-Lr)_+ \exp(-c_1 m p^*w_r) \bigg).
\end{align}

\paragraph{Lower bound}
Next, we prove the shell-wise minimax lower bound. Fix a feasible shell $s<r$.
For each hidden mode location $\bm x \in \mathcal Q$, define
\[
p_{\bm x,s}(\bm q) :=
\begin{cases}
p^*2^{-L\|\bm q-\bm x\|_1},
& \bm q\in \mathcal B_s(\bm x),\\
\beta_s,
& \bm q\notin \mathcal B_s(\bm x),
\end{cases}
\]
where
\begin{align*}
	\beta_s := \frac{1-p^*w_s}{2^n-v_s}.
\end{align*}
By feasibility,
\begin{align*}
	p^*2^{-L(s+1)} \le \beta_s < p^*2^{-Ls}.
\end{align*}
The lower inequality ensures that every point in
\(\mathcal B_r(\bm x)\setminus \mathcal B_s(\bm x)\) has mass at least
\(p^*2^{-L\|\bm q-\bm x\|_1}\), with the binding constraint occurring at
distance \(s+1\).
The upper inequality ensures that the only atoms with regret at most $Ls$ are those in $\mathcal B_s(\bm x)$. 
Thus $P_{\bm x,s}\in\mathcal P$, its mode is $\bm x$, and $P_{\bm x,s}(\mathcal B_s(\bm x))=p^*w_s=\mu_s$.

Let $\bm X \sim U(\mathcal Q)$, and consider an arbitrary policy $\pi$. 
Let $E_m$ denote the event that none of the first $m$ evaluated atoms lies in $\mathcal B_s(\bm X)$. 
On $E_m$, every oracle response is equal to the background value $\beta_s$. 
Hence the transcript gives no information about $\bm X$ except by ruling out hidden mode locations whose radius-$s$ caps contain one of the queried atoms.

After $u$ nonrevealing evaluations, at most $uv_s$ hidden mode locations have been ruled out. 
Therefore, as long as $uv_s\le 2^{n-1}$, the posterior support for $\bm X$ has cardinality at least $2^{n-1}$. 
Conditional on any such nonrevealing transcript, a targeted oracle query hits $\mathcal B_s(\bm X)$ with probability at most $2 v_s / 2^n$.
A random draw from $P_{\bm X,s}$, by contrast, hits $\mathcal B_s(\bm X)$ with probability exactly $\mu_s=p^*w_s$. 
For $m = \mathcal O(\mu_s^{-1})$, the hypothesis $v_s = o\big(\exp(d)w_s\big)$ gives $m v_s = o\big(\exp(n)\big)$, since $2^d = 2^{n}p^*$ and $\mu_s = p^* w_s$. Thus $u v_s \le 2^{n-1}$ for all $u \le m$ and the posterior-cardinality bound applies throughout. Consequently 
\begin{align*}
	\Pr(E_m) \ge \left( 1-\mu_s-\frac{2v_s}{2^n} \right)^m.
\end{align*}
Writing $y := \mu_s + 2v_s/2^n$, the same hypothesis gives $2v_s/2^n = o(\mu_s)$, so $y \to 0$ and in particular $y \le 1/2$ for large $n$. On this range, $1 - y \ge \exp(-2 y)$, giving 
\begin{align*}
    \Pr(E_m) \ge \exp\Big[-2 m\Big(p^* w_s + \tfrac{2v_s}{2^n}\Big)\Big].
\end{align*}

On $E_m$, the returned $\hat{\bm q}_m^\pi$ is $Ls$-good only for hidden locations $\bm X$ with $\hat{\bm q}_m^\pi \in \mathcal B_s(\bm X)$. There are at most $v_s$ such hidden locations.
Hence for large $n$, 
\begin{align}
	\Pr \big(\Delta_m^\pi \le Ls \mid E_m\big) \le \frac{2v_s}{2^n} \le \frac12.
\end{align}
Therefore
\begin{align}
	\Pr\big(\Delta_m^\pi > Ls\big) &\ge \Pr(E_m) - \Pr\big(E_m,\, \Delta_m^\pi \le Ls\big) \\ &\ge \Pr(E_m)\Big(1 - \tfrac{2v_s}{2^n}\Big) \ge \tfrac12\Pr(E_m),
\end{align}
where these probabilities are taken under $\bm X \sim U(\mathcal Q)$. 
By Yao's principle \citep{yao1977},
\begin{align*}
    \sup_{P\in\mathcal P}~ \Pr_P(\Delta_m^\pi > Ls) \ge \mathbb E_{\bm X}\big[\Pr(\Delta_m^\pi > Ls)\big],
\end{align*}
(the worst case dominates the uniform-prior average) and using the bound on $\Pr(E_m)$ from above,
\begin{align*}
    \sup_{P\in\mathcal P}\Pr_P(\Delta_m^\pi > Ls) \ge \tfrac12\exp\Big[-2 m\Big(p^* w_s + \tfrac{2v_s}{2^n}\Big)\Big].
\end{align*}
As $\pi$ was arbitrary, the same bound holds for $\inf_{\pi\in\Pi} ~\sup_{P\in\mathcal P}$.

Finally, $v_s = o(\exp(d)w_s)$ gives $2v_s/2^n = o(p^* w_s)$, so the blind-search term is absorbed and, for constants $C_0, c_0 > 0$,
\begin{align*}
    \inf_{\pi\in\Pi}~ \sup_{P\in\mathcal P} ~\Pr_P(\Delta_m^\pi > Ls) \ge C_0\exp(-c_0 m p^* w_s),
\end{align*}
matching the upper tail rate at shell $s$ up to the exponent constant. $\Box$\\

\noindent \textbf{Theorem 3} (Budget PAC-MAP)\textbf{.} $\pi_{\mathrm B}$ is sound and complete w.r.t. PAC parameters, returning all and only the admissible $(\varepsilon, \delta)$ pairs for a given sample $S$ and corresponding MAP estimate $\hat{\bm q}, \hat p$.
The procedure executes in time $\mathcal O(M)$.

\paragraph{Proof} The theorem makes three claims: that $\pi_{\mathrm B}$ is (1) sound; (2) complete; and (3) executes in linear time. Item (3) is immediately evident from the pseudocode, as we simply draw $M$ samples from $P$ and perform a handful of trivial calculations whose complexity is independent of algorithm inputs. Therefore, the following will focus on claims (1) and (2).

$\pi_{\mathrm B}$ is sound if the empirical maximum $\hat p$ satisfies the target PAC guarantee at any $\varepsilon \in \bm \varepsilon$ and corresponding confidence level $\delta(\hat p, \varepsilon)$ returned by the procedure. This is obvious in the exact case, when $\check p \leq \hat p$, for the same reasons given in the proof above for Thm. 1. If residual mass exceeds the maximal observed probability, however, a range of slack parameters are feasible. Specifically, $\varepsilon$ can take any value in the interval $[0, 1 - \hat p)$. The lower bound is fixed by the nonnegativity constraint, while the upper bound follows from rearranging the inequality $\hat p / (1 - \varepsilon) \leq 1$. Once again, we exploit the i.i.d. sampling procedure to bound our failure probability via $\delta \geq \big(1 - \hat p/(1 - \varepsilon) \big)^M$. 
Pairing any feasible $\varepsilon$ with any $\delta$ that satisfies this inequality suffices for a sound solution. Since we report the minimum $\delta$ consistent with the target guarantee, we conclude that $\pi_{\mathrm B}$ is sound.

To establish completeness, we must show that the PAC parameters returned by $\pi_{\mathrm B}$ form a Pareto frontier of admissible solutions, i.e. that no possible $(\varepsilon, \delta)$ pair dominates those returned by the procedure. Since our $\bm \varepsilon$ set covers the entire feasible region of values for the tolerance parameter, this amounts to demonstrating the optimality of the confidence function:
\begin{align*}
    \delta(\hat p, \varepsilon) = \big( 1 - \hat p/(1 - \varepsilon) \big)^M.
\end{align*}
This follows immediately from the i.i.d. sampling procedure itself, which ensures that the probability of missing the $\varepsilon$-superlevel set is exactly $\big( 1 - \mu_\varepsilon \big)^M$. 
Since $\hat p / (1 - \varepsilon) \leq \mu_\varepsilon$, any $\delta' < \delta(\hat p, \varepsilon)$ would fail to satisfy the PAC guarantee in the worst case, e.g. if just a single atom's mass exceeds $\hat p / (1 - \varepsilon)$ by an infinitesimal amount.  
Thus $\delta(\hat p, \varepsilon)$ represents the infimum of all valid confidence levels, completing the proof. $\Box$
\\

\noindent \textbf{Lemma 2} (Identifiability)\textbf{.} Let $\mathcal Q \subseteq \mathbb R^{d}$ and define $\mathcal B(\bm q, r) = \{\bm q' \in \mathcal Q: \lVert \bm q' - \bm q \rVert \leq r\}$ as a Euclidean ball of radius $r$ with center $\bm q$.
Then $\bm q^*$ is PAC-identifiable iff there exists some $\bm q \in G_\varepsilon$ and $r>0$ such that (a) $\mathcal B(\bm q, r) \subseteq G_\varepsilon$; and (b) $\mu\big(\mathcal B(\bm q, r)\big) > 0$.

\paragraph{Proof} To prove the lemma, we must show that conditions (a) and (b) are necessary and sufficient for PAC-identifiability in the continuous case. Recall that $\bm q^*$ is PAC-identifiable iff there exists a finite $M$ such that we draw at least one sample from the superlevel set $G_\varepsilon$ with probability at least $1 - \delta$. 

Take sufficiency first. If there exists some $\mathcal B(\bm q, r) \subseteq G_\varepsilon$ with strictly positive measure $\mu\big(\mathcal B(\bm q, r)\big) = \gamma > 0$, then by the monotonicity of measures, $\mu(G_\varepsilon) \geq \gamma$. We can now calculate a concrete stopping time:
\begin{align*}
    M \geq \frac{\log \delta^{-1}}{\log (1 - \gamma)^{-1}}. 
\end{align*}
Since $\gamma>0$, the denominator is finite and positive. Thus there exists a finite $M$ that satisfies the PAC criteria.

Now consider the necessity of (a) and (b). The probability of hitting $G_\varepsilon$ in $M$ trials is $1 - \big( 1 - \mu(G_\varepsilon)\big)^M$. For this value to be greater than $0$ (a necessary condition for identifiability), we must have $\mu(G_\varepsilon) > 0$. This integral is strictly positive iff $G_\varepsilon$ contains at least one subset of nonzero Lebesgue measure. Thus any PAC-identifiable atom $\bm q^*$ in continuous space must satisfy criteria (a) and (b). $\Box$
\\

\section{B: Experiments}
Further details on experiments, the benchmark on PGMs, and additional results on runtime are provided in this section. Complete code for reproducing all experiments is included in the code supplement.
Experiments with SPNs were run on a high performance computing cluster with 12 AMD EPYC 7282 CPUs, 64GB of RAM, an NVIDIA 130 graphics card and Ubuntu 22.04.5 LTS operating system.
Experiments with PGMs were run on a computer with a 2.3 GHz 8-Core Intel Core i9 processor, 16GB of RAM and macOS Sonoma 14.7.6 operating system. 

\subsection{Illustration}
\begin{figure}[t]
    \centering
    \includegraphics[width=0.95\linewidth]{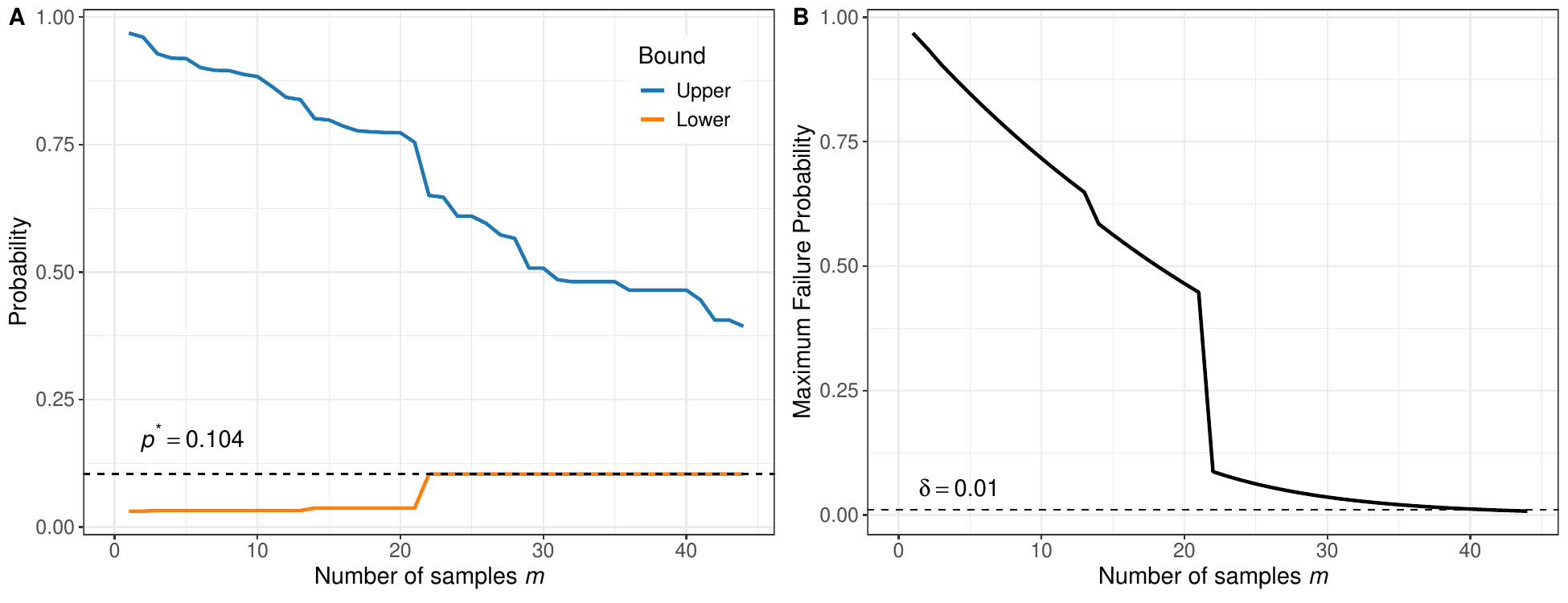}
    \caption{Schematic illustration of purely random PAC-MAP, Alg. $\pi_{\mathrm R}$. (\textbf{A}) Lower and upper bounds approach the MAP probability $p^*$ as $m$ increases. (\textbf{B}) Maximum failure probability approaches $\delta$ as $m$ increases.}
    \label{fig:schematic}
\end{figure}

As a simple proof of concept, we run purely random PAC-MAP $\pi_{\mathrm R}$ on a simulated dataset with $n=6$ binary query variables and a ground truth MAP probability of $p^*=0.104$ (see Fig. \ref{fig:schematic}). PAC parameters are set at $\varepsilon = \delta = 0.01$. Panel (A) monitors the deterministic stopping condition, which would be triggered if the upper bound $\check p$ were to fall to within a factor $1 - \varepsilon$ of the lower bound $\hat p$.
Panel (B) monitors the probabilistic stopping condition, which is triggered when the maximum failure probability hits $\delta$.
In this experiment, the sampler finds the true MAP in $m=22$ samples, but cannot issue a PAC certificate until time $M=44$. Though the solution in this case happens to be exact, the algorithm only guarantees that it is PAC.
For this experiment, we sample $2^6=64$ draws from an exponential distribution, normalized to sum to unity. 
These represent the pmf over a hypothetical Boolean state space, with true MAP probability identified via a simple sort.

\subsection{Benchmark Experiment}
To compare the performance of \texttt{PAC-MAP} and \texttt{smooth-PAC-MAP} against current leading methods from the literature, we designed a benchmark experiment using the Twenty Datasets collection of binary datasets. The datasets, and SPNs that were trained on them, are in the code supplement.
We randomly partition the features into query and evidence variables, with the query proportion taking a value in $\{20\%, 50\%\}$. 
Assignments for evidence variables are drawn randomly.
We repeat the process ten times per dataset and query proportion, resulting in $600$ trials per MAP solver.
We use the following parameters in the majority of cases for \texttt{PAC-MAP} and \texttt{smooth-PAC-MAP}:
\begin{itemize}[noitemsep]
    \item \texttt{batch\_size}$=250$
    \item $\varepsilon=0.01$
    \item $\delta=0.01$
    \item \texttt{sample\_cap}$=2.5 \times10^6$
\end{itemize}
For \texttt{smooth-PAC-MAP}, we search a Hamming ball of radius $1$ around leading candidates every $250$ steps (i.e. after each batch). These $n$ candidates are not added to the sample counter $m$, which only applies to random draws. 
Preliminary trials with larger PAC parameters led to somewhat mixed results, though we did not systematically investigate this.
We tested alternative values of \texttt{batch\_size} ($1000$, $2500$) but found that $250$ produced the best results without increasing runtime too much. Finally, we chose a \texttt{sample\_cap} of $2.5 \times 10^6$ as larger values were computationally impractical. For \texttt{HBP}, we chose a \texttt{num\_iterations} parameter of $5$ to prevent runtime on large datasets from becoming excessive, and the defaults for all others. For \texttt{ITSELF}, we chose three different sets of \texttt{hidden\_sizes} parameters when training networks (that were simple multi-layer perceptrons) on each dataset, depending on the datasets size:  
\begin{itemize}
    \item Small (msnbc, nltcs): [$64, 32$]
    \item Large (book, msweb, nips, tmovie): [$256, 128$]
    \item Medium (all others): [$128$, $64$]
\end{itemize}
We then chose the following parameters that were shared across all datasets:
\begin{itemize}
    \item \texttt{embedding}: '\texttt{DiscreteEmbedder}'
    \item \texttt{loss}: '\texttt{mpe\_log\_likelihood\_loss}'
    \item \texttt{learning\_rate}: $0.001$
    \item \texttt{optimizer}: Adam
    \item \texttt{batch\_size}: $16$
    \item \texttt{n\_train}: $256$
    \item \texttt{n\_epochs}: $50$
\end{itemize}
These parameters were chosen to prevent experiment runtimes exceeding HPC cluster allowances. The other methods in the benchmark (\texttt{AMP}, \texttt{MP} and \texttt{IND}) do not have any hyperparameters to select. The randomly generated query/evidence partitions and value assignments used in each trial are included in the code supplement, ensuring exact replication of the experimental conditions. The stochastic search procedures of \texttt{PAC-MAP} and \texttt{smooth-PAC-MAP} are not seeded across runs, so exact sample trajectories may differ, though this does not affect the reported statistics, which are averaged over ten trials with standard errors. 

For the benchmark on the PGMs, we compare against four approximate MAP solvers:
\begin{itemize}[noitemsep]
    \item \texttt{Distributed And-Or Branch-And-Bound (DAOOPT-AOBB)}, a parallelized version of And-Or Brand-And-Bound search described in \citet{otten2017and} and implemented in the Distributed And-Or Optimization (DAOOPT) package \cite{daooptgithub}.
    \item \texttt{Distributed And-Or Best First Search (DAOOPT-AOBF)}, a parallelized version of And-Or Best-First-Search, also implemented in DAOOPT. 
    \item \texttt{Iterative Join-Graph Propagation (IJGP)}, a method that combines bounded inference with iterative message-passing, described in \citet{dechter2012iterativejoingraphpropagation} and implemented in the Merlin package \cite{merlingithub}.
    \item \texttt{Weighted Mini-Bucket Elimination (WMB)}, a method that uses Hölder’s inequality to compute upper and lower bounds, described in \citet{liu2011bounding} and also implemented in Merlin.
\end{itemize}
Again, we refer readers to the original texts for details on these methods. On large datasets and higher query proportions, these methods would attempt to allocate extremely large amounts of memory, requiring us to remove the four largest datasets from this benchmark and restrict the largest query proportion setting to $20\%$. Several of them also did not natively support MMAP queries, so the benchmark was restricted to the MAP task. The results are shown in Table \ref{tab:PGM benchmark}. Due to the aforementioned out-of-memory errors, results could not be recorded for the experiments on \texttt{dna}, \texttt{kosarek} and \texttt{pumsb\_star} in the higher query proportion setting. Both PAC-MAP algorithms out-performed the competitors across both dimensions however, with $\pi_{\mathrm S}$ dominating $\pi_{\mathrm R}$ again, demonstrating the effectiveness of our algorithms across multiple model classes and competitiveness with the state-of-the-art in each of these settings. We used the following hyperparameters for each competitor method: 
\begin{itemize}
    \item \texttt{WMB}:
    \begin{itemize}
        \item \texttt{ibound}: $10$
        \item \texttt{iterations}: $10$
        \item \texttt{timeout}: $600$
    \end{itemize}
    \item \texttt{IJGP}
    \begin{itemize}
        \item \texttt{ibound}: $10$
        \item \texttt{iterations}: $10$
        \item \texttt{timeout}: $600$
    \end{itemize}
    \item \texttt{DAOOPT-AOBB}:
    \begin{itemize}
        \item \texttt{timeout}: $600$
    \end{itemize}
    \item \texttt{DAOOPT-AOBF}:
    \begin{itemize}
        \item \texttt{timeout}: $600$
    \end{itemize}
\end{itemize}
These were chosen to balance MAP solution quality with computational practicality. Details on the software libraries used to implement each method for both benchmarks is given in table \ref{tab:packages}.

\begin{table}[h]
\centering
\caption{Software packages and versions used for competitor methods. Independent was self-implemented and will be included in the code supplement. Links to the pyspn repo, merlin repo and daoopt repo will be provide after de-anonymisation.
Python 3.8.20 was used to run the \texttt{Independent} and pyspn methods (as well as our own \texttt{PACMAP} methods). R 4.4.3 was used to run .R files in the code supplement, such as files to plot Figs. 1 and 2 in the main text.
Python 3.13.5 was used to run NeuPI. Clang version 16.0.0 was used to compile C++ packages merlin and daoopt.}
\label{tab:packages}
\setlength{\tabcolsep}{1mm}
\begin{tabular}{lll}
\toprule
Method & Package & Version \\
\midrule
\texttt{Independent}    & --      & --                                      \\
\texttt{MaxProduct}     & pyspn   & fork from repo @ 31/05/20   \\
\texttt{ArgMaxProduct}  & pyspn   & fork from repo @ 31/05/20   \\
\texttt{HBP}            & pyspn   & fork from repo @ 31/05/20   \\
\texttt{ITSELF}         & NeuPI   & 1.1.0                                   \\
\texttt{WMB}            & merlin  & fork from repo @ 9/10/25    \\
\texttt{IJGP}           & merlin  & fork from repo @ 9/10/25    \\
\texttt{DAOOPT-AOBB}    & daoopt  & 1.1.2                                   \\
\texttt{DAOOPT-AOBF}    & daoopt  & 1.1.2                                   \\
\bottomrule
\end{tabular}
\end{table}

\begin{table}
\caption{Average rankings of methods by MAP estimates on CLTs trained on each dataset, across ten trials per dataset.}
\setlength{\linewidth}{\textwidth}
\setlength{\hsize}{\textwidth}
\scriptsize
\definecolor{npgRed}{HTML}{E64B35}
\definecolor{npgBlue}{HTML}{4DBBD5}
\definecolor{npgGreen}{HTML}{00A087}
\definecolor{npgNavy}{HTML}{3C5488}
\definecolor{npgPurple}{HTML}{8491B4}
\definecolor{npgOrange}{HTML}{F39B7F}
\definecolor{npgBrown}{HTML}{7E6148}
\setlength{\tabcolsep}{1pt}
\begin{tabular}{l|@{\hspace{0.2em}} c|@{\hspace{0.2em}} c @{\hspace{0.2em}}}
\toprule
Dataset - Dimension & 10\% Q 90\% E & 20\% Q 80\% E \\
\midrule
\texttt{accidents - 111} & \mnum{$\textcolor{npgBrown}{4.4}$}\msep\mnum{$\textcolor{npgOrange}{4.4}$}\msep\mnum{$\textcolor{npgPurple}{3.6}$}\msep\mnum{$\textcolor{npgGreen}{1.2}$}\msep\mnum{$\textcolor{npgNavy}{\mathbf{1.1}}$}\msep\mnum{$\textcolor{npgRed}{3.6}$} & \mnum{$\textcolor{npgBrown}{3.6}$}\msep\mnum{$\textcolor{npgOrange}{3.6}$}\msep\mnum{$\textcolor{npgPurple}{4.4}$}\msep\mnum{$\textcolor{npgGreen}{1.5}$}\msep\mnum{$\textcolor{npgNavy}{\mathbf{1.0}}$}\msep\mnum{$\textcolor{npgRed}{4.4}$} \\
\texttt{adult - 123} & \mnum{$\textcolor{npgBrown}{4.2}$}\msep\mnum{$\textcolor{npgOrange}{4.2}$}\msep\mnum{$\textcolor{npgPurple}{3.8}$}\msep\mnum{$\textcolor{npgGreen}{1.3}$}\msep\mnum{$\textcolor{npgNavy}{\mathbf{1.0}}$}\msep\mnum{$\textcolor{npgRed}{3.8}$} & \mnum{$\textcolor{npgBrown}{4.4}$}\msep\mnum{$\textcolor{npgOrange}{4.4}$}\msep\mnum{$\textcolor{npgPurple}{3.6}$}\msep\mnum{$\textcolor{npgGreen}{1.7}$}\msep\mnum{$\textcolor{npgNavy}{\mathbf{1.0}}$}\msep\mnum{$\textcolor{npgRed}{3.6}$} \\
\texttt{baudio - 100} & \mnum{$\textcolor{npgBrown}{4.8}$}\msep\mnum{$\textcolor{npgOrange}{4.8}$}\msep\mnum{$\textcolor{npgPurple}{3.2}$}\msep\mnum{$\textcolor{npgGreen}{1.2}$}\msep\mnum{$\textcolor{npgNavy}{\mathbf{1.1}}$}\msep\mnum{$\textcolor{npgRed}{3.2}$} & \mnum{$\textcolor{npgBrown}{4.8}$}\msep\mnum{$\textcolor{npgOrange}{4.8}$}\msep\mnum{$\textcolor{npgPurple}{3.2}$}\msep\mnum{$\textcolor{npgGreen}{1.8}$}\msep\mnum{$\textcolor{npgNavy}{\mathbf{1.0}}$}\msep\mnum{$\textcolor{npgRed}{3.2}$} \\
\texttt{bnetflix - 100} & \mnum{$\textcolor{npgBrown}{4.0}$}\msep\mnum{$\textcolor{npgOrange}{4.0}$}\msep\mnum{$\textcolor{npgPurple}{4.0}$}\msep\mnum{$\textcolor{npgGreen}{\mathbf{1.0}}$}\msep\mnum{$\textcolor{npgNavy}{\mathbf{1.0}}$}\msep\mnum{$\textcolor{npgRed}{4.0}$} & \mnum{$\textcolor{npgBrown}{4.0}$}\msep\mnum{$\textcolor{npgOrange}{4.0}$}\msep\mnum{$\textcolor{npgPurple}{4.0}$}\msep\mnum{$\textcolor{npgGreen}{1.3}$}\msep\mnum{$\textcolor{npgNavy}{\mathbf{1.0}}$}\msep\mnum{$\textcolor{npgRed}{4.0}$} \\
\texttt{connect4 - 126} & \mnum{$\textcolor{npgBrown}{4.8}$}\msep\mnum{$\textcolor{npgOrange}{4.8}$}\msep\mnum{$\textcolor{npgPurple}{3.0}$}\msep\mnum{$\textcolor{npgGreen}{1.9}$}\msep\mnum{$\textcolor{npgNavy}{\mathbf{1.0}}$}\msep\mnum{$\textcolor{npgRed}{3.0}$} & \mnum{$\textcolor{npgBrown}{4.4}$}\msep\mnum{$\textcolor{npgOrange}{4.4}$}\msep\mnum{$\textcolor{npgPurple}{3.5}$}\msep\mnum{$\textcolor{npgGreen}{2.2}$}\msep\mnum{$\textcolor{npgNavy}{\mathbf{1.0}}$}\msep\mnum{$\textcolor{npgRed}{3.5}$} \\
\texttt{dna - 180} & \mnum{$\textcolor{npgBrown}{4.8}$}\msep\mnum{$\textcolor{npgOrange}{4.8}$}\msep\mnum{$\textcolor{npgPurple}{3.2}$}\msep\mnum{$\textcolor{npgGreen}{1.2}$}\msep\mnum{$\textcolor{npgNavy}{\mathbf{1.0}}$}\msep\mnum{$\textcolor{npgRed}{3.2}$} & \mnum{--} \\
\texttt{jester - 100} & \mnum{$\textcolor{npgBrown}{4.2}$}\msep\mnum{$\textcolor{npgOrange}{4.2}$}\msep\mnum{$\textcolor{npgPurple}{3.8}$}\msep\mnum{$\textcolor{npgGreen}{\mathbf{1.0}}$}\msep\mnum{$\textcolor{npgNavy}{\mathbf{1.0}}$}\msep\mnum{$\textcolor{npgRed}{3.8}$} & \mnum{$\textcolor{npgBrown}{4.2}$}\msep\mnum{$\textcolor{npgOrange}{4.2}$}\msep\mnum{$\textcolor{npgPurple}{3.8}$}\msep\mnum{$\textcolor{npgGreen}{1.1}$}\msep\mnum{$\textcolor{npgNavy}{\mathbf{1.0}}$}\msep\mnum{$\textcolor{npgRed}{3.8}$} \\
\texttt{kdd - 64} & \mnum{$\textcolor{npgBrown}{4.3}$}\msep\mnum{$\textcolor{npgOrange}{4.3}$}\msep\mnum{$\textcolor{npgPurple}{3.5}$}\msep\mnum{$\textcolor{npgGreen}{1.5}$}\msep\mnum{$\textcolor{npgNavy}{\mathbf{1.0}}$}\msep\mnum{$\textcolor{npgRed}{3.5}$} & \mnum{$\textcolor{npgBrown}{4.4}$}\msep\mnum{$\textcolor{npgOrange}{4.4}$}\msep\mnum{$\textcolor{npgPurple}{3.6}$}\msep\mnum{$\textcolor{npgGreen}{\mathbf{1.1}}$}\msep\mnum{$\textcolor{npgNavy}{\mathbf{1.1}}$}\msep\mnum{$\textcolor{npgRed}{3.6}$} \\
\texttt{kosarek - 190} & \mnum{$\textcolor{npgBrown}{4.8}$}\msep\mnum{$\textcolor{npgOrange}{4.8}$}\msep\mnum{$\textcolor{npgPurple}{3.2}$}\msep\mnum{$\textcolor{npgGreen}{1.7}$}\msep\mnum{$\textcolor{npgNavy}{\mathbf{1.0}}$}\msep\mnum{$\textcolor{npgRed}{3.2}$} & \mnum{--} \\
\texttt{msnbc - 17} & \mnum{$\textcolor{npgBrown}{2.4}$}\msep\mnum{$\textcolor{npgOrange}{2.4}$}\msep\mnum{$\textcolor{npgPurple}{1.6}$}\msep\mnum{$\textcolor{npgGreen}{\mathbf{1.0}}$}\msep\mnum{$\textcolor{npgNavy}{\mathbf{1.0}}$}\msep\mnum{$\textcolor{npgRed}{1.6}$} & \mnum{$\textcolor{npgBrown}{3.8}$}\msep\mnum{$\textcolor{npgOrange}{3.8}$}\msep\mnum{$\textcolor{npgPurple}{2.6}$}\msep\mnum{$\textcolor{npgGreen}{\mathbf{1.0}}$}\msep\mnum{$\textcolor{npgNavy}{\mathbf{1.0}}$}\msep\mnum{$\textcolor{npgRed}{2.6}$} \\
\texttt{mushrooms - 112} & \mnum{$\textcolor{npgBrown}{4.2}$}\msep\mnum{$\textcolor{npgOrange}{4.2}$}\msep\mnum{$\textcolor{npgPurple}{3.8}$}\msep\mnum{$\textcolor{npgGreen}{1.2}$}\msep\mnum{$\textcolor{npgNavy}{\mathbf{1.0}}$}\msep\mnum{$\textcolor{npgRed}{3.8}$} & \mnum{$\textcolor{npgBrown}{4.0}$}\msep\mnum{$\textcolor{npgOrange}{4.0}$}\msep\mnum{$\textcolor{npgPurple}{4.0}$}\msep\mnum{$\textcolor{npgGreen}{1.8}$}\msep\mnum{$\textcolor{npgNavy}{\mathbf{1.0}}$}\msep\mnum{$\textcolor{npgRed}{4.0}$} \\
\texttt{nltcs - 16} & \mnum{$\textcolor{npgBrown}{1.8}$}\msep\mnum{$\textcolor{npgOrange}{1.8}$}\msep\mnum{$\textcolor{npgPurple}{1.8}$}\msep\mnum{$\textcolor{npgGreen}{\mathbf{1.0}}$}\msep\mnum{$\textcolor{npgNavy}{\mathbf{1.0}}$}\msep\mnum{$\textcolor{npgRed}{1.8}$} & \mnum{$\textcolor{npgBrown}{3.8}$}\msep\mnum{$\textcolor{npgOrange}{3.8}$}\msep\mnum{$\textcolor{npgPurple}{3.4}$}\msep\mnum{$\textcolor{npgGreen}{\mathbf{1.0}}$}\msep\mnum{$\textcolor{npgNavy}{\mathbf{1.0}}$}\msep\mnum{$\textcolor{npgRed}{3.4}$} \\
\texttt{ocrletters - 128} & \mnum{$\textcolor{npgBrown}{4.0}$}\msep\mnum{$\textcolor{npgOrange}{4.0}$}\msep\mnum{$\textcolor{npgPurple}{4.0}$}\msep\mnum{$\textcolor{npgGreen}{1.1}$}\msep\mnum{$\textcolor{npgNavy}{\mathbf{1.0}}$}\msep\mnum{$\textcolor{npgRed}{4.0}$} & \mnum{$\textcolor{npgBrown}{4.6}$}\msep\mnum{$\textcolor{npgOrange}{4.6}$}\msep\mnum{$\textcolor{npgPurple}{3.4}$}\msep\mnum{$\textcolor{npgGreen}{1.7}$}\msep\mnum{$\textcolor{npgNavy}{\mathbf{1.1}}$}\msep\mnum{$\textcolor{npgRed}{3.4}$} \\
\texttt{plants - 69} & \mnum{$\textcolor{npgBrown}{4.0}$}\msep\mnum{$\textcolor{npgOrange}{4.0}$}\msep\mnum{$\textcolor{npgPurple}{3.8}$}\msep\mnum{$\textcolor{npgGreen}{1.5}$}\msep\mnum{$\textcolor{npgNavy}{\mathbf{1.0}}$}\msep\mnum{$\textcolor{npgRed}{3.8}$} & \mnum{$\textcolor{npgBrown}{3.6}$}\msep\mnum{$\textcolor{npgOrange}{3.6}$}\msep\mnum{$\textcolor{npgPurple}{4.4}$}\msep\mnum{$\textcolor{npgGreen}{1.6}$}\msep\mnum{$\textcolor{npgNavy}{\mathbf{1.0}}$}\msep\mnum{$\textcolor{npgRed}{4.4}$} \\
\texttt{pumsb\_star - 163} & \mnum{$\textcolor{npgBrown}{4.6}$}\msep\mnum{$\textcolor{npgOrange}{4.6}$}\msep\mnum{$\textcolor{npgPurple}{3.4}$}\msep\mnum{$\textcolor{npgGreen}{1.4}$}\msep\mnum{$\textcolor{npgNavy}{\mathbf{1.0}}$}\msep\mnum{$\textcolor{npgRed}{3.4}$} & \mnum{--} \\
\texttt{tretail - 135} & \mnum{$\textcolor{npgBrown}{4.2}$}\msep\mnum{$\textcolor{npgOrange}{4.2}$}\msep\mnum{$\textcolor{npgPurple}{3.8}$}\msep\mnum{$\textcolor{npgGreen}{1.1}$}\msep\mnum{$\textcolor{npgNavy}{\mathbf{1.0}}$}\msep\mnum{$\textcolor{npgRed}{3.8}$} & \mnum{$\textcolor{npgBrown}{5.0}$}\msep\mnum{$\textcolor{npgOrange}{5.0}$}\msep\mnum{$\textcolor{npgPurple}{3.0}$}\msep\mnum{$\textcolor{npgGreen}{1.1}$}\msep\mnum{$\textcolor{npgNavy}{\mathbf{1.0}}$}\msep\mnum{$\textcolor{npgRed}{3.0}$} \\
\midrule
\quad No. Ranked Highest:& \mnum{$\textcolor{npgBrown}{0}$}\msep\mnum{$\textcolor{npgOrange}{0}$}\msep\mnum{$\textcolor{npgPurple}{0}$}\msep\mnum{$\textcolor{npgGreen}{4}$}\msep\mnum{$\textcolor{npgNavy}{16}$}\msep\mnum{$\textcolor{npgRed}{0}$} & \mnum{$\textcolor{npgBrown}{0}$}\msep\mnum{$\textcolor{npgOrange}{0}$}\msep\mnum{$\textcolor{npgPurple}{0}$}\msep\mnum{$\textcolor{npgGreen}{3}$}\msep\mnum{$\textcolor{npgNavy}{13}$}\msep\mnum{$\textcolor{npgRed}{0}$}\\
\bottomrule
\end{tabular}
\begin{flushleft}
\scriptsize
\textbf{Method:} \quad
\textcolor{npgBrown}{DAOOPT-AOBB} \quad
\textcolor{npgOrange}{DAOOPT-AOBF} \quad
\textcolor{npgPurple}{IJGP} \quad
\textcolor{npgGreen}{PACMAP} \\
\quad \quad \quad \quad \quad \textcolor{npgNavy}{smooth-PACMAP} \quad
\textcolor{npgRed}{WMB} \quad
\end{flushleft}
\label{tab:PGM benchmark}
\end{table}


\subsection{Warm Starts}
For this experiment, we initialize \texttt{smooth-PAC-MAP} with the estimated MAP solution of \texttt{AMP} to derive theoretical guarantees and possibly find a more probable configuration. 
We fix the query proportion at $25\%$ and use the same randomly generated MAP queries from the benchmark experiment for each dataset, along with the same hyperparameters for our procedure.

\subsection{Runtime Experiments}
In this section, we report average runtimes for each algorithm over ten trials at given query proportions. 
We find that \texttt{MP} is consistently fastest, while \texttt{AMP} takes longer on average. Since the complexity of the former is linear in circuit size, while the latter is quadratic, this result accords with expectations. \texttt{Ind} takes somewhat longer in all trials, though the procedure could likely be optimized.
Our random solvers consistently take the longest to execute, reflecting the overhead required for theoretical guarantees. 

\definecolor{npgRed}{HTML}{E64B35}
\definecolor{npgBlue}{HTML}{4DBBD5}
\definecolor{npgGreen}{HTML}{00A087}
\definecolor{npgNavy}{HTML}{3C5488}
\definecolor{npgPurple}{HTML}{8491B4}
\definecolor{npgOrange}{HTML}{F39B7F}
\definecolor{npgBrown}{HTML}{7E6148}

\begin{table*}[t]
\centering
\scriptsize
\setlength{\tabcolsep}{2pt}
\renewcommand{\arraystretch}{0.96}
\caption{Runtime comparison (seconds) for 20\% query and 80\% evidence.  We report the mean ± SE over ten trials.}
\label{tab:runtimes1}
\begin{tabular}{@{}lccccccc@{}}
\toprule
Dataset & AMP & IND & MP & HBP & ITSELF & PAC-MAP & smooth-PAC-MAP \\
\midrule
\texttt{accidents} & $\textcolor{npgRed}{4.51} \pm 0.619$ & $\textcolor{npgPurple}{14.9} \pm 0.411$ & $\textcolor{npgBlue}{\bm{0.525}} \pm 0.031$ & $\textcolor{npgBrown}{17} \pm 0.484$ & $\textcolor{npgOrange}{410} \pm 0$ & $\textcolor{npgGreen}{47.3} \pm 26.4$ & $\textcolor{npgNavy}{50.6} \pm 28.8$ \\
\texttt{adult} & $\textcolor{npgRed}{0.437} \pm 0.0737$ & $\textcolor{npgPurple}{2.16} \pm 0.0663$ & $\textcolor{npgBlue}{\bm{0.0683}} \pm 0.00323$ & $\textcolor{npgBrown}{1.14} \pm 0.0409$ & $\textcolor{npgOrange}{42.4} \pm 0$ & $\textcolor{npgGreen}{1.23} \pm 0.828$ & $\textcolor{npgNavy}{1.25} \pm 0.0874$ \\
\texttt{baudio} & $\textcolor{npgRed}{15.7} \pm 3.16$ & $\textcolor{npgPurple}{22.8} \pm 0.502$ & $\textcolor{npgBlue}{\bm{0.889}} \pm 0.0433$ & $\textcolor{npgBrown}{13.8} \pm 0.29$ & $\textcolor{npgOrange}{574} \pm 0$ & $\textcolor{npgGreen}{372} \pm 218$ & $\textcolor{npgNavy}{380} \pm 218$ \\
\texttt{bnetflix} & $\textcolor{npgRed}{4.67} \pm 0.781$ & $\textcolor{npgPurple}{16.1} \pm 0.668$ & $\textcolor{npgBlue}{\bm{0.656}} \pm 0.113$ & $\textcolor{npgBrown}{6.75} \pm 0.177$ & $\textcolor{npgOrange}{438} \pm 0$ & $\textcolor{npgGreen}{181} \pm 93.4$ & $\textcolor{npgNavy}{190} \pm 100$ \\
\texttt{book} & $\textcolor{npgRed}{13.7} \pm 3.3$ & $\textcolor{npgPurple}{106} \pm 5.19$ & $\textcolor{npgBlue}{\bm{0.92}} \pm 0.113$ & $\textcolor{npgBrown}{4.47} \pm 0.245$ & $\textcolor{npgOrange}{507} \pm 0$ & $\textcolor{npgGreen}{13039} \pm 1430$ & $\textcolor{npgNavy}{13754} \pm 969$ \\
\texttt{connect4} & $\textcolor{npgRed}{6.42} \pm 0.785$ & $\textcolor{npgPurple}{28.6} \pm 0.409$ & $\textcolor{npgBlue}{\bm{0.877}} \pm 0.03$ & $\textcolor{npgBrown}{13.4} \pm 0.145$ & $\textcolor{npgOrange}{711} \pm 0$ & $\textcolor{npgGreen}{11} \pm 5.52$ & $\textcolor{npgNavy}{14.2} \pm 7.03$ \\
\texttt{dna} & $\textcolor{npgRed}{2.45} \pm 0.465$ & $\textcolor{npgPurple}{8.79} \pm 0.174$ & $\textcolor{npgBlue}{\bm{0.194}} \pm 0.0101$ & $\textcolor{npgBrown}{0.76} \pm 0.0677$ & $\textcolor{npgOrange}{80.3} \pm 0$ & $\textcolor{npgGreen}{3203} \pm 530$ & $\textcolor{npgNavy}{3347} \pm 405$ \\
\texttt{jester} & $\textcolor{npgRed}{9.45} \pm 2.08$ & $\textcolor{npgPurple}{11.4} \pm 0.275$ & $\textcolor{npgBlue}{\bm{0.453}} \pm 0.00701$ & $\textcolor{npgBrown}{4.07} \pm 0.144$ & $\textcolor{npgOrange}{288} \pm 0$ & $\textcolor{npgGreen}{354} \pm 450$ & $\textcolor{npgNavy}{354} \pm 433$ \\
\texttt{kdd} & $\textcolor{npgRed}{1.53} \pm 0.321$ & $\textcolor{npgPurple}{2.27} \pm 0.0639$ & $\textcolor{npgBlue}{\bm{0.158}} \pm 0.0362$ & $\textcolor{npgBrown}{0.902} \pm 0.0533$ & $\textcolor{npgOrange}{54} \pm 0$ & $\textcolor{npgGreen}{2.2} \pm 0.804$ & $\textcolor{npgNavy}{2.35} \pm 0.143$ \\
\texttt{kosarek} & $\textcolor{npgRed}{3.51} \pm 0.736$ & $\textcolor{npgPurple}{17.2} \pm 0.645$ & $\textcolor{npgBlue}{\bm{0.361}} \pm 0.0225$ & $\textcolor{npgBrown}{6.57} \pm 0.142$ & $\textcolor{npgOrange}{194} \pm 0$ & $\textcolor{npgGreen}{5091} \pm 630$ & $\textcolor{npgNavy}{5157} \pm 630$ \\
\texttt{msnbc} & $\textcolor{npgRed}{1.5} \pm 0.22$ & $\textcolor{npgPurple}{0.809} \pm 0.029$ & $\textcolor{npgBlue}{\bm{0.18}} \pm 0.00884$ & $\textcolor{npgBrown}{6.07} \pm 0.0993$ & $\textcolor{npgOrange}{126} \pm 0$ & $\textcolor{npgGreen}{1.79} \pm 0.719$ & $\textcolor{npgNavy}{1.71} \pm 0.235$ \\
\texttt{msweb} & $\textcolor{npgRed}{1.69} \pm 0.314$ & $\textcolor{npgPurple}{12.8} \pm 0.374$ & $\textcolor{npgBlue}{\bm{0.177}} \pm 0.00862$ & $\textcolor{npgBrown}{2.32} \pm 0.0748$ & $\textcolor{npgOrange}{72.3} \pm 0$ & $\textcolor{npgGreen}{3797} \pm 1047$ & $\textcolor{npgNavy}{3920} \pm 1063$ \\
\texttt{mushrooms} & $\textcolor{npgRed}{0.445} \pm 0.0653$ & $\textcolor{npgPurple}{2.17} \pm 0.0857$ & $\textcolor{npgBlue}{\bm{0.0774}} \pm 0.00425$ & $\textcolor{npgBrown}{0.804} \pm 0.05$ & $\textcolor{npgOrange}{35.5} \pm 0$ & $\textcolor{npgGreen}{1.6} \pm 0.848$ & $\textcolor{npgNavy}{1.55} \pm 0.0578$ \\
\texttt{nips} & $\textcolor{npgRed}{0.285} \pm 0.039$ & $\textcolor{npgPurple}{11.1} \pm 0.348$ & $\textcolor{npgBlue}{\bm{0.0864}} \pm 0.00481$ & $\textcolor{npgBrown}{0.515} \pm 0.00439$ & $\textcolor{npgOrange}{36.4} \pm 0$ & $\textcolor{npgGreen}{2409} \pm 183$ & $\textcolor{npgNavy}{2613} \pm 179$ \\
\texttt{nltcs} & $\textcolor{npgRed}{0.145} \pm 0.0167$ & $\textcolor{npgPurple}{0.119} \pm 0.00711$ & $\textcolor{npgBlue}{\bm{0.0255}} \pm 0.000876$ & $\textcolor{npgBrown}{0.189} \pm 0.0567$ & $\textcolor{npgOrange}{4.59} \pm 0$ & $\textcolor{npgGreen}{0.531} \pm 0.894$ & $\textcolor{npgNavy}{0.258} \pm 0.0308$ \\
\texttt{ocr\_letters} & $\textcolor{npgRed}{47.5} \pm 8.62$ & $\textcolor{npgPurple}{132} \pm 2.9$ & $\textcolor{npgBlue}{\bm{4.21}} \pm 0.163$ & -- & -- & $\textcolor{npgGreen}{8131} \pm 2672$ & $\textcolor{npgNavy}{9432} \pm 2950$ \\
\texttt{plants} & $\textcolor{npgRed}{5.6} \pm 0.973$ & $\textcolor{npgPurple}{10.8} \pm 0.244$ & $\textcolor{npgBlue}{\bm{0.649}} \pm 0.0312$ & $\textcolor{npgBrown}{38.2} \pm 0.974$ & $\textcolor{npgOrange}{437} \pm 0$ & $\textcolor{npgGreen}{11.3} \pm 4.1$ & $\textcolor{npgNavy}{12.3} \pm 3.94$ \\
\texttt{pumsb\_star} & $\textcolor{npgRed}{2.58} \pm 0.355$ & $\textcolor{npgPurple}{16.7} \pm 0.396$ & $\textcolor{npgBlue}{\bm{0.413}} \pm 0.0185$ & $\textcolor{npgBrown}{2.79} \pm 0.0643$ & $\textcolor{npgOrange}{353} \pm 0$ & $\textcolor{npgGreen}{14.9} \pm 7.85$ & $\textcolor{npgNavy}{16} \pm 7.41$ \\
\texttt{tmovie} & $\textcolor{npgRed}{12.3} \pm 2.96$ & $\textcolor{npgPurple}{142} \pm 8.24$ & $\textcolor{npgBlue}{\bm{1.19}} \pm 0.162$ & $\textcolor{npgBrown}{26.4} \pm 0.843$ & $\textcolor{npgOrange}{566} \pm 0$ & $\textcolor{npgGreen}{10517} \pm 401$ & $\textcolor{npgNavy}{10868} \pm 823$ \\
\texttt{tretail} & $\textcolor{npgRed}{0.121} \pm 0.0153$ & $\textcolor{npgPurple}{1.03} \pm 0.0237$ & $\textcolor{npgBlue}{\bm{0.0288}} \pm 0.00108$ & $\textcolor{npgBrown}{0.17} \pm 0.00282$ & $\textcolor{npgOrange}{7.71} \pm 0$ & $\textcolor{npgGreen}{1.16} \pm 0.951$ & $\textcolor{npgNavy}{1.02} \pm 0.0814$ \\
\bottomrule
\end{tabular}
\end{table*}

\definecolor{npgRed}{HTML}{E64B35}
\definecolor{npgBlue}{HTML}{4DBBD5}
\definecolor{npgGreen}{HTML}{00A087}
\definecolor{npgNavy}{HTML}{3C5488}
\definecolor{npgPurple}{HTML}{8491B4}
\definecolor{npgOrange}{HTML}{F39B7F}
\definecolor{npgBrown}{HTML}{7E6148}

\begin{table*}[t]
\centering
\scriptsize
\setlength{\tabcolsep}{2pt}
\renewcommand{\arraystretch}{0.96}
\caption{Runtime comparison (seconds) for 50\% query and 50\% evidence.  We report the mean ± SE over ten trials.}
\label{tab:runtimes2}
\begin{tabular}{@{}lccccccc@{}}
\toprule
Dataset & AMP & IND & MP & HBP & ITSELF & PAC-MAP & smooth-PAC-MAP \\
\midrule
\texttt{accidents} & $\textcolor{npgRed}{4.23} \pm 0.575$ & $\textcolor{npgPurple}{30.7} \pm 0.618$ & $\textcolor{npgBlue}{\bm{0.456}} \pm 0.0279$ & $\textcolor{npgBrown}{13.9} \pm 0.633$ & $\textcolor{npgOrange}{158} \pm 0$ & $\textcolor{npgGreen}{5353} \pm 819$ & $\textcolor{npgNavy}{4849} \pm 406$ \\
\texttt{adult} & $\textcolor{npgRed}{0.428} \pm 0.0704$ & $\textcolor{npgPurple}{4.98} \pm 0.176$ & $\textcolor{npgBlue}{\bm{0.0651}} \pm 0.00312$ & $\textcolor{npgBrown}{1.03} \pm 0.0202$ & $\textcolor{npgOrange}{91.3} \pm 0$ & $\textcolor{npgGreen}{4.79} \pm 3.27$ & $\textcolor{npgNavy}{5.31} \pm 3.85$ \\
\texttt{baudio} & $\textcolor{npgRed}{15.2} \pm 3.03$ & $\textcolor{npgPurple}{50.1} \pm 1.33$ & $\textcolor{npgBlue}{\bm{0.822}} \pm 0.0464$ & $\textcolor{npgBrown}{11.6} \pm 0.163$ & $\textcolor{npgOrange}{184} \pm 0$ & $\textcolor{npgGreen}{9670} \pm 1547$ & $\textcolor{npgNavy}{9579} \pm 996$ \\
\texttt{bnetflix} & $\textcolor{npgRed}{4.63} \pm 0.867$ & $\textcolor{npgPurple}{35.7} \pm 1.49$ & $\textcolor{npgBlue}{\bm{0.577}} \pm 0.0455$ & $\textcolor{npgBrown}{5.64} \pm 0.15$ & $\textcolor{npgOrange}{136} \pm 0$ & $\textcolor{npgGreen}{5016} \pm 607$ & $\textcolor{npgNavy}{5488} \pm 1053$ \\
\texttt{book} & $\textcolor{npgRed}{14.9} \pm 3.44$ & $\textcolor{npgPurple}{282} \pm 7.77$ & $\textcolor{npgBlue}{\bm{0.995}} \pm 0.0915$ & $\textcolor{npgBrown}{4.78} \pm 0.0386$ & $\textcolor{npgOrange}{245} \pm 0$ & $\textcolor{npgGreen}{11909} \pm 1525$ & $\textcolor{npgNavy}{11142} \pm 629$ \\
\texttt{connect4} & $\textcolor{npgRed}{6.12} \pm 0.742$ & $\textcolor{npgPurple}{62} \pm 1.13$ & $\textcolor{npgBlue}{\bm{0.803}} \pm 0.0353$ & $\textcolor{npgBrown}{11.6} \pm 0.149$ & $\textcolor{npgOrange}{198} \pm 0$ & $\textcolor{npgGreen}{72.3} \pm 41.3$ & $\textcolor{npgNavy}{78.8} \pm 40.9$ \\
\texttt{dna} & $\textcolor{npgRed}{2.35} \pm 0.427$ & $\textcolor{npgPurple}{19.6} \pm 0.441$ & $\textcolor{npgBlue}{\bm{0.184}} \pm 0.013$ & $\textcolor{npgBrown}{0.592} \pm 0.0903$ & $\textcolor{npgOrange}{39} \pm 0$ & $\textcolor{npgGreen}{2423} \pm 232$ & $\textcolor{npgNavy}{2632} \pm 293$ \\
\texttt{jester} & $\textcolor{npgRed}{9.31} \pm 2.02$ & $\textcolor{npgPurple}{25.2} \pm 0.764$ & $\textcolor{npgBlue}{\bm{0.422}} \pm 0.0329$ & $\textcolor{npgBrown}{3.37} \pm 0.0511$ & $\textcolor{npgOrange}{276} \pm 0$ & $\textcolor{npgGreen}{2049} \pm 62.8$ & $\textcolor{npgNavy}{2094} \pm 57.9$ \\
\texttt{kdd} & $\textcolor{npgRed}{1.51} \pm 0.312$ & $\textcolor{npgPurple}{5.6} \pm 0.164$ & $\textcolor{npgBlue}{\bm{0.146}} \pm 0.0167$ & $\textcolor{npgBrown}{0.725} \pm 0.0346$ & $\textcolor{npgOrange}{66.7} \pm 0$ & $\textcolor{npgGreen}{96.8} \pm 137$ & $\textcolor{npgNavy}{101} \pm 138$ \\
\texttt{kosarek} & $\textcolor{npgRed}{3.32} \pm 0.691$ & $\textcolor{npgPurple}{37} \pm 1.11$ & $\textcolor{npgBlue}{\bm{0.325}} \pm 0.0185$ & $\textcolor{npgBrown}{5.55} \pm 0.092$ & $\textcolor{npgOrange}{309} \pm 0$ & $\textcolor{npgGreen}{4153} \pm 211$ & $\textcolor{npgNavy}{4271} \pm 263$ \\
\texttt{msnbc} & $\textcolor{npgRed}{1.45} \pm 0.196$ & $\textcolor{npgPurple}{1.82} \pm 0.0843$ & $\textcolor{npgBlue}{\bm{0.173}} \pm 0.0155$ & $\textcolor{npgBrown}{5.19} \pm 0.109$ & $\textcolor{npgOrange}{38.2} \pm 0$ & $\textcolor{npgGreen}{2.42} \pm 0.214$ & $\textcolor{npgNavy}{3.2} \pm 0.35$ \\
\texttt{msweb} & $\textcolor{npgRed}{1.61} \pm 0.29$ & $\textcolor{npgPurple}{28.7} \pm 0.658$ & $\textcolor{npgBlue}{\bm{0.162}} \pm 0.0103$ & $\textcolor{npgBrown}{1.88} \pm 0.0393$ & $\textcolor{npgOrange}{38.7} \pm 0$ & $\textcolor{npgGreen}{2529} \pm 251$ & $\textcolor{npgNavy}{2800} \pm 293$ \\
\texttt{mushrooms} & $\textcolor{npgRed}{0.419} \pm 0.0568$ & $\textcolor{npgPurple}{4.85} \pm 0.0746$ & $\textcolor{npgBlue}{\bm{0.0715}} \pm 0.00942$ & $\textcolor{npgBrown}{0.683} \pm 0.0185$ & $\textcolor{npgOrange}{14.9} \pm 0$ & $\textcolor{npgGreen}{11.1} \pm 12.7$ & $\textcolor{npgNavy}{13.8} \pm 15.5$ \\
\texttt{nips} & $\textcolor{npgRed}{0.28} \pm 0.0382$ & $\textcolor{npgPurple}{24.7} \pm 0.739$ & $\textcolor{npgBlue}{\bm{0.0815}} \pm 0.00476$ & $\textcolor{npgBrown}{0.471} \pm 0.0145$ & $\textcolor{npgOrange}{17.8} \pm 0$ & $\textcolor{npgGreen}{1873} \pm 68.4$ & $\textcolor{npgNavy}{2864} \pm 113$ \\
\texttt{nltcs} & $\textcolor{npgRed}{0.136} \pm 0.0162$ & $\textcolor{npgPurple}{0.249} \pm 0.00997$ & $\textcolor{npgBlue}{\bm{0.0232}} \pm 0.000899$ & $\textcolor{npgBrown}{0.221} \pm 0.0745$ & $\textcolor{npgOrange}{6.69} \pm 0$ & $\textcolor{npgGreen}{0.633} \pm 0.0595$ & $\textcolor{npgNavy}{0.71} \pm 0.127$ \\
\texttt{ocr\_letters} & $\textcolor{npgRed}{44.6} \pm 8.32$ & $\textcolor{npgPurple}{291} \pm 6.72$ & $\textcolor{npgBlue}{\bm{3.77}} \pm 0.205$ & -- & $\textcolor{npgOrange}{1271} \pm 0$ & $\textcolor{npgGreen}{5885} \pm 223$ & $\textcolor{npgNavy}{7941} \pm 356$ \\
\texttt{plants} & $\textcolor{npgRed}{5.2} \pm 0.9$ & $\textcolor{npgPurple}{23.9} \pm 0.528$ & $\textcolor{npgBlue}{\bm{0.573}} \pm 0.0331$ & $\textcolor{npgBrown}{33.4} \pm 1.85$ & $\textcolor{npgOrange}{144} \pm 0$ & $\textcolor{npgGreen}{1576} \pm 2824$ & $\textcolor{npgNavy}{1570} \pm 2790$ \\
\texttt{pumsb\_star} & $\textcolor{npgRed}{2.42} \pm 0.338$ & $\textcolor{npgPurple}{36.6} \pm 0.592$ & $\textcolor{npgBlue}{\bm{0.37}} \pm 0.0172$ & $\textcolor{npgBrown}{2.23} \pm 0.0303$ & $\textcolor{npgOrange}{89.2} \pm 0$ & $\textcolor{npgGreen}{3001} \pm 3143$ & $\textcolor{npgNavy}{3097} \pm 3245$ \\
\texttt{tmovie} & $\textcolor{npgRed}{11} \pm 2.54$ & $\textcolor{npgPurple}{291} \pm 9.02$ & $\textcolor{npgBlue}{\bm{1.05}} \pm 0.103$ & $\textcolor{npgBrown}{21.9} \pm 0.518$ & $\textcolor{npgOrange}{279} \pm 0$ & $\textcolor{npgGreen}{2142} \pm 129$ & $\textcolor{npgNavy}{2256} \pm 144$ \\
\texttt{tretail} & $\textcolor{npgRed}{0.115} \pm 0.0153$ & $\textcolor{npgPurple}{2.21} \pm 0.0361$ & $\textcolor{npgBlue}{\bm{0.0267}} \pm 0.00154$ & $\textcolor{npgBrown}{0.19} \pm 0.105$ & $\textcolor{npgOrange}{7.43} \pm 0$ & $\textcolor{npgGreen}{28} \pm 50.9$ & $\textcolor{npgNavy}{30.5} \pm 54.5$ \\
\bottomrule
\end{tabular}
\end{table*}

\definecolor{npgRed}{HTML}{E64B35}
\definecolor{npgBlue}{HTML}{4DBBD5}
\definecolor{npgGreen}{HTML}{00A087}
\definecolor{npgNavy}{HTML}{3C5488}
\definecolor{npgPurple}{HTML}{8491B4}
\definecolor{npgOrange}{HTML}{F39B7F}
\definecolor{npgBrown}{HTML}{7E6148}

\begin{table*}[t]
\centering
\scriptsize
\setlength{\tabcolsep}{2pt}
\renewcommand{\arraystretch}{0.96}
\caption{Runtime comparison (seconds) for 20\% query and 50\% evidence.  We report the mean ± SE over ten trials.}
\label{tab:runtimes3}
\begin{tabular}{@{}lccccccc@{}}
\toprule
Dataset & AMP & IND & MP & HBP & ITSELF & PAC-MAP & smooth-PAC-MAP \\
\midrule
\texttt{accidents} & $\textcolor{npgRed}{4.35} \pm 0.591$ & $\textcolor{npgPurple}{12.7} \pm 0.112$ & $\textcolor{npgBlue}{\bm{0.49}} \pm 0.0164$ & $\textcolor{npgBrown}{15} \pm 0.227$ & -- & $\textcolor{npgGreen}{52.2} \pm 31.1$ & $\textcolor{npgNavy}{67.1} \pm 42.1$ \\
\texttt{adult} & $\textcolor{npgRed}{0.436} \pm 0.0713$ & $\textcolor{npgPurple}{2} \pm 0.0641$ & $\textcolor{npgBlue}{\bm{0.0693}} \pm 0.00336$ & $\textcolor{npgBrown}{1.07} \pm 0.045$ & -- & $\textcolor{npgGreen}{1.02} \pm 1.04$ & $\textcolor{npgNavy}{0.857} \pm 0.107$ \\
\texttt{baudio} & $\textcolor{npgRed}{15.5} \pm 3.07$ & $\textcolor{npgPurple}{20.4} \pm 0.529$ & $\textcolor{npgBlue}{\bm{0.879}} \pm 0.0479$ & $\textcolor{npgBrown}{12} \pm 0.147$ & -- & $\textcolor{npgGreen}{449} \pm 476$ & $\textcolor{npgNavy}{429} \pm 428$ \\
\texttt{bnetflix} & $\textcolor{npgRed}{4.68} \pm 0.789$ & $\textcolor{npgPurple}{14.3} \pm 0.493$ & $\textcolor{npgBlue}{\bm{0.607}} \pm 0.0347$ & $\textcolor{npgBrown}{5.67} \pm 0.0955$ & -- & $\textcolor{npgGreen}{283} \pm 155$ & $\textcolor{npgNavy}{265} \pm 140$ \\
\texttt{book} & $\textcolor{npgRed}{15} \pm 3.45$ & $\textcolor{npgPurple}{112} \pm 3.67$ & $\textcolor{npgBlue}{\bm{1.2}} \pm 0.0651$ & $\textcolor{npgBrown}{5.03} \pm 0.087$ & -- & $\textcolor{npgGreen}{10156} \pm 1374$ & $\textcolor{npgNavy}{8031} \pm 293$ \\
\texttt{connect4} & $\textcolor{npgRed}{6.15} \pm 0.78$ & $\textcolor{npgPurple}{24.4} \pm 0.438$ & $\textcolor{npgBlue}{\bm{0.843}} \pm 0.028$ & $\textcolor{npgBrown}{11.9} \pm 0.214$ & -- & $\textcolor{npgGreen}{7.65} \pm 0.647$ & $\textcolor{npgNavy}{9.4} \pm 0.761$ \\
\texttt{dna} & $\textcolor{npgRed}{2.38} \pm 0.448$ & $\textcolor{npgPurple}{7.82} \pm 0.13$ & $\textcolor{npgBlue}{\bm{0.192}} \pm 0.0153$ & $\textcolor{npgBrown}{0.606} \pm 0.0853$ & -- & $\textcolor{npgGreen}{2116} \pm 670$ & $\textcolor{npgNavy}{2888} \pm 897$ \\
\texttt{jester} & $\textcolor{npgRed}{9.17} \pm 1.97$ & $\textcolor{npgPurple}{9.88} \pm 0.219$ & $\textcolor{npgBlue}{\bm{0.431}} \pm 0.026$ & $\textcolor{npgBrown}{3.33} \pm 0.0341$ & -- & $\textcolor{npgGreen}{230} \pm 210$ & $\textcolor{npgNavy}{109} \pm 100$ \\
\texttt{kdd} & $\textcolor{npgRed}{1.49} \pm 0.308$ & $\textcolor{npgPurple}{2.09} \pm 0.0652$ & $\textcolor{npgBlue}{\bm{0.143}} \pm 0.00741$ & $\textcolor{npgBrown}{0.717} \pm 0.0439$ & -- & $\textcolor{npgGreen}{1.57} \pm 0.441$ & $\textcolor{npgNavy}{1.84} \pm 0.601$ \\
\texttt{kosarek} & $\textcolor{npgRed}{3.35} \pm 0.693$ & $\textcolor{npgPurple}{14.7} \pm 0.436$ & $\textcolor{npgBlue}{\bm{0.349}} \pm 0.022$ & $\textcolor{npgBrown}{5.67} \pm 0.08$ & -- & $\textcolor{npgGreen}{4281} \pm 450$ & $\textcolor{npgNavy}{3994} \pm 1400$ \\
\texttt{msnbc} & $\textcolor{npgRed}{1.45} \pm 0.197$ & $\textcolor{npgPurple}{0.736} \pm 0.0271$ & $\textcolor{npgBlue}{\bm{0.172}} \pm 0.0051$ & $\textcolor{npgBrown}{5.38} \pm 0.077$ & -- & $\textcolor{npgGreen}{1.29} \pm 0.105$ & $\textcolor{npgNavy}{1.49} \pm 0.314$ \\
\texttt{msweb} & $\textcolor{npgRed}{1.64} \pm 0.291$ & $\textcolor{npgPurple}{11.3} \pm 0.308$ & $\textcolor{npgBlue}{\bm{0.185}} \pm 0.00856$ & $\textcolor{npgBrown}{1.95} \pm 0.032$ & -- & $\textcolor{npgGreen}{2134} \pm 674$ & $\textcolor{npgNavy}{2060} \pm 697$ \\
\texttt{mushrooms} & $\textcolor{npgRed}{0.43} \pm 0.0586$ & $\textcolor{npgPurple}{1.93} \pm 0.0639$ & $\textcolor{npgBlue}{\bm{0.0755}} \pm 0.00311$ & $\textcolor{npgBrown}{0.703} \pm 0.0255$ & -- & $\textcolor{npgGreen}{0.949} \pm 0.127$ & $\textcolor{npgNavy}{1.29} \pm 0.0995$ \\
\texttt{nips} & $\textcolor{npgRed}{0.303} \pm 0.038$ & $\textcolor{npgPurple}{9.91} \pm 0.33$ & $\textcolor{npgBlue}{\bm{0.101}} \pm 0.00478$ & $\textcolor{npgBrown}{0.553} \pm 0.0125$ & -- & $\textcolor{npgGreen}{2066} \pm 173$ & $\textcolor{npgNavy}{2411} \pm 202$ \\
\texttt{nltcs} & $\textcolor{npgRed}{0.141} \pm 0.0156$ & $\textcolor{npgPurple}{0.103} \pm 0.00149$ & $\textcolor{npgBlue}{\bm{0.0238}} \pm 0.000774$ & $\textcolor{npgBrown}{0.205} \pm 0.0562$ & -- & $\textcolor{npgGreen}{0.226} \pm 0.025$ & $\textcolor{npgNavy}{0.224} \pm 0.0307$ \\
\texttt{ocr\_letters} & $\textcolor{npgRed}{47.8} \pm 9.04$ & $\textcolor{npgPurple}{130} \pm 3.66$ & $\textcolor{npgBlue}{\bm{4.43}} \pm 0.296$ & -- & -- & $\textcolor{npgGreen}{11085} \pm 10567$ & $\textcolor{npgNavy}{11106} \pm 10755$ \\
\texttt{plants} & $\textcolor{npgRed}{5.21} \pm 0.868$ & $\textcolor{npgPurple}{9.15} \pm 0.182$ & $\textcolor{npgBlue}{\bm{0.573}} \pm 0.0349$ & $\textcolor{npgBrown}{34.9} \pm 1.39$ & -- & $\textcolor{npgGreen}{9.27} \pm 5.24$ & $\textcolor{npgNavy}{10.9} \pm 5.59$ \\
\texttt{pumsb\_star} & $\textcolor{npgRed}{2.41} \pm 0.312$ & $\textcolor{npgPurple}{14.4} \pm 0.389$ & $\textcolor{npgBlue}{\bm{0.401}} \pm 0.0207$ & $\textcolor{npgBrown}{2.23} \pm 0.0596$ & -- & $\textcolor{npgGreen}{10.3} \pm 9.59$ & $\textcolor{npgNavy}{11.1} \pm 9.24$ \\
\texttt{tmovie} & $\textcolor{npgRed}{11.2} \pm 2.58$ & $\textcolor{npgPurple}{115} \pm 3.87$ & $\textcolor{npgBlue}{\bm{1.23}} \pm 0.0754$ & $\textcolor{npgBrown}{23.1} \pm 0.313$ & -- & $\textcolor{npgGreen}{9687} \pm 758$ & $\textcolor{npgNavy}{12392} \pm 308$ \\
\texttt{tretail} & $\textcolor{npgRed}{0.12} \pm 0.0149$ & $\textcolor{npgPurple}{0.918} \pm 0.0307$ & $\textcolor{npgBlue}{\bm{0.0291}} \pm 0.0013$ & $\textcolor{npgBrown}{0.207} \pm 0.13$ & -- & $\textcolor{npgGreen}{0.426} \pm 0.0436$ & $\textcolor{npgNavy}{0.534} \pm 0.0531$ \\
\bottomrule
\end{tabular}
\end{table*}

\definecolor{npgRed}{HTML}{E64B35}
\definecolor{npgBlue}{HTML}{4DBBD5}
\definecolor{npgGreen}{HTML}{00A087}
\definecolor{npgNavy}{HTML}{3C5488}
\definecolor{npgPurple}{HTML}{8491B4}
\definecolor{npgOrange}{HTML}{F39B7F}
\definecolor{npgBrown}{HTML}{7E6148}

\begin{table*}[t]
\centering
\scriptsize
\setlength{\tabcolsep}{2pt}
\renewcommand{\arraystretch}{0.96}
\caption{Runtime comparison (seconds) for 50\% query and 30\% evidence.  We report the mean ± SE over ten trials.}
\label{tab:runtimes4}
\begin{tabular}{@{}lccccccc@{}}
\toprule
Dataset & AMP & IND & MP & HBP & ITSELF & PAC-MAP & smooth-PAC-MAP \\
\midrule
\texttt{accidents} & $\textcolor{npgRed}{4.48} \pm 0.656$ & $\textcolor{npgPurple}{31.7} \pm 0.835$ & $\textcolor{npgBlue}{\bm{0.488}} \pm 0.0233$ & $\textcolor{npgBrown}{14.9} \pm 0.573$ & -- & $\textcolor{npgGreen}{2633} \pm 1827$ & $\textcolor{npgNavy}{2549} \pm 1777$ \\
\texttt{adult} & $\textcolor{npgRed}{0.455} \pm 0.0757$ & $\textcolor{npgPurple}{5.25} \pm 0.204$ & $\textcolor{npgBlue}{\bm{0.0719}} \pm 0.00417$ & $\textcolor{npgBrown}{1.17} \pm 0.0581$ & -- & $\textcolor{npgGreen}{3.93} \pm 4.93$ & $\textcolor{npgNavy}{4.1} \pm 5.23$ \\
\texttt{baudio} & $\textcolor{npgRed}{15.4} \pm 2.96$ & $\textcolor{npgPurple}{49.3} \pm 0.854$ & $\textcolor{npgBlue}{\bm{0.845}} \pm 0.0398$ & $\textcolor{npgBrown}{12.1} \pm 0.146$ & -- & $\textcolor{npgGreen}{9538} \pm 1940$ & $\textcolor{npgNavy}{7750} \pm 1536$ \\
\texttt{bnetflix} & $\textcolor{npgRed}{4.59} \pm 0.789$ & $\textcolor{npgPurple}{34.2} \pm 1.29$ & $\textcolor{npgBlue}{\bm{0.581}} \pm 0.0351$ & $\textcolor{npgBrown}{5.6} \pm 0.11$ & -- & $\textcolor{npgGreen}{6147} \pm 379$ & $\textcolor{npgNavy}{6597} \pm 200$ \\
\texttt{book} & $\textcolor{npgRed}{15.8} \pm 3.82$ & $\textcolor{npgPurple}{284} \pm 13.7$ & $\textcolor{npgBlue}{\bm{1.15}} \pm 0.076$ & $\textcolor{npgBrown}{5.21} \pm 0.143$ & -- & $\textcolor{npgGreen}{10534} \pm 485$ & $\textcolor{npgNavy}{7199} \pm 171$ \\
\texttt{connect4} & $\textcolor{npgRed}{6.57} \pm 0.82$ & $\textcolor{npgPurple}{64.3} \pm 1.05$ & $\textcolor{npgBlue}{\bm{0.865}} \pm 0.03$ & $\textcolor{npgBrown}{12.3} \pm 0.198$ & -- & $\textcolor{npgGreen}{32.2} \pm 19$ & $\textcolor{npgNavy}{34.6} \pm 19.8$ \\
\texttt{dna} & $\textcolor{npgRed}{2.55} \pm 0.464$ & $\textcolor{npgPurple}{20.8} \pm 0.458$ & $\textcolor{npgBlue}{\bm{0.203}} \pm 0.00792$ & $\textcolor{npgBrown}{0.662} \pm 0.0873$ & -- & $\textcolor{npgGreen}{3273} \pm 586$ & $\textcolor{npgNavy}{2439} \pm 36.7$ \\
\texttt{jester} & $\textcolor{npgRed}{9.87} \pm 2.14$ & $\textcolor{npgPurple}{26} \pm 0.649$ & $\textcolor{npgBlue}{\bm{0.456}} \pm 0.0399$ & $\textcolor{npgBrown}{3.53} \pm 0.0543$ & -- & $\textcolor{npgGreen}{4521} \pm 147$ & $\textcolor{npgNavy}{2412} \pm 232$ \\
\texttt{kdd} & $\textcolor{npgRed}{1.58} \pm 0.331$ & $\textcolor{npgPurple}{5.73} \pm 0.196$ & $\textcolor{npgBlue}{\bm{0.15}} \pm 0.00991$ & $\textcolor{npgBrown}{0.756} \pm 0.0381$ & -- & $\textcolor{npgGreen}{23.1} \pm 40.3$ & $\textcolor{npgNavy}{26.1} \pm 47.9$ \\
\texttt{kosarek} & $\textcolor{npgRed}{3.47} \pm 0.714$ & $\textcolor{npgPurple}{37.9} \pm 1.03$ & $\textcolor{npgBlue}{\bm{0.359}} \pm 0.0183$ & $\textcolor{npgBrown}{6.14} \pm 0.045$ & -- & $\textcolor{npgGreen}{3943} \pm 126$ & $\textcolor{npgNavy}{2795} \pm 69.9$ \\
\texttt{msnbc} & $\textcolor{npgRed}{1.48} \pm 0.211$ & $\textcolor{npgPurple}{1.83} \pm 0.074$ & $\textcolor{npgBlue}{\bm{0.174}} \pm 0.00974$ & $\textcolor{npgBrown}{5.43} \pm 0.069$ & -- & $\textcolor{npgGreen}{1.23} \pm 0.112$ & $\textcolor{npgNavy}{1.55} \pm 0.187$ \\
\texttt{msweb} & $\textcolor{npgRed}{1.67} \pm 0.298$ & $\textcolor{npgPurple}{29} \pm 0.544$ & $\textcolor{npgBlue}{\bm{0.182}} \pm 0.00853$ & $\textcolor{npgBrown}{2.11} \pm 0.0285$ & -- & $\textcolor{npgGreen}{2237} \pm 79.6$ & $\textcolor{npgNavy}{2277} \pm 176$ \\
\texttt{mushrooms} & $\textcolor{npgRed}{0.435} \pm 0.0597$ & $\textcolor{npgPurple}{4.92} \pm 0.0896$ & $\textcolor{npgBlue}{\bm{0.0752}} \pm 0.00282$ & $\textcolor{npgBrown}{0.742} \pm 0.0205$ & -- & $\textcolor{npgGreen}{2.15} \pm 1.47$ & $\textcolor{npgNavy}{2.61} \pm 1.48$ \\
\texttt{nips} & $\textcolor{npgRed}{0.307} \pm 0.0268$ & $\textcolor{npgPurple}{25} \pm 0.807$ & $\textcolor{npgBlue}{\bm{0.0974}} \pm 0.00455$ & $\textcolor{npgBrown}{0.572} \pm 0.0226$ & -- & $\textcolor{npgGreen}{2642} \pm 293$ & $\textcolor{npgNavy}{3019} \pm 213$ \\
\texttt{nltcs} & $\textcolor{npgRed}{0.141} \pm 0.0156$ & $\textcolor{npgPurple}{0.253} \pm 0.00391$ & $\textcolor{npgBlue}{\bm{0.0238}} \pm 0.000656$ & $\textcolor{npgBrown}{0.222} \pm 0.0336$ & -- & $\textcolor{npgGreen}{0.22} \pm 0.0202$ & $\textcolor{npgNavy}{0.274} \pm 0.0525$ \\
\texttt{ocr\_letters} & $\textcolor{npgRed}{44.6} \pm 8.61$ & $\textcolor{npgPurple}{292} \pm 9.07$ & $\textcolor{npgBlue}{\bm{3.91}} \pm 0.303$ & -- & -- & $\textcolor{npgGreen}{5967} \pm 420$ & $\textcolor{npgNavy}{8616} \pm 2263$ \\
\texttt{plants} & $\textcolor{npgRed}{5.39} \pm 0.925$ & $\textcolor{npgPurple}{24.4} \pm 0.588$ & $\textcolor{npgBlue}{\bm{0.602}} \pm 0.0285$ & $\textcolor{npgBrown}{34.9} \pm 0.77$ & -- & $\textcolor{npgGreen}{59.2} \pm 59.6$ & $\textcolor{npgNavy}{57.2} \pm 52.1$ \\
\texttt{pumsb\_star} & $\textcolor{npgRed}{2.49} \pm 0.321$ & $\textcolor{npgPurple}{36.9} \pm 0.664$ & $\textcolor{npgBlue}{\bm{0.397}} \pm 0.0153$ & $\textcolor{npgBrown}{2.34} \pm 0.0555$ & -- & $\textcolor{npgGreen}{965} \pm 1540$ & $\textcolor{npgNavy}{998} \pm 1584$ \\
\texttt{tmovie} & $\textcolor{npgRed}{11.6} \pm 2.68$ & $\textcolor{npgPurple}{295} \pm 11.2$ & $\textcolor{npgBlue}{\bm{1.22}} \pm 0.111$ & $\textcolor{npgBrown}{24.6} \pm 0.244$ & -- & $\textcolor{npgGreen}{11863} \pm 323$ & $\textcolor{npgNavy}{11153} \pm 467$ \\
\texttt{tretail} & $\textcolor{npgRed}{0.12} \pm 0.0147$ & $\textcolor{npgPurple}{2.25} \pm 0.039$ & $\textcolor{npgBlue}{\bm{0.0287}} \pm 0.00109$ & $\textcolor{npgBrown}{0.206} \pm 0.113$ & -- & $\textcolor{npgGreen}{4.61} \pm 8.79$ & $\textcolor{npgNavy}{5.21} \pm 9.43$ \\
\bottomrule
\end{tabular}
\end{table*}

\definecolor{npgRed}{HTML}{E64B35}
\definecolor{npgBlue}{HTML}{4DBBD5}
\definecolor{npgGreen}{HTML}{00A087}
\definecolor{npgNavy}{HTML}{3C5488}
\definecolor{npgPurple}{HTML}{8491B4}
\definecolor{npgOrange}{HTML}{F39B7F}
\definecolor{npgBrown}{HTML}{7E6148}

\begin{table*}[t]
\centering
\scriptsize
\setlength{\tabcolsep}{1mm}
\renewcommand{\arraystretch}{0.96}
\caption{Runtime comparison (seconds) for 10\% query and 90\% evidence. We report the mean $\pm$ standard deviation over the available trials.}
\label{tab:pgm_runtimes}
\begin{tabular}{@{}lcccccc@{}}
\toprule
Dataset & DAOOPT-AOBB & DAOOPT-AOBF & IJGP & WMB & PAC-MAP & smooth-PAC-MAP \\
\midrule
\texttt{accidents} & $\textcolor{npgRed}{0.0383} \pm 0.00466$ & $\textcolor{npgBlue}{\bm{0.0367}} \pm 0.000524$ & $\textcolor{npgGreen}{0.0492} \pm 0.00172$ & $\textcolor{npgBrown}{0.0499} \pm 0.0051$ & $\textcolor{npgNavy}{0.321} \pm 0.145$ & $\textcolor{npgPurple}{0.3} \pm 0.0155$ \\
\texttt{adult} & $\textcolor{npgRed}{0.0431} \pm 0.0152$ & $\textcolor{npgBlue}{\bm{0.0383}} \pm 0.000592$ & $\textcolor{npgGreen}{0.0514} \pm 0.00193$ & $\textcolor{npgBrown}{0.0521} \pm 0.00283$ & $\textcolor{npgNavy}{0.353} \pm 0.159$ & $\textcolor{npgPurple}{0.324} \pm 0.00265$ \\
\texttt{baudio} & $\textcolor{npgRed}{0.0405} \pm 0.0118$ & $\textcolor{npgBlue}{\bm{0.0368}} \pm 0.00053$ & $\textcolor{npgGreen}{0.0468} \pm 0.000389$ & $\textcolor{npgBrown}{0.0473} \pm 0.000929$ & $\textcolor{npgNavy}{0.398} \pm 0.128$ & $\textcolor{npgPurple}{0.411} \pm 0.162$ \\
\texttt{bnetflix} & $\textcolor{npgRed}{\bm{0.0347}} \pm 0.000616$ & $\textcolor{npgBlue}{0.0348} \pm 0.000347$ & $\textcolor{npgGreen}{0.0455} \pm 0.000864$ & $\textcolor{npgBrown}{0.0458} \pm 0.000722$ & $\textcolor{npgNavy}{0.82} \pm 0.633$ & $\textcolor{npgPurple}{0.834} \pm 0.618$ \\
\texttt{connect4} & $\textcolor{npgRed}{0.0353} \pm 0.00859$ & $\textcolor{npgBlue}{\bm{0.0345}} \pm 0.00772$ & $\textcolor{npgGreen}{0.0447} \pm 0.00938$ & $\textcolor{npgBrown}{0.0488} \pm 0.0117$ & $\textcolor{npgNavy}{51.3} \pm 87.2$ & $\textcolor{npgPurple}{0.448} \pm 0.181$ \\
\texttt{dna} & $\textcolor{npgRed}{\bm{0.0461}} \pm 0.000965$ & $\textcolor{npgBlue}{0.0463} \pm 0.00102$ & $\textcolor{npgGreen}{0.0596} \pm 0.00188$ & $\textcolor{npgBrown}{0.06} \pm 0.00352$ & $\textcolor{npgNavy}{1} \pm 1.13$ & $\textcolor{npgPurple}{1.04} \pm 1.18$ \\
\texttt{jester} & $\textcolor{npgRed}{0.0231} \pm 0.000643$ & $\textcolor{npgBlue}{\bm{0.0231}} \pm 0.000557$ & $\textcolor{npgGreen}{0.0305} \pm 0.00079$ & $\textcolor{npgBrown}{0.0304} \pm 0.00103$ & $\textcolor{npgNavy}{0.402} \pm 0.24$ & $\textcolor{npgPurple}{0.414} \pm 0.246$ \\
\texttt{kdd} & $\textcolor{npgRed}{\bm{0.0196}} \pm 0.000781$ & $\textcolor{npgBlue}{0.0197} \pm 0.000747$ & $\textcolor{npgGreen}{0.0244} \pm 0.000907$ & $\textcolor{npgBrown}{0.0245} \pm 0.00127$ & $\textcolor{npgNavy}{0.26} \pm 0.456$ & $\textcolor{npgPurple}{0.109} \pm 0.00781$ \\
\texttt{kosarek} & $\textcolor{npgRed}{\bm{0.0429}} \pm 0.00155$ & $\textcolor{npgBlue}{0.0437} \pm 0.00269$ & $\textcolor{npgGreen}{0.0587} \pm 0.00202$ & $\textcolor{npgBrown}{0.058} \pm 0.00294$ & $\textcolor{npgNavy}{2.59} \pm 4.15$ & $\textcolor{npgPurple}{0.547} \pm 0.251$ \\
\texttt{msnbc} & $\textcolor{npgRed}{0.0192} \pm 0.00633$ & $\textcolor{npgBlue}{\bm{0.0181}} \pm 0.00158$ & $\textcolor{npgGreen}{0.0192} \pm 0.00228$ & $\textcolor{npgBrown}{0.0217} \pm 0.00869$ & $\textcolor{npgNavy}{0.0341} \pm 0.00189$ & $\textcolor{npgPurple}{0.03} \pm 0.00125$ \\
\texttt{mushrooms} & $\textcolor{npgRed}{0.025} \pm 0.00247$ & $\textcolor{npgBlue}{\bm{0.0242}} \pm 0.00127$ & $\textcolor{npgGreen}{0.0324} \pm 0.000753$ & $\textcolor{npgBrown}{0.0329} \pm 0.00208$ & $\textcolor{npgNavy}{0.214} \pm 0.0742$ & $\textcolor{npgPurple}{0.207} \pm 0.0592$ \\
\texttt{nltcs} & $\textcolor{npgRed}{\bm{0.0155}} \pm 0.000496$ & $\textcolor{npgBlue}{0.0162} \pm 0.000678$ & $\textcolor{npgGreen}{0.0171} \pm 0.000573$ & $\textcolor{npgBrown}{0.017} \pm 0.000521$ & $\textcolor{npgNavy}{0.0309} \pm 0.000481$ & $\textcolor{npgPurple}{0.0271} \pm 0.00119$ \\
\texttt{ocr\_letters} & $\textcolor{npgRed}{\bm{0.028}} \pm 0.000625$ & $\textcolor{npgBlue}{0.0282} \pm 0.000709$ & $\textcolor{npgGreen}{0.0379} \pm 0.00164$ & $\textcolor{npgBrown}{0.0403} \pm 0.00827$ & $\textcolor{npgNavy}{0.43} \pm 0.15$ & $\textcolor{npgPurple}{0.393} \pm 0.142$ \\
\texttt{plants} & $\textcolor{npgRed}{\bm{0.0202}} \pm 0.000808$ & $\textcolor{npgBlue}{0.0205} \pm 0.000552$ & $\textcolor{npgGreen}{0.025} \pm 0.000451$ & $\textcolor{npgBrown}{0.0251} \pm 0.00104$ & $\textcolor{npgNavy}{0.113} \pm 0.00305$ & $\textcolor{npgPurple}{0.114} \pm 0.00347$ \\
\texttt{pumsb\_star} & $\textcolor{npgRed}{\bm{0.0318}} \pm 0.000376$ & $\textcolor{npgBlue}{0.0318} \pm 0.000555$ & $\textcolor{npgGreen}{0.0448} \pm 0.000908$ & $\textcolor{npgBrown}{0.044} \pm 0.00172$ & $\textcolor{npgNavy}{5.49} \pm 9.26$ & $\textcolor{npgPurple}{0.422} \pm 0.359$ \\
\texttt{tretail} & $\textcolor{npgRed}{0.0274} \pm 0.000975$ & $\textcolor{npgBlue}{\bm{0.0271}} \pm 0.000667$ & $\textcolor{npgGreen}{0.0379} \pm 0.00156$ & $\textcolor{npgBrown}{0.0385} \pm 0.0037$ & $\textcolor{npgNavy}{0.219} \pm 0.00819$ & $\textcolor{npgPurple}{0.23} \pm 0.00776$ \\
\bottomrule
\end{tabular}
\end{table*}

\definecolor{npgRed}{HTML}{E64B35}
\definecolor{npgBlue}{HTML}{4DBBD5}
\definecolor{npgGreen}{HTML}{00A087}
\definecolor{npgNavy}{HTML}{3C5488}
\definecolor{npgPurple}{HTML}{8491B4}
\definecolor{npgOrange}{HTML}{F39B7F}
\definecolor{npgBrown}{HTML}{7E6148}

\begin{table*}[t]
\centering
\scriptsize
\setlength{\tabcolsep}{1mm}
\renewcommand{\arraystretch}{0.96}
\caption{Runtime comparison (seconds) for 20\% query and 80\% evidence. We report the mean $\pm$ standard deviation over the available trials.}
\label{tab:pgm_runtimes}
\begin{tabular}{@{}lcccccc@{}}
\toprule
Dataset & DAOOPT-AOBB & DAOOPT-AOBF & IJGP & WMB & PAC-MAP & smooth-PAC-MAP \\
\midrule
\texttt{accidents} & $\textcolor{npgRed}{0.134} \pm 0.0117$ & $\textcolor{npgBlue}{\bm{0.132}} \pm 0.00829$ & $\textcolor{npgGreen}{0.139} \pm 0.00774$ & $\textcolor{npgBrown}{0.152} \pm 0.0302$ & $\textcolor{npgNavy}{1.76} \pm 2.62$ & $\textcolor{npgPurple}{0.696} \pm 0.261$ \\
\texttt{adult} & $\textcolor{npgRed}{\bm{0.384}} \pm 0.03$ & $\textcolor{npgBlue}{0.39} \pm 0.0407$ & $\textcolor{npgGreen}{0.409} \pm 0.0414$ & $\textcolor{npgBrown}{0.427} \pm 0.07$ & $\textcolor{npgNavy}{3.88} \pm 8.53$ & $\textcolor{npgPurple}{0.923} \pm 0.341$ \\
\texttt{baudio} & $\textcolor{npgRed}{0.0575} \pm 0.00426$ & $\textcolor{npgBlue}{\bm{0.0572}} \pm 0.00265$ & $\textcolor{npgGreen}{0.0642} \pm 0.0027$ & $\textcolor{npgBrown}{0.0661} \pm 0.0071$ & $\textcolor{npgNavy}{6.5} \pm 5.09$ & $\textcolor{npgPurple}{4.54} \pm 4.37$ \\
\texttt{bnetflix} & $\textcolor{npgRed}{0.0615} \pm 0.00564$ & $\textcolor{npgBlue}{\bm{0.0588}} \pm 0.00548$ & $\textcolor{npgGreen}{0.0681} \pm 0.00688$ & $\textcolor{npgBrown}{0.0717} \pm 0.00646$ & $\textcolor{npgNavy}{22.2} \pm 20.3$ & $\textcolor{npgPurple}{21.9} \pm 21.3$ \\
\texttt{connect4} & $\textcolor{npgRed}{0.763} \pm 0.0684$ & $\textcolor{npgBlue}{\bm{0.751}} \pm 0.0654$ & $\textcolor{npgGreen}{0.761} \pm 0.0826$ & $\textcolor{npgBrown}{0.781} \pm 0.0658$ & $\textcolor{npgNavy}{516} \pm 422$ & $\textcolor{npgPurple}{118} \pm 264$ \\
\texttt{jester} & $\textcolor{npgRed}{\bm{0.061}} \pm 0.005$ & $\textcolor{npgBlue}{0.0614} \pm 0.00364$ & $\textcolor{npgGreen}{0.067} \pm 0.00396$ & $\textcolor{npgBrown}{0.077} \pm 0.0191$ & $\textcolor{npgNavy}{20.4} \pm 14.9$ & $\textcolor{npgPurple}{20.5} \pm 14.9$ \\
\texttt{kdd} & $\textcolor{npgRed}{\bm{0.0289}} \pm 0.000852$ & $\textcolor{npgBlue}{0.0296} \pm 0.00105$ & $\textcolor{npgGreen}{0.0349} \pm 0.000814$ & $\textcolor{npgBrown}{0.0346} \pm 0.00224$ & $\textcolor{npgNavy}{0.157} \pm 0.0499$ & $\textcolor{npgPurple}{0.135} \pm 0.0184$ \\
\texttt{msnbc} & $\textcolor{npgRed}{0.0254} \pm 0.00669$ & $\textcolor{npgBlue}{\bm{0.023}} \pm 0.000877$ & $\textcolor{npgGreen}{0.0252} \pm 0.00111$ & $\textcolor{npgBrown}{0.0285} \pm 0.0101$ & $\textcolor{npgNavy}{0.0573} \pm 0.0516$ & $\textcolor{npgPurple}{0.0352} \pm 0.00332$ \\
\texttt{mushrooms} & $\textcolor{npgRed}{0.132} \pm 0.0116$ & $\textcolor{npgBlue}{\bm{0.132}} \pm 0.00968$ & $\textcolor{npgGreen}{0.145} \pm 0.0127$ & $\textcolor{npgBrown}{0.145} \pm 0.00924$ & $\textcolor{npgNavy}{15.6} \pm 24.4$ & $\textcolor{npgPurple}{0.786} \pm 0.321$ \\
\texttt{nltcs} & $\textcolor{npgRed}{0.0238} \pm 0.000836$ & $\textcolor{npgBlue}{\bm{0.0234}} \pm 0.000453$ & $\textcolor{npgGreen}{0.0263} \pm 0.000875$ & $\textcolor{npgBrown}{0.0299} \pm 0.0118$ & $\textcolor{npgNavy}{0.0386} \pm 0.00188$ & $\textcolor{npgPurple}{0.0323} \pm 0.00105$ \\
\texttt{ocr\_letters} & $\textcolor{npgRed}{0.761} \pm 0.0644$ & $\textcolor{npgBlue}{\bm{0.745}} \pm 0.0687$ & $\textcolor{npgGreen}{0.78} \pm 0.0803$ & $\textcolor{npgBrown}{0.817} \pm 0.159$ & $\textcolor{npgNavy}{96.8} \pm 95.5$ & $\textcolor{npgPurple}{84.8} \pm 102$ \\
\texttt{plants} & $\textcolor{npgRed}{0.0221} \pm 0.00128$ & $\textcolor{npgBlue}{\bm{0.022}} \pm 0.00104$ & $\textcolor{npgGreen}{0.0276} \pm 0.00138$ & $\textcolor{npgBrown}{0.0293} \pm 0.00699$ & $\textcolor{npgNavy}{0.236} \pm 0.272$ & $\textcolor{npgPurple}{0.159} \pm 0.0495$ \\
\texttt{tretail} & $\textcolor{npgRed}{\bm{2.94}} \pm 0.321$ & $\textcolor{npgBlue}{3.39} \pm 1.34$ & $\textcolor{npgGreen}{3.14} \pm 0.888$ & $\textcolor{npgBrown}{3.2} \pm 0.87$ & $\textcolor{npgNavy}{7.76} \pm 2.22$ & $\textcolor{npgPurple}{9.31} \pm 1.31$ \\
\bottomrule
\end{tabular}
\end{table*}

\section{C: Pseudocode}

In this section, we provide pseudocode for $\texttt{smooth-PAC-MAP}$, which adds an exploitation step to the pure exploration strategy of $\texttt{PAC-MAP}$. This can lead to stronger results and faster convergence under mild smoothness assumptions. In the pseudocode, the exploration/exploitation tradeoff is probabilistically controlled by a hyperparameter $\eta$; in our implementation, we deterministically run an exploit loop every $250$ samples, effectively imposing $\eta = 0.004$.

\begin{algorithm}[tb] 
\caption{Smooth PAC-MAP: $\pi_{\mathrm S}(P, p, \varepsilon, \delta, L, r, \eta)$}
\label{alg:pacmap_exploit}
\textbf{Input}: Sampler $P$, oracle $p$, tolerance $\varepsilon$, level $\delta$, Lipschitz constant $L$, radius $r$, exploitation probability $\eta$\\
\textbf{Output}: PAC-MAP solution $\hat{\bm q}, \hat p$\\
\begin{algorithmic}[1]
\vspace{-1em}
\STATE $S \gets \emptyset$ \quad \texttt{// Initialize candidate set}
\STATE $m \gets 0$ \quad \texttt{// Initialize sample count}
\STATE $M \gets \infty$ \quad \texttt{// Initialize stop time}
\STATE $\hat p_0 \gets 2^{-n}$ \quad \texttt{// Initialize MAP probability}
\STATE $v_r \gets \sum_{j=0}^r \binom{n}{j}$ \quad \texttt{// Hamming ball volume}
\STATE $k \gets \min\{r, \lfloor \frac{\log_2 \left(1 / (1 - \varepsilon)\right)}{L} \rfloor\}$ \quad \texttt{// Relevant shell radius}
\STATE $w_k \gets \sum_{j=1}^k \binom{n}{j} 2^{-Lj}$ \quad \texttt{// Weighted ball volume}

\WHILE{$m < M$}
    \STATE $\text{Explore} \sim \text{Bern}(1 - \eta)$ \quad \texttt{Explore or exploit?} 
    \IF{$\text{Explore} = \texttt{TRUE}$}
        \STATE $m \gets m + 1$ \quad \texttt{// Update count}
        \STATE $\bm q \sim P(\bm Q \mid \bm e)$ \quad \texttt{// Draw sample} 
        \STATE $S \gets S \cup \bm q$ \quad \texttt{// Admit new candidate}
    \ELSIF{$\hat p_m > \hat p_{m-1}$} 
        \STATE $S \gets S \cup \mathcal B_r(\hat{\bm q})$ \quad \texttt{// Hamming ball sweep}
    \ENDIF
    \STATE $\hat{\bm{q}}_m \leftarrow \argmax_{\bm q \in S} ~p(\bm q \mid \bm{e})$ \quad \texttt{// Leading candidate}
    \STATE $\hat p_m \gets p(\hat{\bm q} \mid \bm e)$ \quad \texttt{// Associated probability}
    
    \STATE $\check p_m \gets 1 - \sum_{\bm q \in S} ~p(\bm q \mid \bm e)$ \texttt{// Residual mass}
    \IF{$\hat p_m \geq \check p_m(1 - \varepsilon)$}
        \STATE \textbf{break} \quad \texttt{// Algorithm has converged}
    \ELSE 
        \STATE $M \gets \frac{(1 - \varepsilon) ~\log  \delta^{-1}}{\hat p_m w_k} $ \quad \texttt{// Update stop time}
    \ENDIF
\ENDWHILE

\RETURN $\hat{\bm q}_m, ~\hat p_m$
\end{algorithmic}
\end{algorithm}

\section{D: Formal PC definitions}
Formally, let $\mathcal C = (\mathcal G, \bm \theta)$ denote a PC over variables \(\bm{Z}\) with graphical structure $\mathcal G$ and mixture weights $\bm \theta$. Let \(\node\) denote a node in \(\pc\), and \(\ch{\node}\) the set of \(\node\)'s children. The probability distribution at \(\node\) is then recursively defined as follows:
\begin{equation*}
    \node(\bm{z}) = 
    \begin{cases}
        p_{\node}(\bm{z}) \: &\text{if \(\node\) is a leaf node}\\
        \prod_{c \in \ch{\node}} c(\bm{z}) &\text{if \(\node\) is a product node} \\
        \sum_{c \in \ch{\node}} \theta_{\node, c} \cdot c(\bm{z}) &\text{if \(\node\) is a sum node}.
    \end{cases}
\end{equation*}
A major advantage of PCs is that structural properties of circuits can guarantee polytime inference for various probabilistic inference tasks. We consider three properties in particular: smoothness, decomposability, and determinisim.

A sum node is \textit{smooth} if all of its children depend on all of the same variables. 
A product node is \textit{decomposable} if all of its children depend on disjoint sets of variables.
We say that a PC is smooth if all of its sum nodes are smooth, and decomposable if all of its product nodes are decomposable. A smooth, decomposable PC can compute marginal queries of the form $\sum_{\bm v} p(\bm q, \bm v)$ in linear time \citep{darwiche_ACs, peharz2015}.\footnote{The complexity of inference tasks in PCs is measured w.r.t. circuit size $|\mathcal C|$ (i.e., the number of edges in $\mathcal G$).} 
This makes them well suited for PAC-MAP---plugging a smooth, deterministic circuit $\mathcal C$ in for our sampler and oracle in the algorithms above increases runtime by a constant factor.

A sum node is \textit{deterministic} if, for all fully instantiated inputs, the output of at most one of its children is nonzero.
Deterministic PCs impose a partition on the feature space, such that any full assignment of values to variables eliminates all but a single path through the sum nodes. It has been shown that decomposable, deterministic PCs can solve MPE in linear time \citep{chan2006}.\footnote{In fact, decomposability can be replaced with a weaker property known as \textit{consistency} \citep[§6.2]{choi2020}. Since our target in this section is determinism, not decomposability, we will not explore this direction further.} 
For example, a decision tree that models joint distributions via recursive partitions and piecewise constant likelihoods can be compiled into a deterministic PC \citep{correia_joints_2020}. This might seem to resolve MPE outright, given that DTs are universally consistent under mild assumptions \citep{lugosi1996}. However, this guarantee only holds for asymptotically deep trees. Linear dependence on circuit size is not so benign when the tree from which a PC is compiled may be indefinitely large. Furthermore, DTs are unstable predictors—hence their typical deployment within ensembles like random forests \citep{Breiman2001} or gradient boosting \citep{friedman_boost}, which do not preserve determinism. 

Several PC architectures are based on tree ensembles \citep{dimauro_cutset_forest2015, correia_joints_2020, liu2021tractable, watson_adversarial_2023}, but they do not generally satisfy determinism. 
Though rigorous results are lacking, empirical evidence suggests that determinism has a negative impact on expressive efficiency, which may explain why most leading PCs do not impose it by default.
Structural constraints for tractable MMAP are even more restrictive \citep{oztok2015}.
These challenges highlight the need for approximate MAP solvers in non-deterministic PCs \citep{conaty2017, mei_maxSPN, maua2020, choi_solving_2022, arya_nips2024}.
We test against several such methods in section 5 of the main text.

%% file: bibbythebib.bib
@InProceedings{maua2020,
  title = 	 {Two Reformulation Approaches to Maximum-A-Posteriori Inference in Sum-Product Networks},
  author =       {Mau\'a, Denis Deratani and Reis, Heitor Ribeiro and Katague, Gustavo Perez and Antonucci, Alessandro},
  booktitle = 	 {Proceedings of the 10th International Conference on Probabilistic Graphical Models},
  pages = 	 {293--304},
  year = 	 {2020}
}

@inproceedings{arya_nips2024,
	author = {Arya, Shivvrat and Rahman, Tahrima and Gogate, Vibhav},
	booktitle = {Advances in Neural Information Processing Systems},
	pages = {33538--33601},
	title = {A Neural Network Approach for Efficiently Answering Most Probable Explanation Queries in Probabilistic Models},
	volume = {37},
	year = {2024}
}

@InProceedings{peharz2015,
  title = 	 {{On Theoretical Properties of Sum-Product Networks}},
  author = 	 {Peharz, Robert and Tschiatschek, Sebastian and Pernkopf, Franz and Domingos, Pedro},
  booktitle = 	 {Proceedings of the 18th International Conference on Artificial Intelligence and Statistics},
  pages = 	 {744--752},
  year = 	 {2015},
  volume = 	 {38}
}

@inproceedings{watson_adversarial_2023,
	title = {Adversarial {Random} {Forests} for {Density} {Estimation} and {Generative} {Modeling}},
	booktitle = {Proceedings of {The} 26th {International} {Conference} on {Artificial} {Intelligence} and {Statistics}},
	publisher = {PMLR},
	author = {Watson, David S. and Blesch, Kristin and Kapar, Jan and Wright, Marvin N.},
	year = {2023},
	pages = {5357--5375}
}

@article{SHIMONY1994399,
	author = {Solomon Eyal Shimony},
	journal = {Artificial Intelligence},
	number = {2},
	pages = {399-410},
	title = {Finding {MAP}s for belief networks is {NP}-hard},
	volume = {68},
	year = {1994}
}

@inproceedings{correia_joints_2020,
	title = {Joints in {Random} {Forests}},
	volume = {33},
	urldate = {2024-10-22},
	booktitle = {Advances in {Neural} {Information} {Processing} {Systems}},
	author = {Correia, Alvaro and Peharz, Robert and de Campos, Cassio P},
	year = {2020},
	pages = {11404--11415},
}

@inproceedings{conaty2017,
	title = {Approximation Complexity of Maximum A Posteriori Inference in Sum-Product Networks},
	booktitle = {Proceedings of {The} 32nd {Uncertainty} in {Artificial} {Intelligence} {Conference}},
	author = {Conaty, D. and Mau\'a, D. and de Campos, C.},
	year = {2017}
}

@unpublished{choi2020,
author = {YooJung Choi and Antonio Vergari and Guy {Van den Broeck}},
title = {Probabilistic Circuits: A Unifying Framework for Tractable Probabilistic Models},
year = {2020}, 
note = {Technical Report, University of California, Los Angeles}
}

@inproceedings{choi_solving_2022,
	title = {Solving {Marginal} {MAP} {Exactly} by {Probabilistic} {Circuit} {Transformations}},
	booktitle = {Proceedings of {The} 25th {International} {Conference} on {Artificial} {Intelligence} and {Statistics}},
	author = {Choi, Yoojung and Friedman, Tal and Van den Broeck, Guy},
	year = {2022},
	pages = {10196--10208}
}

@article{darwiche_ACs,
author = {Darwiche, Adnan},
title = {A differential approach to inference in {B}ayesian networks},
year = {2003},
volume = {50},
number = {3},
journal = {J. ACM},
pages = {280–305}
}

@inproceedings{chan2006,
author = {Chan, Hei and Darwiche, Adnan},
title = {On the robustness of most probable explanations},
year = {2006},
booktitle = {Proceedings of the Twenty-Second Conference on Uncertainty in Artificial Intelligence},
pages = {63–71}
}

@InProceedings{decampos_no_news,
  title = 	 {Almost No News on the Complexity of {MAP} in {B}ayesian Networks},
  author =       {de Campos, Cassio P.},
  booktitle = 	 {Proceedings of the 10th International Conference on Probabilistic Graphical Models},
  pages = 	 {149--160},
  year = 	 {2020}
}

@book{darwiche2009modeling,
  title={Modeling and reasoning with Bayesian networks},
  author={Darwiche, Adnan},
  year={2009},
  publisher={Cambridge university press}
}

@article{park_darwiche_2004,
  title={Complexity Results and Approximation Strategies for {MAP} Explanations},
  author={Park, J.D. and Darwiche, A.},
  journal={Journal of Artificial Intelligence Research},
  volume={21},
  pages={101--133},
  year={2004}
}

@inproceedings{poon2011sum,
  title={Sum-product networks: A new deep architecture},
  author={Poon, Hoifung and Domingos, Pedro},
  booktitle={2011 IEEE International Conference on Computer Vision Workshops (ICCV Workshops)},
  pages={689--690},
  year={2011},
  organization={IEEE}
}

@book{Lattimore2020bandit,
  author    = {Lattimore, Tor and Szepesv{\'a}ri, Csaba},
  title     = {Bandit Algorithms},
  publisher = {Cambridge University Press},
  year      = {2020}
}

@inproceedings{vergari_tutorial,
  title={Probabilistic circuits: Representations, inference, learning and applications},
  author={Vergari, Antonio and Choi, YooJung and Peharz, Robert and Van den Broeck, Guy},
  booktitle={AAAI Tutorial},
  year={2020}
}

@article{liu2021tractable,
  title={Tractable regularization of probabilistic circuits},
  author={Liu, Anji and Van den Broeck, Guy},
  journal={Advances in Neural Information Processing Systems},
  volume={34},
  pages={3558--3570},
  year={2021}
}

@inproceedings{dimauro_cutset_forest2015,
	author = {Di Mauro, Nicola and Vergari, Antonio and Basile, Teresa M. A.},
	booktitle = {Foundations of Intelligent Systems},
	pages = {122--132},
	title = {Learning {B}ayesian Random Cutset Forests},
	year = {2015}
}

@inproceedings{mei_maxSPN,
author = {Mei, Jun and Jiang, Yong and Tu, Kewei},
title = {Maximum a posteriori inference in sum-product networks},
year = {2018},
booktitle = {Proceedings of the 32nd AAAI Conference on Artificial Intelligence}
}

@inproceedings{deCampos2011,
author = {de Campos, Cassio P.},
title = {New complexity results for {MAP} in {B}ayesian networks},
year = {2011},
booktitle = {Proceedings of the 22nd International Joint Conference on Artificial Intelligence},
pages = {2100–2106}
}

@inproceedings{audibert2010,
  TITLE = {Best Arm Identification in Multi-Armed Bandits},
  AUTHOR = {Audibert, Jean-Yves and Bubeck, S{\'e}bastien},
  BOOKTITLE = {23rd Annual Conference on Learning Theory},
  YEAR = {2010}
}

@InProceedings{garivier2016,
  title = 	 {Optimal Best Arm Identification with Fixed Confidence},
  author = 	 {Garivier, Aurélien and Kaufmann, Emilie},
  booktitle = 	 {29th Annual Conference on Learning Theory},
  pages = 	 {998--1027},
  year = 	 {2016}
}

@InProceedings{russo2016,
  title = 	 {Simple Bayesian Algorithms for Best Arm Identification},
  author = 	 {Russo, Daniel},
  booktitle = 	 {29th Annual Conference on Learning Theory},
  pages = 	 {1417--1418},
  year = 	 {2016}
}

@article{kaufman2016,
  author  = {Emilie Kaufmann and Olivier Capp{{\'e}} and Aur{{\'e}}lien Garivier},
  title   = {On the Complexity of Best-Arm Identification in Multi-Armed Bandit Models},
  journal = {J. Mach. Learn. Res.},
  year    = {2016},
  volume  = {17},
  number  = {1}
}

@inproceedings{wang2008,
  title     = {Algorithms for Infinitely Many-Armed Bandits},
  author    = {Wang, Yizao and Audibert, Jean-Yves and Munos, R{\'e}mi},
  booktitle = {Advances in Neural Information Processing Systems},
  volume    = {21},
  year      = {2008}
}

@article{bubeck2011,
  title   = {{X}-Armed Bandits},
  author  = {Bubeck, S{\'e}bastien and Munos, R{\'e}mi and Stoltz, Gilles and Szepesv{\'a}ri, Csaba},
  journal = {J. Mach. Learn. Res.},
  volume  = {12},
  number  = {46},
  pages   = {1655--1695},
  year    = {2011}
}

@inproceedings{carpentier2015,
  title     = {Simple Regret for Infinitely Many Armed Bandits},
  author    = {Carpentier, Alexandra and Valko, Michal},
  booktitle = {Proceedings of the 32nd International Conference on Machine Learning},
  volume    = {37},
  pages     = {1133--1141},
  year      = {2015}
}

@inproceedings{kleinberg2008,
author = {Kleinberg, Robert and Slivkins, Aleksandrs and Upfal, Eli},
title = {Multi-armed bandits in metric spaces},
year = {2008},
booktitle = {Proceedings of the Fortieth Annual ACM Symposium on Theory of Computing},
pages = {681–690}
}

@inproceedings{munos2011,
  title     = {Optimistic Optimization of a Deterministic Function without the Knowledge of its Smoothness},
  author    = {Munos, R{\'e}mi},
  booktitle = {Advances in Neural Information Processing Systems},
  volume    = {24},
  pages     = {783--791},
  year      = {2011}
}

@inproceedings{aziz2018,
  title     = {Pure Exploration in Infinitely-Armed Bandit Models with Fixed-Confidence},
  author    = {Aziz, Maryam and Anderton, Jesse and Kaufmann, Emilie and Aslam, Javed},
  booktitle = {Proceedings of the 29th International Conference on Algorithmic Learning Theory},
  pages     = {3--24},
  year      = {2018}
}

@inproceedings{gong2023,
	author = {Gong, Evelyn Xiao-Yue and Sellke, Mark},
	booktitle = {Advances in Neural Information Processing Systems},
	pages = {18551--18581},
	title = {Asymptotically Optimal Quantile Pure Exploration for Infinite-Armed Bandits},
	volume = {36},
	year = {2023}
}

@article{otten2017and,
  title={AND/OR branch-and-bound on a computational grid},
  author={Otten, Lars and Dechter, Rina},
  journal={Journal of Artificial Intelligence Research},
  volume={59},
  pages={351--435},
  year={2017}
}

@misc{daooptgithub,
  author       = {Otten, Lars and Lam, William},
  title        = {DAOOPT: Distributed AND/OR Optimization},
  year         = {2014},
  publisher    = {GitHub},
  journal      = {GitHub repository},
  howpublished = {\url{https://github.com/lotten/daoopt/tree/uai14}},
}

@misc{merlingithub,
  author       = {Marinescu, Radu},
  title        = {Merlin: An extensible C++ library for probabilistic inference in graphical models.},
  year         = {2016},
  publisher    = {GitHub},
  journal      = {GitHub repository},
  howpublished = {\url{https://github.com/radum2275/merlin}},
}

@misc{dechter2012iterativejoingraphpropagation,
      title={Iterative Join-Graph Propagation}, 
      author={Rina Dechter and Kalev Kask and Robert Mateescu},
      year={2012},
      eprint={1301.0564},
      archivePrefix={arXiv},
      primaryClass={cs.AI},
      url={https://arxiv.org/abs/1301.0564}, 
}

@inproceedings{liu2011bounding,
  title={Bounding the Partition Function using Holder's Inequality.},
  author={Liu, Qiang and Ihler, Alexander},
  booktitle={ICML},
  pages={849--856},
  year={2011}
}

@article{Breiman2001,
author = {Breiman, Leo},
journal = {Mach. Learn.},
number = {1},
pages = {1--33},
title = {Random Forests},
volume = {45},
year = {2001}
}

@article{lugosi1996,
	author = {G{\'a}bor Lugosi and Andrew Nobel},
	journal = {Ann. Statist.},
	number = {2},
	pages = {687 -- 706},
	title = {{Consistency of data-driven histogram methods for density estimation and classification}},
	volume = {24},
	year = {1996}
}

@article{friedman_boost,
	author = {Jerome H. Friedman},
	journal = {Ann. Stat.},
	number = {5},
	pages = {1189 -- 1232},
	title = {{Greedy function approximation: A gradient boosting machine}},
	volume = {29},
	year = {2001}
}

@book{koller2009,
address = {Cambridge, MA},
author = {Daphne Koller and Nir Friedman},
publisher = {The MIT Press},
title = {Probabilistic Graphical Models},
year = {2009}
}

@inproceedings{Davis_2021, 
title={Markov Network Structure Learning: A Randomized Feature Generation Approach}, 
volume={26}, 
booktitle={Proceedings of the AAAI Conference on Artificial Intelligence}, 
author={Van Haaren, Jan and Davis, Jesse}, 
year={2012}, 
pages={1148-1154} 
}

@article{Tolpin_Wood_2021,
	author = {Tolpin, David and Wood, Frank},
	journal = {Proceedings of the International Symposium on Combinatorial Search},
	number = {1},
	pages = {201-205},
	title = {Maximum a Posteriori Estimation by Search in Probabilistic Programs},
	volume = {6},
	year = {2021}
}

@article{wang2018,
	author = {Wang, Kainan and Bui-Thanh, Tan and Ghattas, Omar},
	journal = {SIAM J. Sci. Comput.},
	number = {1},
	pages = {A142-A171},
	title = {A Randomized Maximum A Posteriori Method for Posterior Sampling of High Dimensional Nonlinear Bayesian Inverse Problems},
	volume = {40},
	year = {2018}
}

@ARTICLE{hazan2019,
  author={Hazan, Tamir and Orabona, Francesco and Sarwate, Anand D. and Maji, Subhransu and Jaakkola, Tommi S.},
  journal={IEEE Transactions on Information Theory}, 
  title={High Dimensional Inference With Random Maximum A-Posteriori Perturbations}, 
  year={2019},
  volume={65},
  number={10},
  pages={6539-6560}
}

@INPROCEEDINGS{yao1977,
  author={Yao, Andrew Chi-Chin},
  booktitle={18th Annual Symposium on Foundations of Computer Science}, 
  title={Probabilistic computations: Toward a unified measure of complexity}, 
  pages={222-227},
  year={1977}
}

@book{downey2013,
    address = {London},
    author = {Rodney Downey and Michael Fellows},
    publisher = {Springer},
    title = {Fundamentals of parameterized complexity},
    year = {2013}
}

@inproceedings{oztok2015,
author = {Oztok, Umut and Darwiche, Adnan},
title = {A top-down compiler for sentential decision diagrams},
year = {2015},
booktitle = {Proceedings of the 24th International Conference on Artificial Intelligence},
pages = {3141–3148}
}
